\documentclass[12pt]{article}

\usepackage{amsmath,amssymb,mathtools,amsthm}
\DeclareMathOperator*{\argmax}{arg\,max}
\usepackage{bm}                       

\usepackage{times}                    
\usepackage{graphicx}
\usepackage{subcaption}
\usepackage{float}
\usepackage{placeins}
\usepackage{tabularx}
\usepackage[dvipsnames]{xcolor}       
\usepackage{graphicx}
\usepackage{multirow}
\usepackage{rotating}
\usepackage{subcaption} 
\usepackage{algorithm}
\usepackage{algpseudocode}            

\usepackage[authoryear]{natbib}

\usepackage{bbm}
\usepackage{latexsym}

\usepackage{hyperref}

\newcommand{\captionfonts}{\normalsize}
\makeatletter
\long\def\@makecaption#1#2{%
  \vskip\abovecaptionskip
  \sbox\@tempboxa{{\captionfonts #1: #2}}%
  \ifdim \wd\@tempboxa >\hsize
    {\captionfonts #1: #2\par}
  \else
    \hbox to\hsize{\hfil\box\@tempboxa\hfil}%
  \fi
  \vskip\belowcaptionskip}
\makeatother

\newcommand{\E}{\mathbb{E}}

\newcommand{\dd}{\mathsf{d}}

\newcommand{\bfx}{{\mathbf{x}}}
\newcommand{\bfy}{{\mathbf{y}}}

\newcommand{\bff}{{\mathbf{f}}}
\newcommand{\bfg}{{\mathbf{g}}}

\newcommand{\bmeps}{{\bm{\varepsilon}}}
\newcommand{\bmmu}{{\bm{\mu}}}

\newcommand{\bmpsi}{{\bm{\psi}}}

\newcommand{\bmeta}{{\bm{\eta}}}

\newcommand{\bmomega}{{\bm{\omega}}}
\newcommand{\bmlambda}{{\bm{\lambda}}}
\newcommand{\bmtheta}{{\bm{\theta}}}

\newcommand{\e}{\mathsf{e}}

\newcommand{\inv}{^{-1}}

\renewcommand{\exp}{\mathsf{exp}}
\renewcommand{\log}{\,\mathsf{log}\,}

\newcommand{\diag}{\mathsf{diag}}

\newcommand{\Id}[1]{{I}_{#1}}

\newcommand{\normal}{\mathsf{N}}

\newcommand{\gcx}{{\tilde{\mathbf{x}}}}
\newcommand{\gcy}{{\tilde{\mathbf{y}}}}

\newcommand{\gceps}{{\widetilde{\bm{\varepsilon}}}}
\newcommand{\gcmu}{{\widetilde{\bm{\mu}}}}

\newcommand{\vfe}{\mathsf{F}}

\begin{document}

\noindent\hfill 
\vspace{20mm}
\begin{center}
{\LARGE Online Generalised Predictive Coding}

\ \\
{\bf Mehran H. Z. Bazargani$^{ \dagger 1,2,3,4}$}, 
{\bf Szymon Urbas$^{ \dagger 5,6}$},
{\bf Adeel Razi$^{ 3,4,7,8}$},
{\bf Thomas Brendan Murphy$^{ 1,2}$},
{\bf Karl Friston$^{8}$}

\vspace*{2mm}
\noindent$^{\dagger}$Equal contribution

\vspace*{2mm}
\noindent {$^{1}$Insight Research Ireland Centre, UCD, Dublin, Ireland.}\\
{$^{2}$UCD School of Mathematics and Statistics, UCD, Belfield, Dublin, Ireland.}\\
{$^{3}$Turner Institute for Brain and Mental Health, School of Psychological Sciences, Monash University, Melbourne, Victoria, Australia.}\\
{$^{4}$Monash Biomedical Imaging, Monash University, Melbourne, Victoria, Australia.}\\
{$^{5}$Department of Mathematics \& Statistics,  Maynooth University, Maynooth, Ireland.}\\
{$^{6}$Hamilton Institute, Maynooth University, Maynooth, Ireland.}\\
{$^{7}$CIFAR Azrieli Global Scholars Program, Toronto, Canada.}\\
{$^{8}$Queen Square Institute of Neurology, Dept. Imaging Neuroscience. University College London, London WC1N 3AR, UK.
}
\end{center}

\thispagestyle{empty}
\markboth{}{NC instructions}

\vspace*{0mm}

\begin{center} {\bf Abstract} \end{center}
This paper introduces an extension of generalised filtering for online applications. Generalised filtering refers to data assimilation schemes that jointly infer latent states, learn unknown model parameters, and estimate uncertainty in an integrated framework---e.g., estimate state and observation noise---at the same time (i.e., triple estimation). This framework appears across disciplines under different names, including variational Kalman-Bucy filtering in engineering, generalised predictive coding in neuroscience, and Dynamic Expectation Maximisation (DEM) in time-series analysis. Here, we specialise DEM for ``online'' data assimilation, through a separation of temporal scales. We describe the variational principles and procedures that allow one to assimilate data in a way that allows for a slow updating of parameters and precisions, which contextualise fast Bayesian belief updating about the dynamic hidden states. Using numerical studies, we demonstrate the validity of online DEM (ODEM) using a non-linear---and potentially chaotic---generative model, to show that the ODEM scheme can track the latent states of the generative process, even when its functional form differs fundamentally from the dynamics of the generative model. Framed from a neuro-mimetic predictive coding perspective, ODEM offers a biologically inspired solution to online inference, learning, and uncertainty estimation in dynamic environments.

\vspace{2mm}

\section{Introduction} \label{sec:introduction}
Despite being confined within the interior darkness of the skull, the human brain possesses a remarkable ability to interpret, understand and analyse the world \textit{out there}, plan for unseen futures, and make decisions that can alter the course of events. This extraordinary capability is conjectured to come from the brain's function as a predictive machine, constantly inferring the hidden causes of its sensory inputs to maintain a coherent model of its environment. This view, which dates back to Helmholtz's idea of ``perception as unconscious inference'' \citep{helmholtz1866_optics}---evolving into the ``Bayesian brain'' hypothesis \citep{doya2007bayesian}---suggests that the brain operates as a constructive statistical organ. It updates its beliefs about the external world based on incoming sensory data under a generative model (GM). The GM furnishes the brain with a structured representation that supports probabilistic beliefs over both the latent dynamical states of the external world, corresponding to the generative process (GP), as well as the observation mappings through which these states give rise to sensory signals. Essentially, the brain continually refines its probabilistic beliefs about both the latent states and the causal mechanisms of the world through a process of \textit{online triple estimation}, jointly optimising beliefs over: hidden states, model parameters, and their associated uncertainties in accordance with the principles of Bayesian inference \citep{bayes_theorem,parr2022active}. More technically, given a sensory observation $\bfy_t$ at time $t$, perception can be formulated as an online triple estimation scheme, whose three components are: 1) online hidden state inference, 2) online parameter learning, and 3) online uncertainty estimation, all three of which are the core components of our proposed online generalised PC scheme and are elaborated in Section.~\ref{sec:odem}. The main challenge of online triple estimation can be attributed to its online nature where once an observation $\bfy_t$ is assimilated by the GM at time $t$, it cannot be revisited by the GM at a later stage. Moreover, upon receiving $\bfy_t$, the GM does not have the opportunity to iteratively refine its internal states, parameters, or precision estimates until convergence, before proceeding to the subsequent observation $\bfy_{t+1}$ at time $t+1$. Consequently, this setting necessitates the development of computational schemes that enable rapid and accurate updates of hidden states, model parameters, and uncertainty estimates under stringent temporal constraints.

One of the most promising frameworks for developing brain-inspired computation is through the Free Energy Principle (FEP) \citep{friston2010free}, an information-theoretic principle which posits that the brain operates to minimise a quantity known as the \textit{Variational Free Energy} (VFE). Specifically, VFE provides an upper bound on the negative logarithm of the Bayesian model evidence, defined as $-\log(p(\bfy_t|M))$, where $M$ is the GM and $\bfy_t$ is the observation. Under variational Laplace (detailed in Section.~\ref{sec:optimisation}), the functional form of the VFE reduces to a sum of precision-weighted prediction errors and complexity terms (detailed in Section.~\ref{sec:optimisation}). The FEP suggests that the brain seeks to reduce this discrepancy to sustain a non-equilibrium steady state, allowing the ``self'' to survive and persist, adaptively, over time in an unpredictable and dynamic environment. In this paper, we introduce an online generalised predictive coding (PC) framework---inspired by neuronal message-passing mechanisms in the brain and derived from the FEP---to model human perception as an online triple estimation problem within a PC formulation.

The main contributions are:
\begin{enumerate}
    \item The \emph{Online Dynamic Expectation Maximisation (ODEM)} algorithm for online triple estimation. 
    \item A Python/Pytorch implementation of ODEM, which is available in a public code repository, for our proposed online generalised predictive coding scheme, that is, ODEM \footnote{\href{https://github.com/MLDawn/ODEM}{https://github.com/MLDawn/ODEM}}.
\end{enumerate}

Crucially, ODEM is an innovation upon the original Dynamic Expectation Maximisation (DEM) algorithm introduced in \cite{FrTr2008}, by bringing its flexible filtering scheme into an ``online'' inference paradigm. In contrast to the online triple estimation in \cite{MeWi2021} and \cite{MeWi2022}, the proposed scheme considers non-linear dynamics in the state space. The scheme highlights the benefit of utilising generalised coordinates of motion (GCM) (detailed in Section.~\ref{sec:gcm}) to model the dynamics of the hidden state trajectories without the need for closed-form transition densities or Monte Carlo simulation, unlike Kalman filters or particle filters. The online parameter/\allowbreak hyperparameter inference is done by applying variational Bayesian updates to the posteriors, where the mean-field posterior approximation at one time point serves as the prior at the next. This approximation can incur additional bias compared to an exact method like SMC$^2$ \citep{ChJa2013}; however, it results in a constant computational cost per iteration, thus allowing a truly online implementation.

The remainder of the paper is organised as follows. Section~\ref{sec:Inference, learning, and uncertainty estimation} introduces the three components of triple estimation. Section~\ref{sec:generalised preditive coding} presents the framework of generalised predictive coding. Section~\ref{sec:related work} reviews the relevant literature. Section~\ref{sec:optimisation} describes the optimisation scheme employed in this work. Section~\ref{sec:odem} introduces the proposed Online Dynamic Expectation Maximisation (ODEM) framework. Section.~\ref{sec: experimental design} describes the experimental design. Section~\ref{sec: experiment results} presents and analyses the experimental results. Finally, Section~\ref{sec:conclusion_and_futurework} concludes the paper, and discusses future work.

\section{Inference, learning, and uncertainty estimation}\label{sec:Inference, learning, and uncertainty estimation}
For a brain-inspired model to function effectively in a dynamic world, it must continuously adapt to new sensory inputs (i.e., observations). It requires a GM that encapsulates and reflects the dynamics of the GP, which causes the \textit{incoming stream} of sensory data, in real-time. The GP is not directly accessible to the GM, much like how the true external world lies beyond sensory organs and is hidden from our brain. Thus, determining the hidden states of the world directly is not possible, and therefore becomes an inference problem, where the model seeks to reverse-engineer the GP from observed sensory inputs. More specifically, the GM only receives the sensory stimuli caused by the GP which are impressed upon the boundary that the GM shares with the GP. This boundary is also referred to as the \textit{Markov Blanket}. Indeed the process of inferring the most likely values for the hidden states of the GP, given the sensory stimuli that would land on the Markov blanket is referred to as \textit{model inversion} (i.e., perception).

Let $\bmpsi_t$ denote the set of all three quantities to be inferred at time $t$; i.e.,\  hidden states, the GM parameters and GM estimation of its own uncertainty. By Bayes' theorem, the posterior distribution of $\bmpsi_t$ given the observed data $\bfy_t$, is expressed as: $p(\bmpsi_t|\bfy_t)=p(\bfy_t|\bmpsi_t)p(\bmpsi_t)/p(\bfy_t)$. The calculation of the \emph{Bayesian model evidence} term, $p(\bfy_t)$, often involves a complex, high-dimensional integral that is generally intractable, which makes the exact calculation of the posterior belief, $p(\bmpsi_t|\bfy_t)$, infeasible. Under such circumstances, perception as \textit{exact} Bayesian inference is computationally intractable. To circumvent this difficulty, \textit{approximate} Variational Inference (VI) is considered, which provides a simpler optimisation problem. In VI, the aim is to find a surrogate distribution $q(\bmpsi_t)$ that approximates the true intractable posterior $p(\bmpsi_t|\bfy_t)$ by minimising the VFE \citep{friston2010free}: 
\begin{equation}\label{VFE}
        \vfe(q;\bfy_t) = \underbrace{D_{\mathrm{KL}}[q(\bmpsi_t)\|p(\bmpsi_t)]}_{\text{Complexity}}-\underbrace{\mathbb{E}_{q(\bmpsi_t)}[\log p(\bfy_t|\bmpsi_t)]}_{\text{Accuracy}}
                        = -\mathbb{E}_{q(\bmpsi_t)}\left[\log\frac{ p(\bfy_t,\bmpsi_t)}{q(\bmpsi_t)}\right],
\end{equation}
where $D_{\mathrm{KL}}$ is the \textit{Kullback-Leibler} divergence. Crucially, the intractable model evidence term, $p(\bfy_t)$, is now absorbed into the variational objective in Eq.\eqref{VFE}, where the optimisation variable is the approximate posterior 
$q(\bmpsi_t)$. Note that VFE is simply a negative Evidence Lower Bound, $\text{VFE}=-\text{ELBO}$. Thus, minimising VFE (i.e., maximising ELBO) serves two purposes: 1) It approximates the otherwise intractable model evidence, and 2) its value provides a robust criterion for selecting among different GMs. Crucially, since VFE is a functional of $q$ (i.e.,~it takes in a function and returns a scalar), the calculus of variation is used for its minimisation \citep[e.g.,~Chapter 10 of][]{BiNa2006}. The VFE balances two opposing quantities: The \textit{accuracy} term, which ensures that the model's predictions match the observed data closely, and the \textit{complexity} term, which penalises overly complex models that might overfit. Specifically, complexity measures the extent to which the approximate posterior belief, $q(\bmpsi_t)$ (after having observed, $\bfy_t$) will diverge from the prior belief of the model (before having seen, $\bfy_t$) about the state of the world, $p(\bmpsi_t)$. Thus, the more a GM needs to ``move'' from its prior belief in order to explain the observations, the more complex those explanations will be. By minimising the VFE, the model achieves an optimal trade-off between fitting the data and maintaining simplicity, adhering to the principle of Occam's razor, that is, the GM will provide the least complex and yet most accurate account for the observed data.

Online triple estimation is essential for constructing truly brain-inspired models for human perception, which can adapt and generalise across different contexts, much like biological neural networks. Our proposed approach, the Online Dynamic Expectation Maximisation (ODEM) scheme, is capable of addressing the challenging task of triple estimation in an online fashion. In the next section we will discuss generalised PC as a key component of ODEM.

\section{Generalised predictive coding}\label{sec:generalised preditive coding}
To maintain stability (i.e.,\ homeostasis),  and ensure survival, biological systems like the brain must continuously minimise the dissipation or entropy of their states. This process is akin to minimising the brain's ``surprise'' about its sensory states, which from a statistical perspective translates to maximising the Bayesian model evidence for its sensory input---a process known as Bayesian filtering. Predictive coding (PC) \citep{friston2005theory,bogacz2017tutorial, rao1999predictive, millidge2024predictive} is a prominent and neurobiologically-plausible approach to Bayesian filtering, which frames the brain's function as a constant interplay between sensory data prediction and error correction.
Under the PC framework, the brain is seen as a hierarchical GM that optimises its internal model of the world by minimising prediction error. This is accomplished by minimising the VFE or equivalently, maximising the ELBO, as we will discuss later. In this framework, perception is conceptualised as the minimisation of precision-weighted prediction errors through the continual updating of expectations (i.e., top-down predictions) that propagate down the cortical hierarchy. Predictions flow downward from deeper cortical layers to more superficial ones, while the resulting precision-weighted prediction errors travel upward, refining the brain's expectations and improving future predictions. In essence, the brain functions as a self-correcting system, constantly seeking to reduce the discrepancies between its expectations and sensory reality, thereby optimising its internal model (i.e., the GM) of the world---through online triple estimation. For more detail on hierarchical generalised PC models see Appendix.~\ref{app:Hierarchical generalised predictive coding}. In its current implementation, the ODEM scheme utilises a one-layer PC model for tackling the online triple estimation problem.

The VFE provides a bound approximation for the Bayesian model evidence, whose functional form---under certain simplifying conditions---is equivalent to a series of precision-weighted prediction error terms. This simplification is achieved using the Laplace approximation \citep[e.g.][]{ZeFr2023}, a method that approximates the posterior distribution of the model with simpler Gaussian distributions by matching the modes and the curvatures about them. Under the variational paradigm, the Laplace approximation leads to the mean-field approximate posterior being represented as a product of Gaussian terms. This greatly simplifies the computation and the consequent gradient-based optimisation of the VFE, allowing for efficient inference in a biologically plausible manner.

At its simplest form, online generalised PC scheme is based on a simple single-layer PC network
\begin{equation}\label{eq:ssm}
\tfrac{\dd}{\dd t} \bfx_t = \bff(\bfx_t, \bmtheta) + \bmomega_x(t)~~~\mbox{and}~~~    \bfy_t = \bfg(\bfx_t, \bmtheta) + \bmomega_y(t),
\end{equation}
Such a PC network underlying GM in ODEM and it is mathematically equivalent to a basic continuous State Space Model (SSM), where both the hidden states and observations take continuous real values in continuous time. The temporal evolution of the generalised hidden states $\gcx_t$, and their relationship to observations $\bfy_t$, are defined through a system of stochastic differential equations parametrised by $\bmtheta$. Crucially, the two random processes $\bmomega_y(t)$ and $\bmomega_x(t)$ are assumed to be independent (e.g.,~\citep{friston2010generalised}). In the most basic case, these could be Wiener processes \citep[e.g.][]{CoMi1965}, but other smoother processes such as the Mat\'ern process could be used here as well \citep[e.g.][]{HaSa2010}. Hereafter, we use shorthand notation: $\bfx_t' = \frac{\dd}{\dd t} \bfx_t$, $\bfx_t'' = \frac{\dd^2}{\dd^2 t} \bfx_t$, etc. 

The first equation is referred to as the \textit{state dynamics equation} and it describes the evolution of hidden states over time through a deterministic function $\bff(\bfx_t, \bmtheta)$ and stochastic fluctuations  $\bmomega_x(t)$, where the evolution of the hidden states can be modelled as differential equations, e.g.,~the change from some $t_0>0$ to some other $t_1>t_0$ comprises infinitesimally small increments in time. The second equation is sometimes referred to as the \textit{observation model} and it expresses how the observations are \textit{believed} to be generated from the hidden states through a deterministic function $\bfg(\bfx_t, \bmtheta)$ and stochastic fluctuations $\bmomega_{y}(t)$. Interestingly, if we assume the fluctuations to be zero-mean Gaussian processes, these two equations form a GM that underwrites the \textit{Kalman-Bucy filter} \citep[e.g.][]{ruymgaart2013mathematics} in the engineering literature. Importantly, even though we may be observing this continuous state space model at discrete times, the underlying dynamics of the system are continuous in time (e.g.,~the evolution of the hidden states, VFE minimisation, etc.). Here, the hidden states and their motion are collapsed into one latent variable $\gcx_t=\{ \bfx_t, \bfx_t'\}$, and the corresponding surrogate posterior distribution becomes $q(\gcx_t)$. The $\gcx_t$ is known as the generalised states, since it allows us to generalise the inference of the hidden states to inferring its velocity, $\bfx_t'$, in addition to its position $\bfx_t$. As we will see in the next section, one can push this further and generalise to even higher orders of motion. Indeed, this extended version with higher orders of motion is our proposed online generalised PC model.

\subsection{Higher orders of Generalised Coordinates of Motion}\label{sec:gcm}
As we have seen in Eq.~\eqref{eq:ssm}, in its most basic form, a GM uses the position, $\bfx$, and the velocity, $\tfrac{\dd}{\dd t} \bfx_t$ in its state dynamics equation. However, higher orders of motion (i.e., acceleration, jerk, snap, crackle, pop, etc.) will equip the GM with the ability to capture the dynamics of the hidden states more accurately without the need for problem-dependent closed-form transition densities---this is the core foundation of our proposed online generalised PC model. This is an important factor in designing the GM, as---depending on the form of the \textit{true} dynamics in the GP---higher orders of the generalised motion may be necessary. This will result in the expansion of the SSM in Eq.~\eqref{eq:ssm}, and can be looked at as a potential improvement on the original simple continuous SSM. Importantly, random fluctuations within the data-generating mechanism in the GP are generally assumed to have uncorrelated increments over time (i.e.,~Wiener assumption); however, in most complex systems (e.g.,~biological systems)---where the random fluctuations themselves are generated by some underlying dynamical system---these fluctuations possess a certain degree of smoothness. Indeed, by relaxing the Wiener assumption and imposing smoothness on the model functions $\bff(\,\cdot\,)$ and $\bfg(\,\cdot\,)$, we have the opportunity to not only consider the rate of change of the hidden state, $\bfx_t'$, and the observation but also their corresponding higher-order temporal derivatives (i.e.,~acceleration, jerk, etc.); see, for example, \citep{friston2010generalised}. 

The resultant pair of $\{\bfx_t, \bfx_t', \bfx_t'',\ldots\}$ and $\{\bfy_t, \bfy_t', \bfy_t'',\ldots\}$ are called the generalised coordinates of motion \citep[GCM; ][]{balaji2011bayesian}, which provide an opportunity for further capturing the dynamics that govern the evolution of the hidden states and observations. An estimated trajectory over time can be calculated using a Taylor series expansion around the present time, which results in a function that can extrapolate to the near future as well as remember the recent past. To motivate this, we note that Eq.~\eqref{eq:ssm} implies that if we treat the time derivative of $\bfx_t$ as another variable, its conditional distribution is $\frac{\dd}{\dd t}\bfx_t|\bfx_t \sim \normal\left(\bff\left(\bfx_t, \bmtheta\right), \Pi_x\inv\right),$ where $\Pi_x$ is some precision matrix---determined by $\bmomega_x$---to be estimated (i.e., uncertainty estimation). We can extend this to construct hierarchical temporal relationships between higher-order time-derivatives of $\bfx_t$, if we assume that $\bfx_t$ itself is analytic in $t$ and thus has continuous derivatives. We use the standard notation $\bfx'_t = \frac{\dd}{\dd t}\bfx_t$, $\bfx''_t = \frac{\dd^2}{\dd^2 t}\bfx_t$, ..., and concatenate the latent variables to $\gcx_t = \left(\bfx_t,\bfx'_t,\bfx''_t,\ldots\right)^\top$. While $\gcx_t$ can in principle be infinite dimensional, for all practical purposes it is truncated at some order, because the precision of generalised fluctuations falls quickly with increasing order. The truncation is a modelling choice that reflects prior expectations about the smoothness of random fluctuations; i.e., smooth processes require higher orders of motion because the precision of the generalised motion falls more slowly with their order. The $\gcx_t$ vector could in fact be thought of being made up of sufficient statistics characterising the latent-space process. To describe the GM for these additional terms, we will differentiate Eq.~\eqref{eq:ssm} to obtain the following approximate relationships:
\begin{align*}
        \bfx''_t &= \frac{\dd}{\dd t}\left[\bff\left(\bfx_t, \bmtheta\right) + \bmomega_x(t)\right]\\
        &=\left[\nabla_x\bff\left(\bfx_t, \bmtheta\right)\right]\bfx'_t + \bmomega_x'(t),
\end{align*}
where $\nabla_x\bff\left(\bfx_t, \theta\right)$ denotes the Jacobian $\frac{\partial \bff}{\partial \bfx}$. Under a local linearity assumption on $\bff(\,\cdot\,)$ (its Hessian is close to zero), we instead can approximate higher derivatives by
\begin{align*}
\bfx'''_t &\approx \left[\nabla_x\bff\left(\bfx_t, \bmtheta\right)\right]\bfx''_t + \bmomega_x''(t),\\
\bfx^{(4)}_t &\approx \left[\nabla_x\bff\left(\bfx_t, \bmtheta\right)\right]\bfx'''_t + \bmomega_x'''(t),\\
&~\vdots\\
\bfx^{(k_x+1)}_t &\approx \left[\nabla_x\bff\left(\bfx_t, \bmtheta\right)\right]\bfx^{(k_x)}_t + \bmomega_x^{(k_x)}(t).
\end{align*}
Applying the same assumption to $\bfg(\,\cdot\,)$, we have 
\begin{align*}
\bfy'_t & \approx \left[\nabla_x\bfg\left(\bfx_t, \bmtheta\right)\right]\bfx'_t + \bmomega_y'(t),\\
\bfy''_t & \approx \left[\nabla_x\bfg\left(\bfx_t, \bmtheta\right)\right]\bfx''_t + \bmomega_y''(t),\\
&~\vdots\\
\bfy^{(k_y+1)}_t & \approx \left[\nabla_x\bfg\left(\bfx_t, \bmtheta\right)\right]\bfx^{(k_y)}_t + \bmomega_y^{(k_y)}(t).
\end{align*}
We can collate the above equations into
\begin{align*}
    D\gcx_t &\approx \tilde{\bff}\left(\gcx_t,\bmtheta\right)+\widetilde{\bmomega}_x(t),\\
    \gcy_t &\approx \tilde{\bfg}\left(\gcx_t,\bmtheta\right)+\widetilde{\bmomega}_y(t),
    \end{align*}
where $D$ is a matrix with  identity matrices, $\Id{d_x}$, above the leading block diagonals and 0 elsewhere, and $\widetilde{\bmomega}_x(t) = (\bmomega_x(t), \bmomega'_x(t),...)^\top$ and $\widetilde{\bmomega}_y(t) = (\bmomega_y(t), \bmomega'_y(t),...)^\top$.

\begin{figure}[H]
    \centering
    \begin{subfigure}{\textwidth}
        \centering
        \includegraphics[width=0.85\textwidth]{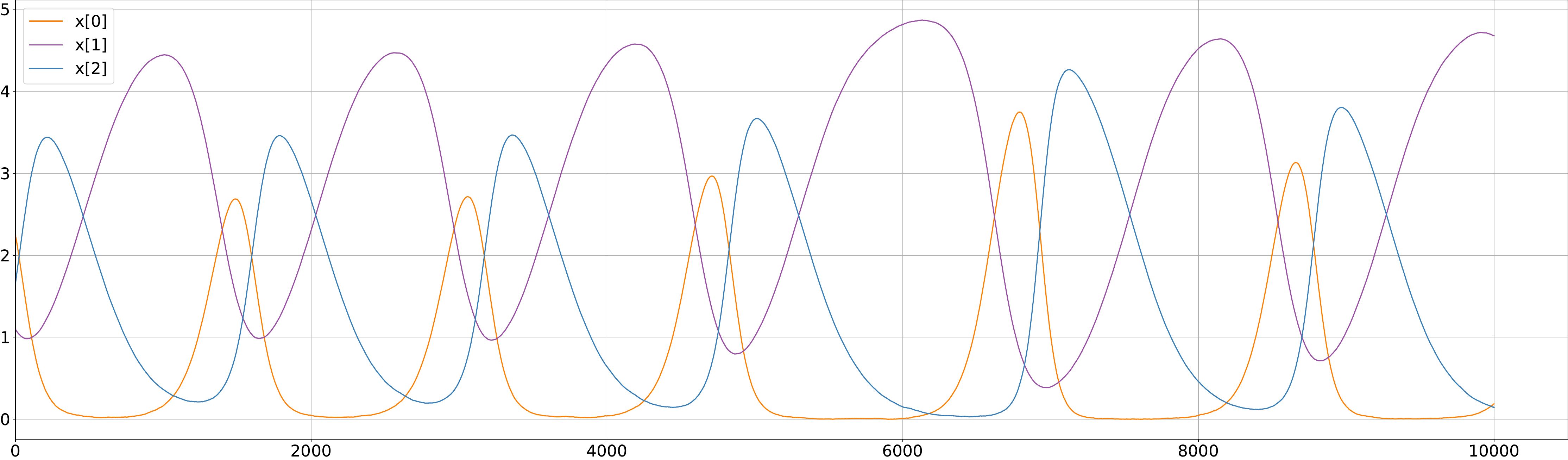}
        \caption{A realisation of a GLV-GP, that is, the true states with smooth state noise, $x$.}
        \label{fig:x_noisy}
    \end{subfigure}
    
    \vspace{0.35cm}

    \begin{subfigure}{\textwidth}
        \centering
        \includegraphics[width=0.85\textwidth]{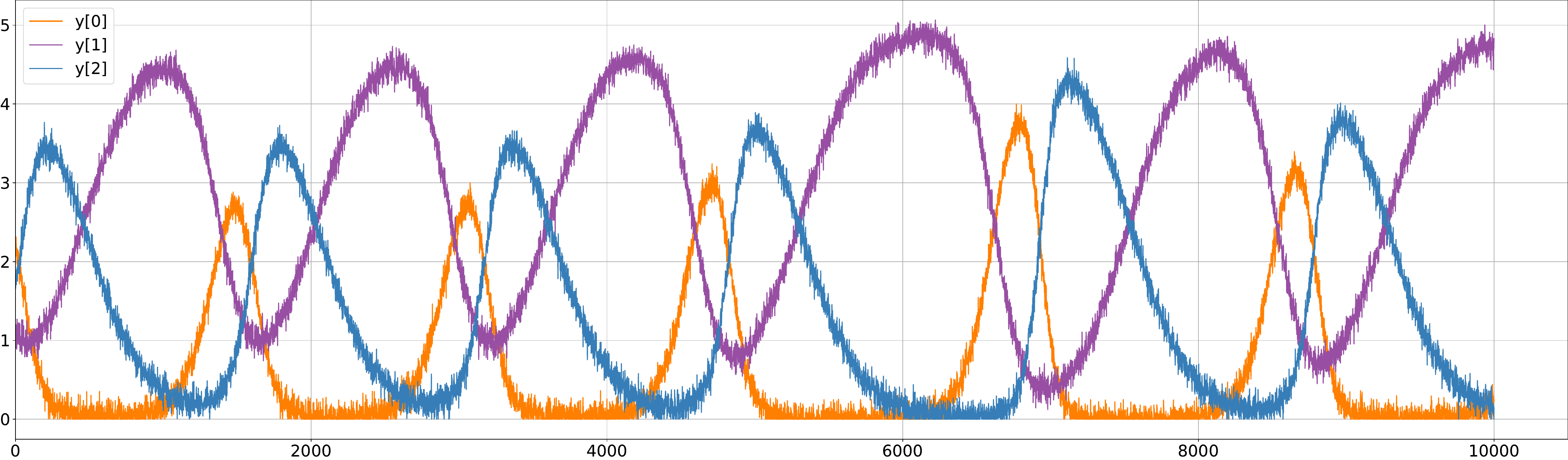}
        \caption{The observations $y$ with smooth observation noise.}
        \label{fig:y_noisy}
    \end{subfigure}

    \caption{A realisation of a GLV-GP state trajectories with smooth state noise (top), and observations with smooth observation noise (bottom).}
    \label{fig:x_noisy_y_noisy}
\end{figure}

The generalised observations, $\gcy_t$, are constructed by augmenting each discrete observation with approximations to its higher-order temporal derivatives. Let $\bfy_t$ denote the observation at discrete time index $t$, where successive samples are separated by a fixed time step $dt$. The first temporal derivative is approximated using a finite difference scheme as $\bfy_t' \approx \frac{\bfy_t - \bfy_{t-1}}{dt}$. Higher-order derivatives are then constructed recursively by applying the same operation to the lower-order derivative estimates. For example, the second derivative is approximated as $\bfy_t'' \approx \frac{\bfy_t' - \bfy_{t-1}'}{dt}$. Proceeding in this manner yields successive approximations to higher orders of motion. These quantities are then stacked to form the vector of generalised observations, $\gcy_t = \{\bfy_t, \bfy_t', \bfy_t'',\ldots\}$, which provides a local description of the temporal evolution of the observations in GCM. In our simulations, the same time step $dt$ used by the GP to produce the data is also used by the GM to construct these finite-difference approximations. Under a smooth Gaussian assumption on $\bmomega_x(t)$ and $\bmomega_y(t)$, the equations will result in approximate conditionals 
\begin{align}
    D\gcx_t \mid \gcx_t
    &\sim \normal\left(
        \tilde{\bff}\left(\gcx_t, \bmtheta\right),
        \Pi_{\gcx}^{-1}
    \right)
    \label{eq:generalised_PC_ssm} \\
    \gcy_t \mid \gcx_t
    &\sim \normal\left(
        \tilde{\bfg}\left(\gcx_t, \bmtheta\right),
        \Pi_{\gcy}^{-1}
    \right),
    \nonumber
\end{align}
where $\Pi_\gcx = S_{k_x}(\sigma^2)\otimes\Pi_x$ and $\Pi_\gcy = S_{k_y}(\sigma^2)\otimes\Pi_y$; $\Pi_x$ and $\Pi_y$ are the marginal precisions of $\bmomega_x(t)$ and $\bmomega_y(t)$ respectively and  $S_k(\sigma^2)$ is the ``smoothness'' matrix with elements corresponding to covariances among all derivatives of the  respective noise processes at stationarity, up to order $k$ \citep[e.g.,~Section 2.1 of][]{DaDa2025}. Eq.~\eqref{eq:generalised_PC_ssm} is the main component of our proposed online generalised PC model. The dependence between the temporal derivatives of the latent and the observed processes are modelled explicitly through the generalised precisions $\Pi_\gcx$ and $\Pi_\gcy$. This dependence structure is what drives the inference of the dynamics present in the predictive coding formulation. In order to provide an intuition for noisy state trajectories and noisy observations, Fig.~\ref{fig:x_noisy} and Fig.~\ref{fig:y_noisy} show hidden states of a Generalised Lotka-Volterra (GLV) \citep{malcai2002theoretical} generative process (i.e., GLV-GP) and the corresponding generated observations with smooth state noise and observation noise, respectively. Further information on how the noisy states and observations are generated will be provided in Section.~\ref{sec:state_observation_smooth_noise}.

\section{Related work}\label{sec:related work}
The task of triple estimation can also be implemented in an off-line fashion. This is where the filtering scheme has the opportunity to go over the entire sensory time-series as many times as required until it has converged (i.e.,~minimised VFE). In contrast to off-line implementations, in an online implementation, the GM may only ever process each datum of the series once; the scheme cannot iteratively improve upon the inference and parameter estimation it by considering the whole (or partial) series. In biological terms, the GM does not have the luxury of re-visiting any part of the sensory stimuli time-series, as it cannot go back in time and re-visit the previously sensed sensory stimuli in the past.
This makes online triple estimation especially challenging. This is why in this work, human perception has been modelled as an \textit{online} triple estimation task, since the way we perceive the sensory stimuli is also purely online: once a stimulus has been sampled and sensed, there is no going back. In this section, we briefly review online methods in the literature. For a comprehensive review of offline methods, see, for example, \cite{Chen2003}, \cite{KaDo2015} or \cite{Gods2019}.

\cite{ChJa2013} introduced the SMC$^2$ algorithm which, in principle, is able to perform online parameter inference for an SSM by computing the parameter (and hyperparameter) posteriors at each consecutive time point. The inference is exact in the Monte Carlo sense; however, the statistical efficiency of the scheme relies on a linearly increasing computational cost, limiting any widespread use. Methods such as those in \cite{DeJa2017} incur a quantifiable bias to SMC$^2$ in order to improve the efficiency of the algorithm.

Variational methods offer computationally-efficient alternatives at the price of only performing approximate parameter inference. \cite{CaSh2021} provides a flexible variational scheme for inferring hidden states along with parameter-learning. By exploiting the time-reversible Markovian structure of the model, a fixed-cost inference algorithm is obtained. Interestingly, whilst in principle the scheme could be used for complete triple-estimation, in the work, state-space noise has been treated as fixed; that is, the covariance matrices appearing in the Gaussian transition and emission densities are assumed to be known.

\cite{MeWi2021} tackles the triple estimation problem using DEM \citep{FrTr2008}, focusing on the problem of noise precision estimation. It was found that a carefully set-up model is able to perform accurate inference state inference and parameter/\allowbreak precision hyperparameter estimation, when utilising generalised coordinates. \cite{MeWi2022} implements DEM for a tracking task where the underlying model is linear (i.e.,\ Kalman filter). \cite{MeLa2023} further innovates by considering an adaptive scheme for the precision estimation task based on Gauss-Newton updates; a key contribution is the provision of sufficient conditions for the existence of globally optimal precision estimates.

\section{Optimisation}\label{sec:optimisation}
The ultimate objective of the online generalised PC network is to continuously estimate the hidden states, parameters and noise precision hyperparameters by minimising the VFE, given an observation, $\bfy_t$, at each time point. The main challenge is that the exact calculation of the VFE is computationally intractable, unless certain simplifying assumptions are made so that the VFE can be expressed in closed-form. This is where VL can be used to approximate the VFE for filtering, parameter learning and uncertainty estimation. We first re-parametrise the hyperparameters to be on $\mathbb{R}$ which will allow us to approximate the posterior with a multivariate Gaussian which results in a fully tractable approximate VFE. The optimal mean-field approximation to the posterior is obtained through gradient-based minimisation of the VFE; hidden states are inferred continuously as the data come in, and the parameters/hyperparameters posteriors are updated once every set number of time points by accumulated gradient updates

We parametrise the generative model covariance matrices in Eq.~\eqref{eq:generalised_PC_ssm} as
\[
\Pi_x
= \diag\!\bigl(\e^{\lambda^x_1},\ldots,\e^{\lambda^x_{d_x}}\bigr)\in \mathbb{R}^{d_x \times d_x},
\qquad
\Pi_y
= \diag\!\bigl(\e^{\lambda^y_{1}},\ldots,\e^{\lambda^y_{d_y}}\bigr)\in \mathbb{R}^{d_y \times d_y},
\]
where $d_x$ and $d_y$ are the dimensionalities of the hidden states and observations, respectively; we let $\bm{\lambda}= (\lambda^x_1,\ldots,\lambda^x_{d_x},\lambda^y_{1},\ldots,\lambda^y_{d_y})$
From the data, we wish to estimate
$\left(\widetilde{\bfx}_t,\,\bm{\theta},\,\bm{\lambda}\right)\eqqcolon\bm{\psi}$.

For the posterior inference of the quantities, we must specify appropriate parameter priors
$\pi_\theta(\bm{\theta})$ and $\pi_\lambda(\bm{\lambda})$; these will be $\normal(\bmeta_\theta,\Pi_\theta)$ and $\normal(\bmeta_\lambda,\Pi_\lambda)$ respectively.
We let the approximating mean-field density be the product density
\[
q_{\bm{\psi}}(\bm{\psi} ; \bmmu_{\bm{\psi}},\Sigma_{\bm{\psi}})
= q_x(\gcx_t; \gcmu_x,\Sigma_x)\,
  q_\theta(\bm{\theta}; \bmmu_\theta,\Sigma_\theta)\,
  q_\lambda(\bm{\lambda}; \bmmu_\lambda,\Sigma_\lambda).
\]
Under the Laplace approximation, each of those factors is a multivariate Gaussian density. 
The full VFE is
\begin{equation}
\vfe\left[q_{\bm{\psi}}(\,\cdot\,;\bmmu_{\bm{\psi}},\Sigma_{\bm{\psi}}); \gcy_t\right]
= \E_{q_{\bm{\psi}}(\,\cdot\,;\bmmu_{\bm{\psi}},\Sigma_{\bm{\psi}})}
\left[-\log p(\bm{\psi}, \gcy_t) + \log q(\bm{\psi};\bmmu_{\bm{\psi}},\Sigma_{\bm{\psi}})\right]
\label{eqn:vfe_exact}
\end{equation}
which is intractable apart from special cases. However, we eschew this by applying the Laplace approximation: the mean of
$q_{\bm{\psi}}$ is set to be the posterior mode,
$\bmmu_{\bm{\psi}} = \argmax_{\bm{\psi}} \, p(\bm{\psi}|\gcy_t)= \argmax_{\bm{\psi}} \, p(\bm{\psi},\gcy_t)$, and the optimal choice of
mean-field covariances conditional on $\bmmu_{\bm{\psi}}$ is achieved at
\begin{equation}
\Sigma_{\bm{\psi}_i}^{-1}
= \left.\frac{\partial^2}{\partial \bm{\psi}_i^2}U(\bm{\psi}, \gcy)\right|_{\bm{\psi}=\bmmu_{\bm{\psi}}},
\qquad
\bm{\psi}_i\in\{\gcx_t, \bm{\theta}, \bm{\lambda}\},\label{eqn:cov_update}
\end{equation}
where $U(\bm{\psi}, \gcy)=-\log p(\bm{\psi}, \gcy)$
\citep[e.g.][]{ZeFr2023}. The approximate VFE objective function based on the distribution assumptions
given in Eq.~\eqref{eq:generalised_PC_ssm} and the Laplace approximation is
\begin{align}
\vfe_L
\coloneqq&\tfrac{1}{2}\Bigl(
\gceps^\top\Pi_\gceps\gceps
+\bmeps_\theta^\top\Pi_\theta\bmeps_\theta
+\bmeps_\lambda^\top\Pi_\lambda\bmeps_\lambda
-\log\!\bigl(|\Pi_\gceps||\Pi_\theta||\Pi_\lambda||\Sigma_x||\Sigma_\theta||\Sigma_\lambda|\bigr)
+d_yk_y\log 2\pi
\Bigr)\nonumber\\
=&\underbrace{\tfrac{1}{2}\Bigl(
\gceps_x^\top\Pi_\gcx\gceps_x
+\bmeps_\theta^\top\Pi_\theta\bmeps_\theta
+\bmeps_\lambda^\top\Pi_\lambda\bmeps_\lambda
+\log\!\bigl(|\Sigma_x||\Sigma_\theta||\Sigma_\lambda||\Pi_\gcx||\Pi_\theta||\Pi_\lambda|\bigr)
\Bigr)}_{\eqqcolon\,\text{Complexity}}\nonumber\\
&~~~\quad -\underbrace{\tfrac{1}{2}\Bigl(
-\gceps_y^\top\Pi_\gcy\gceps_y
+\log|\Pi_\gcy|
-d_yk_y\log 2\pi
\Bigr)}_{\eqqcolon\,\text{Accuracy}},
\label{eq:vfe_under_vl}
\end{align}
where $\gceps = \left(\gceps_y,\gceps_x\right)^\top = \left(\gcy - \tilde{\bfg}\left(\widetilde\bmmu_{x}, \bmmu_\theta\right),D\widetilde\bmmu_{x} - \tilde{\bff}\left(\widetilde\bmmu_{x}, \bmmu_\theta\right)\right)^\top$, $\bmeps_\theta = \bmmu_\theta-\bmeta_\theta$ and $\bmeps_\lambda = \bmmu_\lambda-\bmeta_\lambda$;
see Appendix \ref{app:errors_example} for an example of an explicit form. It is essential to characterise the distinct roles of the accuracy and complexity terms, and the trade-off between them, in VFE minimisation. The objective of a GM is to minimise its own VFE, which can be decomposed into an accuracy term and a complexity term. Minimisation of VFE therefore entails maximising accuracy—i.e., providing a precise account of the observed data under the model—while simultaneously minimising complexity—i.e., constraining departures from prior beliefs. Accordingly, the optimal GM is one that explains the data accurately while doing so in the simplest possible manner. This balance formalises the principle of parsimony: models that achieve high accuracy with minimal complexity are preferred because they avoid unnecessary adjustments to prior beliefs. Importantly, this simplicity underwrites generalisation. In the absence of a complexity penalty, the GM would be free to arbitrarily adapt its beliefs to the data, thereby overfitting and compromising predictive performance on unseen observations.

From a technical perspective, the complexity term corresponds to the Kullback–Leibler divergence between the posterior and prior distributions over latent variables and parameters. It quantifies the extent to which posterior beliefs must diverge from prior expectations in order to explain the data. In this sense, complexity measures the effective degrees of freedom utilised by the GM during inference. When prior beliefs are informative and well-aligned with the data-generating process (i.e.,~ the GP), posterior updates can remain close to the prior, yielding high accuracy at a low complexity cost. Consequently, belief updating becomes both efficient and minimally expressive, reflecting an optimal balance between data fit and model parsimony.

The standard VFE can be derived and minimised, during the time intervals between observations. More specifically, after observing $\gcy_t$, we can minimise the integration of point estimates of the VFE along a continuous time interval until the next observation $\gcy_{t+\Delta}$, $\Delta>0$. This quantity is called \emph{Free Action (FA)} \citep{friston2011action} and it is defined as $\overline{\mathcal{A}}[q(\gcx)]=\int_{t}^{t+\Delta} \vfe(q(\gcx_s);\gcy_s) \,\mathsf{d}s$, and it is an upper bound on the accumulated surprise, $
-\int_{t}^{t+\Delta} \log p(\gcy_s)) \,\mathsf{d}s$, over the same time period. In practice, one does not observe the data in continuous time and as such $\gcy_t$ is used to approximate $\{\gcy_s, s\in[t,t+\Delta)\}$ in the integrand. By minimising $\overline{\mathcal{A}}$ in-between observations, the GM is constantly minimising the VFE along a path of length $\Delta$, and thus continuously striving to improve the approximate estimation of the posteriors over the hidden states, parameters and noise precision hyperparameters.

\section{Online Dynamic Expectation Maximisation (ODEM)}\label{sec:odem}
We have implemented an online generalised predictive coding scheme through our proposed method: namely, \textit{Online Dynamic Expectation Maximisation (ODEM)}, designed to tackle the \textit{online} triple estimation problem. Since ODEM uses generalised predictive coding as its underlying GM, its neuronal message-passing is divided into two types: 1) Top-down predictions and 2) Bottom-up (precision weighted) prediction error signals. More specifically, given a sensation $\bfy_t$, ODEM strives to keep track of the GP by inferring the hidden states (known as the \emph{D}-step), learning the parameters of the underlying GM (known as the \emph{E}-step) and estimating its own uncertainty regarding both the state dynamics and observations (known as the \emph{M}-step).

There is a crucial assumption of a \textit{separation of temporal scales} in ODEM, which is consistent with principles of Haken’s \emph{Synergetics} \citep{haken2004introduction}, particularly the \textit{slaving principle} \citep{haken1996slaving}. The field of synergetics studies how macroscopic order emerges in complex systems (e.g., brain) composed of many interacting microscopic components. A key result is that, near instabilities or phase transitions, a small set of slowly evolving macroscopic variables---referred to as \textit{order parameters}---emerge and constrain, or “enslave”, the dynamics of the many fast microscopic degrees of freedom. This mechanism naturally induces a hierarchy of temporal scales: fast microscopic processes rapidly relax toward a low-dimensional manifold defined by the order parameters, whose dynamics evolve on slower timescales. In the context of biological self-organisation, this provides a principled motivation for separating inference processes according to their intrinsic temporal scales. Fast neuronal dynamics can support rapid state estimation (fast \emph{D}-step), while slower processes---such as synaptic plasticity (slow \emph{E}-step) and neuromodulatory control of uncertainty (slow \emph{M}-step)---govern parameter learning and precision modulation. Accordingly, our ODEM implementation introduces a separation of temporal scales in which slow macroscopic variables---namely the generative model parameters and precision hyperparameters---coordinate and constrain faster microscopic dynamics corresponding to the generalised hidden states.

This natural \emph{separation of temporal scales} has been introduced into the ODEM scheme by updating the state, parameters and uncertainty estimations at different temporal scales. Here, the details of each of the \emph{D}, \emph{E}, and \emph{M} steps are presented along with the pseudo-code of the ODEM scheme.

\textbf{The \emph{D}-step}: In this step, given \emph{every} observation $\bfy_t$, the estimation of the dynamic hidden states of the world are continuously and instantly inferred and updated. Crucially, the GCM (See. Section.~\ref{sec:gcm}) are used to track the dynamics of the states of the world. As discussed above, in ODEM, it is assumed that the states of the world change fast (but smoothly). As such, after each observation, $\bfy_t$, a \emph{D}-step is taken, which facilitates the continuous tracking of the external fast moving states of the world by the the internal hidden states of the GM. Specifically, given observation $\bfy_t$ at time $t$, during a \emph{D}-step, the estimations of the dynamic hidden states are updated using the gradient of VFE at time $t$. The fast moving states can be likened to the fast moving neuronal activity in the brain. The \emph{D}-step is done according to the Ozaki scheme \citep{ozaki19852} which uses curvature information of the VFE surface to regularise the gradient, resulting in a more efficient and robust update rule \citep[e.g.][]{ZeFr2023}. Specifically, the mean-field parameter of the state-space vector is updated through
 \begin{equation}\gcmu_x \gets \gcmu_x + J_0^{-1} \left[\exp\left(J_0 \Delta s\right)-I\right]\left(D\gcmu_x - \kappa\left.\nabla_{\gcmu_x} \vfe_L(q; \gcy_t)\right|_{\gcmu_x}\right),\label{eqn:ozaki}\end{equation}
 where $J_0\coloneqq \mathsf{J}\left[D\gcmu_x - \kappa\nabla_{\gcmu_x} \vfe_L(q; \gcy_t)\right]_{\gcmu_x}$ is the Jacobian matrix, $\Delta s$ is a step-size parameter and $\kappa$ is the Lagrange multiplier; Section \ref{sec:experiment_setting} discusses the specific settings for these tuning parameters. The Lagrange multiplier $\kappa$ represents the influence of the VFE gradient on the updates relative to the ``momenta'' given by $D\gcmu_x$; low $\kappa$ yields updates largely driven by the existing dynamics, and high $\kappa$ yields updates primarily based on the gradient information.

\textbf{The \emph{E}-step}: This refers to parameter learning, where the goal is to learn the parameters, $\bmtheta$ , of the GM. The parameters endow the GM with the ability to model the slowly moving aspect of the GP (e.g., the slowly moving context upon which faster dynamics can unfold as content). In ODEM, it is assumed that the parameters change orders of magnitude more slowly (i.e., slowly changing context) than the states (i.e., quickly unfolding content). Consequently, the \emph{E}-step happens less frequently than the \emph{D}-step and this equips ODEM with prior knowledge of the fact that there are slowly moving contexts in the world that it needs to track, in order to track the fast moving content. In contrast to the instantaneous updates performed in the \emph{D}-step, the parameter estimates in the \emph{E}-step are updated using the accumulated gradient of the VFE. Specifically, these gradients are aggregated over a window---whose size determines the separation of temporal scale and it is denoted by $\mathsf{inter}_{EM}$ in Algorithm.\ref{alg:odem}---of previously observed data, such that parameter updates reflect information integrated across multiple observations rather than a single time point. The slowly evolving parameters can be likened to the strength of the synaptic weights in the brain, since they too update more slowly than fast neuronal activity.

\textbf{The \emph{M}-step}: This allows the ODEM scheme to estimate the noise precision of both the sensations and the states of the world. Crucially, by examining the likelihood and state noise precision terms,$\Pi_{\gcy}$ and $\Pi_{\gcx}$ , in the definition of the VFE under VL (see Eq.~\eqref{eq:vfe_under_vl}) one can see how the ability to change these estimated precisions can allow for modelling an \textit{attention} mechanism (i.e., attending to particular error terms in the VFE in Eq.~\eqref{eq:vfe_under_vl}). In ODEM, we are assuming that the observation and state noise precisions are moving slowly (at the same time scale as the parameters), and as such, the \emph{M}-step for updating the observation and state noise precision estimates takes place at the same temporal scale as the \emph{E}-step. Similarly, the noise precision estimation in the \emph{M}-step are updated using the accumulated gradient of the VFE. Specifically, these gradients are aggregated over a window---whose size is determined by $\mathsf{inter}_{EM}$ in Algorithm.\ref{alg:odem}---of previously observed data, such that the updates for noise precision estimates reflect information integrated across multiple observations rather than a single time point. The ensuing precision estimates can be likened to attentional selection in the brain, whereby precision determines the weighting of prediction errors during inference. In this sense, the interpretation of precision as attention is primarily normative—arising from the role of precision in optimising the weighting of prediction errors within the predictive coding framework—while also being consistent with neurobiological observations that shifts in attentional allocation, often mediated by neuromodulatory systems, evolve on slower timescales than the fast neuronal dynamics that encode moment-to-moment activity. Consequently, the rate at which the brain can deploy attention across cortical regions is typically slower than the rapid neuronal activity underlying ongoing perceptual processing.

\begin{algorithm}[H]
\caption{Online Dynamic Expectation Maximisation (ODEM), where the \emph{D}-step occurs at every observation but the \emph{E} and \emph{M} steps occur at every $\mathsf{inter}_{EM}$ observations.}
\label{alg:odem}
\begin{algorithmic}[1]
\State $\mathrm{FA} \gets 0$ \Comment{\textbf{Initialise Free Action}}
\For{$t, \bfy_t \in \mathcal{Y}$} \Comment{\textbf{Loop over observations}}
    \State $\gcy_t = \mathsf{generalise}(\bfy_t,\mathsf{d}t)$ \Comment{\textbf{By finite differentiation (Section.~\ref{sec:gcm})}}
    
    \State $J_0^{x}\gets \mathsf{J}\left[D\gcmu_x - \kappa\nabla_{\gcmu_x} \vfe_L(q; \gcy_t)\right]_{\gcmu_x}$ \Comment{\textbf{Calculate Jacobian}}
    
    \State Update $\gcmu_x$ according to Eq. \eqref{eqn:ozaki} \Comment{\textbf{\emph{D}-step (Section.~\ref{sec:odem})}}

    \State $\bmlambda_{acc} \gets \beta_\Lambda \bmlambda_{acc} + (1 - \beta_\Lambda)\left.\nabla_{\bmmu_\lambda} \vfe_L(q; \gcy_t)\right|_{\bmmu_\lambda}$ \Comment{\textbf{$\Lambda$ gradients accumulation}}
    \State $\bmtheta_{acc} \gets \beta_\theta \bmtheta_{acc} + (1 - \beta_\theta)\left.\nabla_{\bmmu_\theta} \vfe_L(q; \gcy_t)\right|_{\bmmu_\theta}$ \Comment{\textbf{$\Theta$ gradient accumulation}}

    \If{$t \bmod \mathsf{inter}_{EM} == 0$} \Comment{\textbf{E \& M step interval}}
        \State $j \gets \left\lfloor \dfrac{t + 1}{\mathsf{inter}_{EM}} \right\rfloor$ \Comment{\textbf{EM-iteration index (integer division)}}

        \State $\{\alpha^\lambda_j, \alpha^\theta_j\} 
       = \left\{\frac{\alpha^\lambda}{(j + t_0^{\lambda})^{\gamma^\lambda}},\; 
                \frac{\alpha^\theta}{(j + t_0^{\theta})^{\gamma^\theta}} \right\}$ \Comment{\textbf{Robbins-Monroe learning rate}}

        \State $\bmmu_\lambda \gets \bmmu_\lambda - \alpha^\lambda_j \bmlambda_{acc}$ \Comment{\textbf{\emph{M}-step}}

        \State{$\Sigma_{\lambda} \gets U(\gcy_t, \gcmu_x, \bmmu_\theta, \bmmu_\lambda)_{\lambda\lambda}^{-1}$}
        \Comment{\textbf{Hyperparameter covariance}}


        \State $\bmeta_\lambda,\Pi_\lambda \gets \bmmu_\lambda, \Sigma_\lambda^{-1}$ \Comment{\textbf{Prior update for $\Lambda$}}  

        \State $\bmmu_\theta \gets \bmmu_\theta - \alpha^\theta_j \bmtheta_{acc}$ \Comment{\textbf{\emph{E}-step}}

        \State{$\Sigma_{\theta} \gets  U(\gcy_t, \gcmu_x, \bmmu_\theta, \bmmu_\lambda)_{\theta\theta}^{-1}$}
        \Comment{\textbf{Parameter covariance}}

        \State $\bmeta_\theta,\Pi_\theta \gets \bmmu_\theta, \Sigma_\theta^{-1}$ \Comment{\textbf{Prior update for $\Theta$}}
        
        \State $\{\bmtheta_{acc}, \bmlambda_{acc}\} \gets \{\bm{0}, \bm{0}\}$ \Comment{\textbf{Reset the accumulated gradients}}
    \EndIf
    \State $\Sigma_x \gets U(\gcy_t, \gcmu_x, \bmmu_\theta, \bmmu_\lambda)_{\tilde{x}\tilde{x}}^{-1}$
    \Comment{\textbf{State covariance}}
    \State $\mathrm{FA} \gets \mathrm{FA} + \vfe_L\left(q(\,\cdot\,;\gcmu_x, \bmmu_\theta, \bmmu_\lambda, \Sigma_{x}, \Sigma_\theta, \Sigma_\lambda); \gcy_t\right)$ \Comment{\textbf{Update Free Action}}
\EndFor
\end{algorithmic}
\end{algorithm}

\section{Experimental design}\label{sec: experimental design}

Having established the variational principles and procedures behind ODEM, this section provides the big picture of how the experiments are designed to provide a validation of the ODEM scheme and illustrate its application. In brief, to generate data for online assimilation, we chose a normal form autonomous dynamical system used ubiquitously in the life sciences; namely, a \textbf{Generalised Lotka-Volterra (GLV)} generative process (GLV-GP). The tracking and identification of such systems by a given GM can be extremely difficult; especially if the parameters and precisions used by the GP for generating the data are unknown. We have therefore constrained the problem by using two GMs with the ``same'' and ``different'' functional forms to that of the GP, namely, a Generalised Lotka-Volterra (GLV) generative model (GLV-GM) and a Lorenz generative model (Lorenz-GM), and we show that ODEM is capable of online triple estimation by inverting generative models of a given functional form using data generated by a GP of the same and---even more interestingly---different functional form. As we will see below, this functionality rests upon working in GCM (detailed in Section.~\ref{sec:gcm}). More specifically, only when equipped with higher orders of motion, under the ODEM scheme, the Lorenz-GM was able to infer the dynamics generated by a GLV-GP.

The implementation of ODEM rests upon prior constraints that render the simultaneous state inference, uncertainty estimation and parameter learning tractable. Crucially, these constraints include implicit beliefs about how fast the precision and parameter posterior estimates change---with respect to the state posterior estimates---which is entailed by the frequency of corresponding updates. This means that the choice of the scheduling (i.e., $\mathsf{inter}_{EM}$ in Algorithm.~\ref{alg:odem}) becomes a tuning parameter---of a discrete sort---that itself has to be optimised with respect to model evidence. This affords the opportunity to illustrate a final level of model optimisation within ODEM; namely, Bayesian model selection (BMS).

In what follows, we first describe how the GP is designed and solved in order to generate the trajectory of noisy hidden states (to be inferred by the GM) and how by adding smooth observation noise to this trajectory, the observation time-series can be created.  To keep things simple, we used a fixed level of observation and state noise throughout, when generating data using the GLV-GP. Then the functional form of two GMs---to which these observations are fed---are introduced. After detailing how Bayesian prior belief is chosen over the parameters and precision hyperparameters of the GMs in Table.~\ref{tab:priors_parameters} and Table.~\ref{tab:prior_hyperparameters}, the tuning parameters are introduced in Table.~\ref{tab:tuning_parameters}. Importantly, we then describe how BMS is used to select the best GM over the grid of tuning parameters, over two different experiment scenarios. These scenarios correspond to whether the GM had the same functional form as the GP, or not, under low ($k_x=2$) and high ($k_x=3$) orders of GCM. Crucially, we have also considered different prior precision ratios $C:=\frac{E_{\Pi_y}}{E_{\Pi_x}}$, which represents the degree to which the GM trusts the observations compared to the dynamics during inference. We have found that performance was very sensitive to the value of $C$.

In order to clarify how the experiments are conducted---and how the results of the best GMs are reported---we provide an illustrative example.

Let us say we have a Lorenz-GM---with prior belief defined over $rho$ according to Table.~\ref{tab:priors_parameters}---that will observe the observation time-series generated by the GLV-GP. Here are the main steps:
\begin{enumerate}
    \item We use ODEM to execute online triple estimation for Lorenz-GM using the data generated by GLV-GP for 1512 combinations (i.e., $2\times3\times4\times3\times3\times7=1512$) of the tuning parameters in Table.~\ref{tab:tuning_parameters}. This results in 1512 GMs---all of whose functional form are Lorenz systems---with 1512 FA values (later used in BMS).
    \item Given the major effect of the $C$ parameter and the order of motion, $k_x$, on the performance of ODEM, we have reported the best performing GM under each $k_x$ and $C$ settings. For instance, for a given order of motion (e.g., $k_x=2$), and a given value of $C$ (e.g., $C=1/10$) we find the GM with the lowest FA and then report its: 1) inferred states, 2) learned parameters and 3) estimated precisions.
\end{enumerate}

The remainder of this section elaborates on the details of the experimental design.

\subsection{GP and GM with smooth state noise}\label{sec:state_observation_smooth_noise}
The generative process (GP) is a Generalised Lotka-Volterra (GLV) process, which is an expressive mathematical framework describing how multiple species interact and change in population over time, using a system of differential equations to capture growth rates, carrying capacities, and pairwise interactions. The GLV system was selected due to its smooth non-linear dynamics, which admit well-defined higher-order temporal derivatives and thereby enable the explicit use of higher orders of motion within the GM. Furthermore, the GLV formulation provides a flexible yet interpretable dynamical system with non-linear interactions among states, making it a convenient test-bed for evaluating inference schemes that operate in generalized coordinates of motion. We consider the following variation of a three-dimensional GLV system:
\[
\dot{\boldsymbol{x}} = \boldsymbol{x} \odot A\boldsymbol{x},
\qquad 
\boldsymbol{x} \in \mathbb{R}^3 > 0,
\]
where \(A \in \mathbb{R}^{3 \times 3}\) is the interaction matrix. Each element of \(A\) encodes how one population influences another, while the diagonal entries typically represent self-interaction terms. The symbol \(\odot\) denotes the \emph{element-wise (Hadamard) product}, such that for two given vectors $\boldsymbol{m}$ and $\boldsymbol{n}$: \((\boldsymbol{m} \odot \boldsymbol{n})_i = m_i n_i\) for each component \(i\). To obtain oscillations in the three-species dynamics, we choose \(A\) to be \emph{anti-symmetric}, i.e.,\ \(A = -A^\top\). This configuration removes any net amplification or decay, leading to energy-neutral rotations in the state space rather than convergence to a fixed point. Specifically, the $A$ matrix in the GLV-GP is chosen to be:
\[
A = 
\begin{bmatrix}
0.0 & 0.2 & -0.4 \\
-0.2 & 0.0 & 0.1 \\
0.4 & -0.1 & 0.0
\end{bmatrix}
\]

Substituting these into the standard GLV formulation yields the final GLV-GP model (see. Appendix.~\ref{app:glv_gp}).

State noise was produced by adding a coloured (i.e., correlated) noise to the drift terms, $\dot{x}$, for the GLV-GP. This means adding noise to the integration process when solving the GP, which will result in noisy trajectories of external hidden states, $x$; the coloured noise is generated by convolving a smoothing kernel with a Wiener process (details in the code). Crucially, we inject independent realisations of the noise process into each of the three channels of the GLV-GP, ensuring that stochastic perturbations are channel-specific. Finally, in order to solve the system of ODE's in the GP, we have used the forward Euler method for integration and we have picked a step size of $\dd t=0.01$ for a total duration of $T=100$, which generates noisy trajectories of the external hidden states of size $100/0.01=10,000$.

\textbf{Generating observations with smooth observation noise}: The observations, $\bfy_t$ are generated by adding a coloured (i.e., correlated) noise to the noisy solutions, $\bfx_t$, of the GLV-GP. Mutually independent noise processes are applied to the components of the hidden states, $\bfx_t$. Again, each noise process is a smooth Gaussian process with a squared-exponential covariance kernel \citep[e.g.][]{ScSp2018}; simulation details are given in the accompanying code. 

\textbf{Generative model (GM)}: We will consider GLV and Lorenz attractor as the functional forms of the state dynamic equations of two separate GMs. This allows us to evaluate a scenario where the functional form of the GM is the same as the GP (i.e.,~GLV-GM vs. GLV-GP) as well as when it is not (i.e.,~ Lorenz-GM vs. GLV-GP). Let us first discuss the functional form of the state dynamics equations (i.e., the $\bff(\,\cdot\,)$ function). 

The Lorenz system is a canonical non-linear dynamical system, which exhibits deterministic chaos, and it is widely used as a benchmark for evaluating inference and learning algorithms in non-linear SSMs. The system dynamics are defined by the following set of coupled ordinary differential equations:
\begin{align*}
\dot{x}_0 &= \sigma \left(x_1 - x_0\right), \\
\dot{x}_1 &= x_0 \left(\rho - x_2\right) - x_1, \\
\dot{x}_2 &= x_0 x_1 - \beta x_2,
\end{align*}
where \(x_0, x_1, x_2\) denote the hidden states, and \(\sigma > 0\), \(\rho > 0\), and \(\beta > 0\) are system parameters. The parameter \(\sigma\) controls the strength of the coupling between \(x_0\) and \(x_1\), effectively setting the rate of energy transfer between these modes. Crucially, the parameter \(\rho\) acts as a bifurcation parameter that controls the qualitative regime of the system dynamics. For sufficiently small values of \(\rho\), the system converges to stable fixed points, whereas larger values of \(\rho\) give rise to chaotic behaviour characterized by sensitive dependence on initial conditions and complex attractor geometry.

The functional form of the GLV-GM is $\dot{\boldsymbol{x}} = \boldsymbol{x} \odot A\boldsymbol{x}$, which is isomorphic to the GLV-GP.

Importantly, the observation model of both GM's is always as identity mapping, that is, $\bfg(\bfx)=\bfx$, which means that $\bfg(\,\cdot\,)$ has no parameters. The learnable parameters will be outlined in the next section.

\subsection{Priors over parameters}
In all of the experiments, and for both GM's, the priors are always defined as Normal distributions, in accordance with variational Laplace \citep{ZeFr2023}. Since, the $\bfg(\,\cdot\,)$ function is defined as identity for both GM's, we will be only focusing on the parameters of the $\bff(\,\cdot\,)$ function, i.e.,\ state dynamics equations for both GM's.

For the Lorenz-GM, we focus on learning $\rho$ from observations, as it directly modulates the transition between ordered and chaotic regimes. Accurately inferring \(\rho\) therefore provides a stringent test of the model’s ability to capture and adapt to changes in the underlying dynamical structure of the GP. We have fixed the values of the remaining two parameters as such: \((\sigma,~ \beta) =(10,~ 8/3)\) and define a Gaussian prior over the $\rho$ parameter (see Table.~\ref{tab:priors_parameters}).

For the GLV-GM, we have fixed the values of the $r$ vector to be zero. We have further assumed an anti-symmetric form for the $A$ matrix, where the learnable parameters are only the elements lying above the diagonal, for which we define appropriate priors. Since $A$ is assumed as anti-symmetric, the elements under the diagonal are simply fixed to the negated values of the estimations for the above-diagonal elements, making the total number of learnable parameters as 3 for the GLV-GM: $(a_{12},~a_{13},~a_{23})$. The parameter priors are given in  Table.~\ref{tab:priors_parameters}.

\begin{table}[t]
\centering
\caption{The priors defined over the parameters of the Lorenz-GM and GLV-GM models.}
\label{tab:priors_parameters}

\renewcommand{\arraystretch}{1.1}

\begin{tabular}{c|c|p{5cm}}
\hline
\textbf{GMs} & \textbf{Parameters} & \multicolumn{1}{c}{\textbf{Priors}} \\ \hline

Lorenz 
& $\Theta = (\sigma = 10,\, \rho,\, \beta = 8/3)$
& $\rho \sim \mathcal{N}(30,\,81)$
\\ \hline

\multirow{3}{*}{GLV}
& \multirow{3}{*}{\raisebox{-0.5\height}{$
A=\begin{bmatrix}
0 & a_{12} & a_{13} \\
-a_{12} & 0 & a_{23} \\
-a_{13} & -a_{23} & 0
\end{bmatrix}$}}
& $a_{12} \sim \mathcal{N}(0.3,\,0.0625)$
\\

&
&
$a_{13} \sim \mathcal{N}(-0.2,\,0.0625)$
\\

&
&
$a_{23} \sim \mathcal{N}(0.3,\,0.0625)$
\\ \hline

\end{tabular}
\end{table}

\subsection{Priors over precision hyperparameters}\label{sec:Priors over precision hyperparameters}

Rather than specifying prior means $\eta_{\lambda^x}$ and $\eta_{\lambda^y}$ directly, we define them through expected precisions $E_{\Pi_x}\coloneqq\E[\e^{\lambda^x_i}]= \exp\!\left(\eta_{\lambda^x} + \sigma_{\lambda^x}^2/2\right)$ and $E_{\Pi_y}\coloneqq\E[\e^{\lambda^y_j}]= \exp\!\left(\eta_{\lambda^y} + \sigma_{\lambda^y}^2/2\right)$ for given prior variance terms $\sigma^2_{\lambda^x}$ and $\sigma^2_{\lambda^y}$ respectively. 

In PC formulations of variational inference, precision modulates the gain of prediction error signals and thus their influence on belief updates. A key quantity is the ratio between expected observation and state noise precisions,
\[
C \coloneqq \frac{E_{\Pi_y}}{E_{\Pi_x}},
\]
which determines the relative weighting of sensory prediction errors versus dynamical prediction errors during VFE minimisation. In other words, $C$ represents the degree to which the GM trusts the observations compared to the dynamics during inference. Larger values of $C$ bias inference toward closely tracking observations, whereas smaller values favour consistency with the latent dynamics encoded by the generative mode (c.f., Kalman gain).

In our experiments, we fix the expected state noise precision to $E_{\Pi_x}=500$ and vary the expected observation noise precision over a predefined grid, yielding seven distinct values of $C$. For both $\bmlambda^x$ and $\bmlambda^y$, we additionally consider two values of the log-precision standard deviation, resulting in a total prior grid of size $28$. All precision prior configurations are summarised in Table~\ref{tab:prior_hyperparameters}.

\begin{table}[t]
\centering
\caption{Priors over the observation and state noise precisions.}
\begin{tabular}{c|c}
\hline
\textbf{\boldmath $C 
\;\textnormal{with}\; E_{\Pi_x}=500$}
& \textbf{\boldmath $\sigma_{\lambda}$} 
\\ \hline

\begin{tabular}{c}
$E_{\Pi_y}=[10, 20, 50, 500, 5000, 12500, 25000]$ \\
$C = \left[\tfrac{1}{50}, \tfrac{1}{25}, \tfrac{1}{10}, 1, 10, 25, 50\right]$
\end{tabular}
&
\begin{tabular}{c}
$\sigma_{\lambda^y}=[0.1, 0.5]$ \\
$\sigma_{\lambda^x}=[0.1, 0.5]$
\end{tabular}
\\ \hline

\end{tabular}
\label{tab:prior_hyperparameters}
\end{table}

Given the strong influence of the precision ratio $C$ on inference behaviour, results are reported separately for each value of $C$. For each ratio, candidate GM's are evaluated over a grid of experimental settings, and the model achieving the lowest FA---i.e., path integral of VFE in Algorithm.\ref{alg:odem}---is selected  (see below). We then report state inference, parameter learning, and precision estimation results for this selected model, enabling a principled comparison of ODEM performance across different precision prior regimes. Care must be taken to avoid a potential Bayesian variant of the \emph{Heywood} case, where one of the posterior modes implies a zero-variance noise process \citep{Heyw1931,Faro2022}, which can happen in very flexible latent variable models such as the ones considered here. This would be reflected in the log-precision terms $\bmlambda^y$ being effectively infinite, and the inferred states $\bfx_t$ would be near equal to the observed data $\bfy_t$. Imposing stronger hyperparameter priors limits this risk of overfitting.

\subsection{Experimental settings}\label{sec:experiment_setting}
There are also a range of experiment settings, which play a key role in the numerical experiments. We defined a grid over a range of possible values for these settings.

\begin{table}[t]
\centering
\caption{Tuning parameters}
\begin{tabular}{c|m{5.5cm}|c}
\hline
\textbf{Tuning parameters} & \centering\textbf{Description} & \textbf{Value Range} \\
\hline
$(k_x, k_y)$ & Orders of motion along $x$ and $y$ & [(2,1), (3,2)] \\
\hline
$\kappa$ & Lagrange multiplier in GCM & [1.0, 0.5, 0.25] \\
\hline
$\mathsf{inter}_{EM}$ & EM update interval & [64, 128, 256, 512] \\
\hline
$\beta_\Lambda$ & Precision forgetting rate & [0, 0.1, 0.2] \\
\hline
$\beta_\theta$ & Parameter forgetting rate & [0, 0.1, 0.2] \\
\hline
$C$ & Prior precision ratio & $\left[\tfrac{1}{50}, \tfrac{1}{25}, \tfrac{1}{10}, 1, 10, 25, 50\right]$ \\
\hline
\end{tabular}
\label{tab:tuning_parameters}
\end{table}
From the full model in Eq.~\eqref{eq:generalised_PC_ssm}, the $k$-th order derivative of the hidden states, $\bfx_t^{(k)}$ depends on the observation $\bfy_t^{(k)}$ and its derivative $\bfx_t^{(k+1)}$. Thus the orders of motion for the states $k_x$ is always taken to be one more than the order of motion along observations, $k_y$, that is: $k_x = k_y +1$, and $k_y$ is chosen based on numerical stability. For instance, the order of motion being 3 implies $(k_x, k_y)=(3,2)$. The Lagrange multiplier hyperparameter, $\kappa$, is used in the $D$-step, in order to balance the influence of the higher order generalised state estimations, i.e., $D\gcmu_x$ and the current gradient of the VFE with respect to the generalised states, i.e., $\left.\nabla_{\gcmu_x} \vfe_L(q; \tilde{y}^t)\right|_{\gcmu_x}$ (see Eq. \eqref{eqn:ozaki} in Section.~\ref{sec:odem}). A low $\kappa$ encourages more exploration during tracking, as it affords more influence to the momentum endowed by higher orders of motion. The $\mathsf{inter}_{EM}$ hyperparameter determines the separation of temporal scales between the $D$-step and the $EM$-steps: the higher it is, the more the separation (used in the \textit{E \& M step interval} line of Algorithm.~\ref{alg:odem}). The $\beta_\Lambda$ and $\beta_\theta$ hyperparameters, control the level of exponential smoothing for the accumulation of the gradients of the VFE with respect to the precision and parameter estimates, respectively (used in lines \textbf{$\Lambda$ gradients accumulation} and \textbf{$\Theta$ gradients accumulation} of Algorithm.~\ref{alg:odem}). Finally, the  (log) descent step-size parameter of the \emph{D}-step is $\Delta s = \exp(\nu)/\alpha$ where $\nu=-4$ and $\alpha$ is dynamically determined based on the curvature of the VFE, following the recommendations in \cite{ZeFr2023}.

Other experimental settings remained fixed throughout the experiments:  the hyperparameter values for the Robbins-Monro scheduling $\alpha_j = {\alpha}/{(j + t_0)^{\gamma}}$ algorithm for adaptive learning rates, for both E \& M steps are set as: $(\alpha, t_0, \gamma) = (0.0001,10,0.3)$. Starting with a higher learning rates---and gradually decreasing it---is a good optimisation practice to encourage high exploratory behaviour before exploitation, and Robbins-Monro scheduling provides such scheduling. Furthermore, the standard deviation values of the white noise Gaussian kernels used for the state and observation noise (Wiener process) were $(0.05, 0.1)$, respectively, which correspond to precisions of $(399.99, 100)$. The kernel size and standard deviation of the smoothing kernel---for generating both the states and observation smooth noise---are $(51, 0.005)$, respectively. For each combination of experiment hyperparameter and precision prior ratio \textit{C}, we conducted the online triple estimation using ODEM, which results in an FA value.

\subsection{Model Selection}
Given an observation time-series, we can use the FA as a model selection criterion, where the model with the lowest FA is selected as the best model; in that it can provide the simplest and the most accurate account for the data (i.e., trade-off between complexity and accuracy). It is important to emphasise that there is no \textit{true} model of the world in any absolute sense, since for any given set of observations, there can exist many candidate GMs capable of providing plausible---and simpler---accounts of how the observations were generated. There is, however, a \textit{best} model--i.e., explanation---which is the one that minimises VFE within the assumed model space.

It should be noted that one cannot use the VFE or FA for comparing the performance of two GMs when they have a different order of motion. For instance, in the case where $GM_1$ and $GM_2$ have $k_x=2$ and $k_x=3$ orders of motion, respectively. The reason for this is that if we change the order of GCM, we are changing the data, as we expand the generalised observations, i.e., $\{\bfy_t, \bfy_t', \bfy_t'',\ldots\}$. Consequently, if the data has changed, we cannot compare the marginal likelihoods of the data. In other words, model evidence can only be used to compare GM's of the \textit{same} data.

The process of model selection using the FA values was as follows: suppose we have two different generative models $GM_1$ and $GM_2$, both of which are inverted given \emph{identical} data generated by the GP.
For each of the two GMs we perform a grid-search across different values of experiment tuning parameters in Table.~\ref{tab:tuning_parameters}. This gives a total of 1512 combinations (i.e., $2\times3\times4\times3\times3\times7=1512$), which can be thought of 1512 variations of $GM_1$ and 1512 variations of $GM_2$. Then given a fixed order of motion, $k_x$, and a prior precision ratio, $C$, we find the best $GM_1$ variation with the lowest FA, that is, ${GM_1}^*$). Then we the same for the other generative model and find ${GM_2}^*$. Then, since they have the same order of motion, we can use FA to compare them, hence, we pick the generative model with the lower FA.

\subsection{Experiment conditions}
We consider two main experimental scenarios, each with $k_x=2$ and $k_x=3$ orders of motion (constituting 4 different conditions):

\begin{itemize}
    \item \textbf{\textit{Scenario-different}: Lorenz-GM vs. GLV-GP}: This is the case where the functional form of the GM differs from that of the GP, and follows a Lorenz dynamics whereas the true form of the GP follows a GLV dynamic. This case is explored for $k_x=2$ and $k_x=3$ orders of GCM.

    \item \textbf{\textit{Scenario-same}: GLV-GM vs. GLV-GP}: In this case the functional form of the GM is identical to that of the GP, as it too is a GLV system. This case too is explored for $k_x=2$ and $k_x=3$ orders of GCM.
\end{itemize}

For a given experiment scenario, we examine all combinations of experimental settings (See. Table.~\ref{tab:tuning_parameters}) and precision prior ratios (See Table.~\ref{tab:prior_hyperparameters}). For each combination, we implement online triple estimations using ODEM. The outputs for each inversion are: FA value, trajectories of the inferred states, trajectories of parameters estimates, trajectories of the precision hyperparameter estimates (for both the states and observations) along with the posterior covariances for the parameters and precision hyperparameters. These encode the GM's confidence regarding the estimated parameter and precision hyperparameter estimates at any given time.

\section{Experimental results}\label{sec: experiment results}
In this section, we discuss the results for state/observation noise precision estimation, parameter learning and state inference. We also demonstrate how the FA and its components: 1) Accuracy and 2) Complexity, change with respect to the orders of motion over different precision prior ratios, \emph{C}. It is worth noting that, throughout all experiments, the reported results and corresponding plots are based on single runs. No experiments were conducted across multiple random initialisations, nor were the results averaged over different seeds.

\subsection{\texorpdfstring{Convergence for $\mu_{\lambda^x}$, $\mu_{\lambda^y}$ and $\mu_\theta$ under different orders of motion}{Convergence for mu lambda x, mu lambda y and mu theta under different orders of motion}}
The separation of temporal scales embedded in ODEM, assumes that both the observation/state noise precisions and parameters change orders of magnitude more slowly than the states. 

\begin{figure}[H]
    \centering

    \begin{subfigure}{0.9\textwidth}
        \centering
        \includegraphics[width=\linewidth]{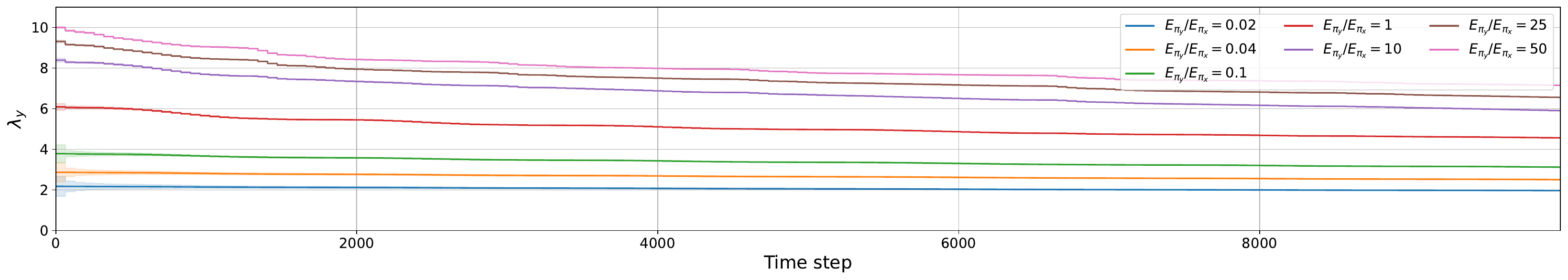}
        \caption{}
        \label{fig:lambda_y_scenario=1_kx=3_sample}
    \end{subfigure}

    \vspace{0.1cm}

    \begin{subfigure}{0.9\textwidth}
        \centering
        \includegraphics[width=\linewidth]{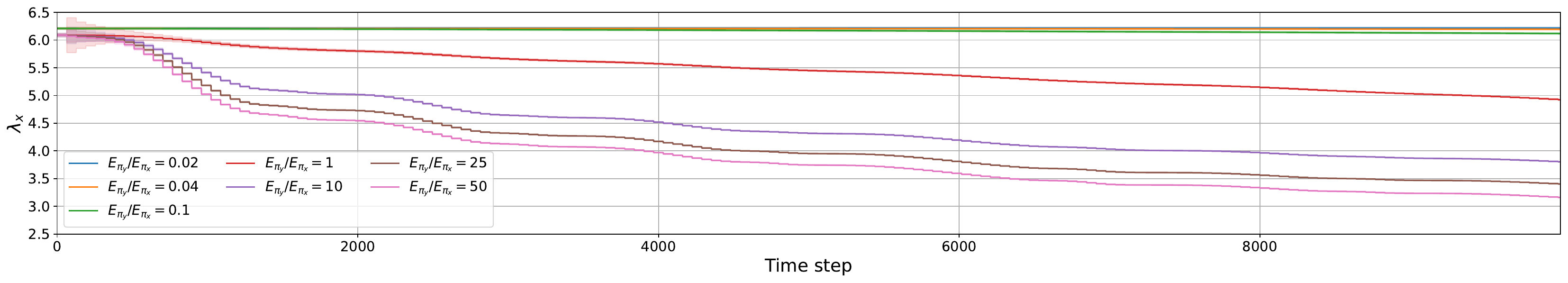}
        \caption{}
        \label{fig:lambda_x_scenario=1_kx=3_sample}
    \end{subfigure}

    \caption{The evolution of a) observation noise posterior estimate $\mu_{\lambda^y}$ and b) state noise posterior estimate $\mu_{\lambda^x}$, in scenario-different with $k_x=3$ orders of motion and along 7 precision prior ratios. The solid lines represent the posterior means at a given time, and the shaded bands represent credible regions within two standard deviations of the mean.}
    \label{fig:lambda_ylambda_x_kx=3_sample}
\end{figure}

We observed that for both $k_x=2$ and $k_x=3$ orders of motion, the posterior uncertainty estimate of the observation noise, $\mu_{\lambda^y}$, in scenario-same remains nearly constant. This indicates that when the underlying dynamics of the GM match those of the GP, the \emph{D}-step alone is largely sufficient for minimising the VFE, and consequently $\mu_{\lambda^y}$ requires minimal updating. In contrast, in scenario-different the posterior estimate $\mu_{\lambda^y}$ must also adapt to compensate for the mismatch between the dynamics of the GM and the GP. A similar behaviour is observed for the posterior estimate of the state noise, $\mu_{\lambda^x}$. Furthermore, the results show that, given sufficient time, the posterior estimates $\mu_{\lambda^y}$ and $\mu_{\lambda^x}$ eventually stabilise, consistent with the fact that the observation and state noise precisions in the GLV-GP were constant. An example of convergence of the posterior estimates over observation and state-noise precisions for $k_x=3$ orders of motion and scenario-different is provided in Fig.~\ref{fig:lambda_ylambda_x_kx=3_sample}, where the presented Bayesian credible intervals get narrower with the amount of data
; see Appendix.~\ref{app:lambda_y}, Appendix.~\ref{app:lambda_x} for equivalent results for all of the experiment conditions concerning $\mu_{\lambda^y}$ and $\mu_{\lambda^x}$.

Furthermore, for both $k_x=2$ and $k_x=3$ orders of motion, the posterior parameter estimates, $\mu_{\theta}$, were found to stabilise over time. This behaviour is consistent with the fact that the parameters in the GLV-GP are constant. Interestingly, increasing the order of motion from $k_x=2$ to $k_x=3$ results in a faster reduction of the posterior variance around $\mu_{\theta}$, particularly in scenario-different. This highlights the value of GCM in enabling the GM to reduce its uncertainty over parameter estimates. An example of convergence of the posterior over parameters is provided in Fig.~\ref{fig:theta_scenario1_kx3_sample},giving both the posterior means and Bayesian credible interval bands
See Appendix.~\ref{app:theta} for equivalent results for all of the experiment conditions concerning $\mu_{\theta}$.

\begin{figure}[H]
    \centering
    \includegraphics[width=0.9\textwidth]{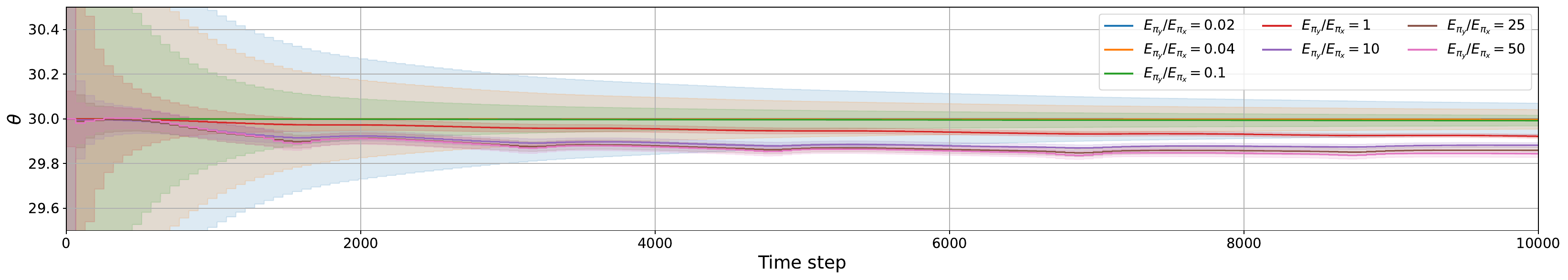}
    \caption{The evolution of the posterior expectation over $\rho$ in scenario-different, with $k_x = 3$ orders of motion across seven precision prior ratios.  The solid lines represent the posterior means at a given time, and the shaded bands represent credible regions within two standard deviations of the mean.}
    \label{fig:theta_scenario1_kx3_sample}
\end{figure}

\subsection{\texorpdfstring{Inference of states, $\mu_x$, under different orders of motion}{Inference of states, mu x, under different orders of motion}}

\begin{figure}[H]
    \centering
    \begin{subfigure}{0.9\textwidth}
        \centering
        \includegraphics[width=\textwidth]{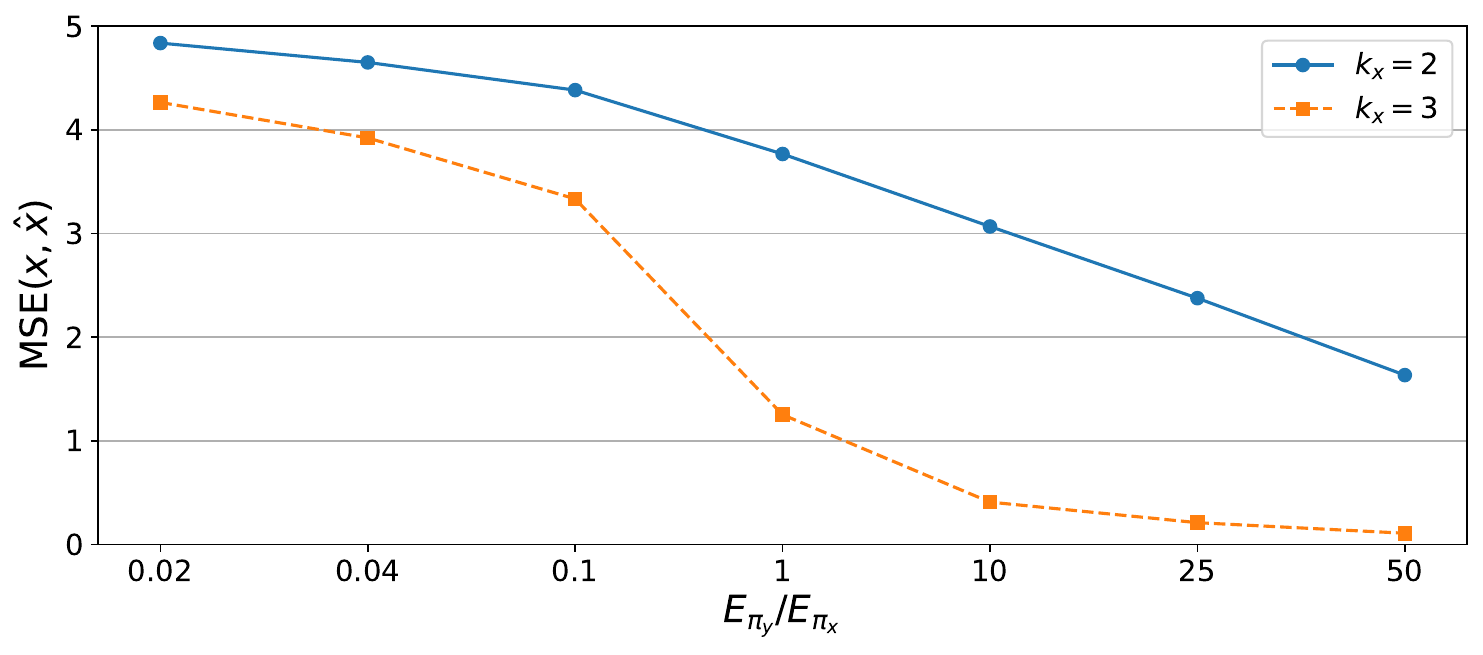}
        \caption{}
        \label{fig:mse_lorenz_glv}
    \end{subfigure}

    \vspace{0.1cm}

    \begin{subfigure}{0.9\textwidth}
        \centering
        \includegraphics[width=\textwidth]{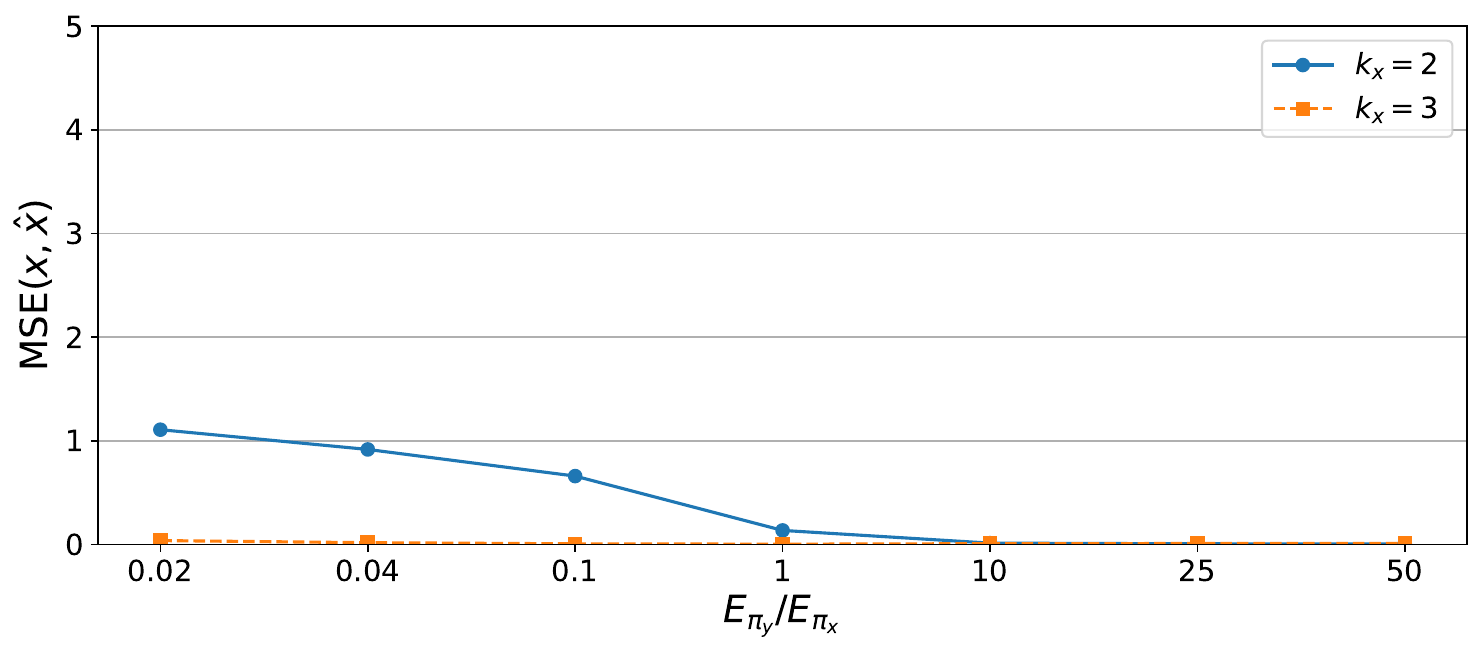}
        \caption{}
        \label{fig:mse_glv_glv}
    \end{subfigure}

    \caption{
    The mean squared error (MSE) between the true states $x$ and the inferred states $\hat{x}$,
    for a) scenario-different: Lorenz-GM vs.\ GLV-GP, and b) scenario-same: GLV-GM vs.\ GLV-GP,
    across $k_x=2$ (blue curve) and $k_x=3$ (orange curve) orders of motion and across different noise precision prior ratio.
    }
    \label{fig:x_xhat}
\end{figure}

The accuracy of the inference schemes, quantified by the mean squared error (MSE) between the true states and the hidden state estimates $\hat{\bfx}_t$, is illustrated in Fig.~\ref{fig:x_xhat}; the results are provided for two orders of motion and varying noise precision prior ratios \emph{C}.

We can see a consistent reduction in the MSE loss across both scenarios as we increase the orders of motion, highlighting the value of the generalised orders of motion. A GLV-GM in Fig.~\ref{fig:mse_glv_glv} is better suited to the data generated by the GLV-GP since they are both from the same model family, compared to the Lorenz-GM in Fig.~\ref{fig:mse_lorenz_glv}.

In order to see the actual inferred states trajectories---relative to the true trajectories---for both scenarios with $k_x=2$ and $k_x=3$ orders of motion, see Appendix.~\ref{app:x}. The main finding is that in the scenario-different setting, the Lorenz-GM struggles to track the GLV-GP when only $k_x=2$ orders of motion are employed. However, increasing the order of motion to $k_x=3$ leads to a substantial improvement in tracking performance. This result highlights the value of GCM, particularly when the structural assumptions of the GM differ significantly from those of the GP. In contrast, in the scenario-same setting the GLV-GM is able to track the GLV-GP effectively with both $k_x=2$ and $k_x=3$, indicating that when the dynamics of the GM closely match those of the GP, increasing the order of motion provides comparatively limited additional benefit.

It should be noted that the temporal derivatives of $\bfy_t$ are obtained through a numerical approximation (see Section.~\ref{sec:gcm}) which becomes increasingly unstable for higher-order terms.

\subsection{Free action under different orders of motion}
\begin{figure}[t]
    \centering
    \begin{subfigure}{\textwidth}
        \centering
        \includegraphics[width=\textwidth]{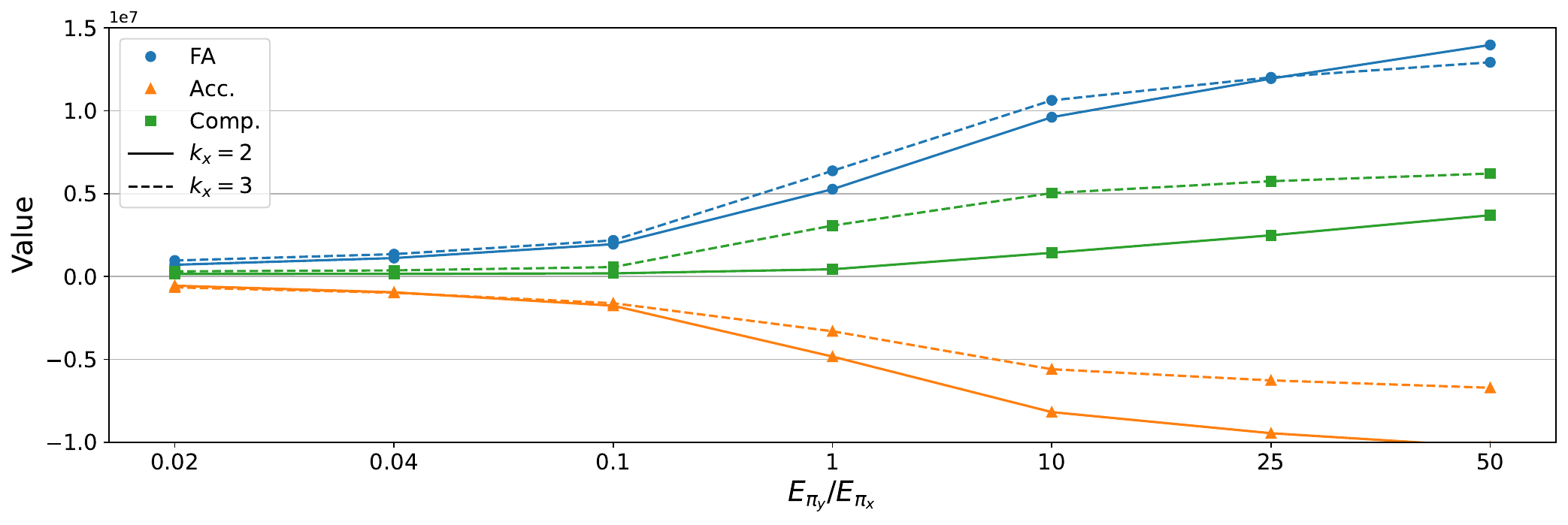}
        \caption{}
        \label{fig:fa_lorenz_glv}
    \end{subfigure}

    \vspace{0.1cm}

    \begin{subfigure}{\textwidth}
        \centering
        \includegraphics[width=\textwidth]{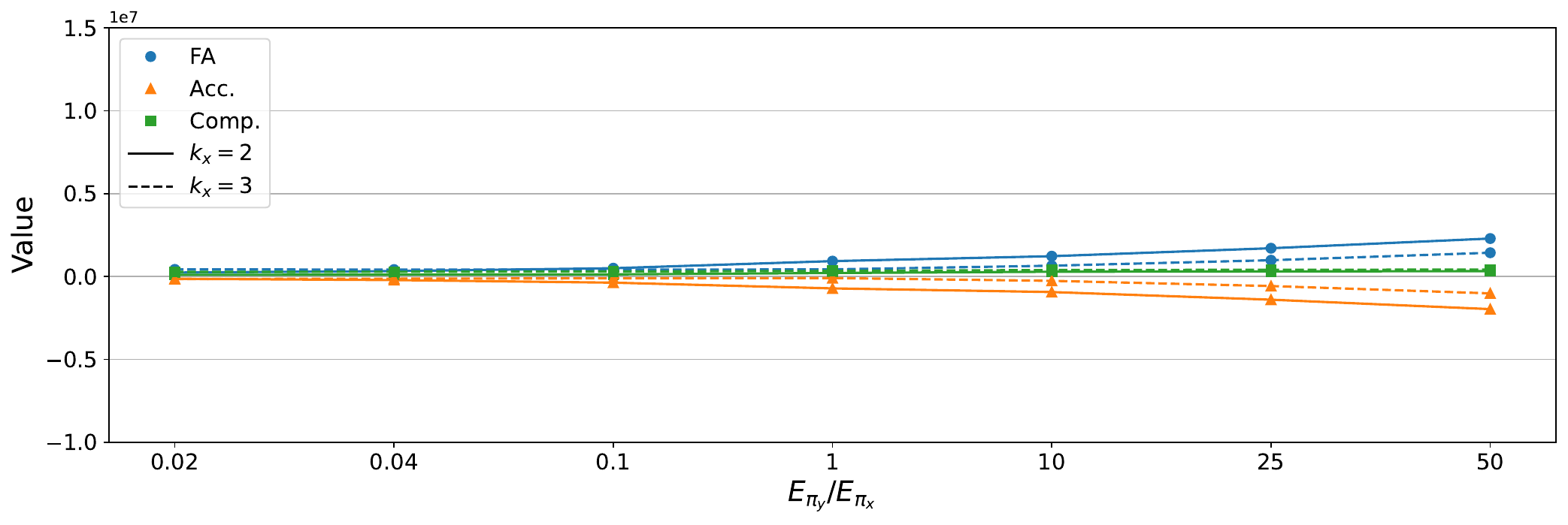}
        \caption{}
        \label{fig:fa_glv_glv}
    \end{subfigure}

    \caption{The blue curves show FA values, while the orange and green curves correspond to the accuracy and complexity values for a) scenario-different: Lorenz-GM and a GLV-GP, and b) scenario-same: GLV-GM and GLV-GP across two different orders of motion and varying noise precision prior ratios $\frac{E_{\Pi_y}}{E_{\Pi_x}}$. The solid and dashed markers correspond to $k_x=2$ and $k_x=3$ orders of motion. The horizontal axis shows the seven precision prior ratios.}
    \label{fig:fa}
\end{figure}

Fig.~\ref{fig:fa} shows the free action (FA) values for each precision prior ratio along with the corresponding constituent accuracy and complexity values, for the GM with the lowest FA across the entire experiment setting in Table.~\ref{tab:tuning_parameters}.

Fig.~\ref{fig:fa_lorenz_glv} corresponds to scenario-different, where we have Lorenz-GM fitted to a GLV-GP. It can be seen that the addition of orders of motion (i.e., from $k_x=2$ to $k_x=3$) does not change the FA much, however, looking at the constituent accuracy and complexity curves, it becomes clear that both accuracy and complexity terms increase with the orders of motion, which results in small changes in FA. This means that the increase in accuracy when increasing the order of motion comes with a complexity cost. In contrast, for scenario-same, where we have a GLV-GM, we can see in Fig.~\ref{fig:fa_glv_glv} that FA values are much lower compared to scenario-different, across all ratios, highlighting the fact that in the second scenario, a GLV-GM is consistently able to provide a simpler explanation that offers an accurate account of the observations, compared to a Lorenz-GM, under a GLV-GP. It is evident that higher order motion can improve the accuracy, but at the cost of an increase in complexity, which is why the FA values across different orders of motion remain almost identical.

\subsection{The generative power of the GM}
The online generalised PC model is a generative model, and as such, at any given time $t$, it can predict sensations through its observation model (i.e.,~the $\bfg(\,\cdot\,)$ function). More specifically, given the current inferred generalised state of the world, $\gcmu_x$, the GM can use the observation model to map $\gcmu_x$ into the expected generalised sensation, $\hat{\tilde{y}}$. In our generalised PC model, the observation model, is defined as an identity mapping, which means the inferred generalised states are actually the predicted generalised sensations $\hat{y}$. The plots of the GM-generated sensations against the ground truth for both scenarios---and for $k_x=2$ and $k_x=3$ orders of motion---are given in Appendix.~\ref{app:y}.

\section{Conclusion and future work} \label{sec:conclusion_and_futurework}
In this work, we introduced Online Dynamic Expectation Maximisation (ODEM), an online generalised predictive coding framework for simultaneous state inference, parameter learning, and uncertainty estimation in non-linear dynamical systems. Unlike offline approaches that revisit the full dataset iteratively, ODEM operates under strict temporal constraints, making it particularly relevant for biologically plausible inference and real-time adaptive systems. Within this formulation, top-down predictions recursively explain away bottom-up, precision-weighted prediction errors, thereby implementing a variational message-passing scheme consistent with the principles of predictive coding.

A key methodological feature of ODEM is the explicit separation of temporal scales. Fast Bayesian belief updating supports the continuous tracking of dynamic hidden states, while slower updates govern the adaptation of model parameters and precision estimates. This separation yields a principled variational framework for performing online triple estimation, in which inference, learning, and uncertainty estimation unfold concurrently but at different temporal scales.

Through numerical experiments involving non-linear and potentially chaotic dynamical systems, we demonstrated that ODEM can reliably track latent states even when there is a substantial mismatch between the functional form of the generative process and that of the candidate generative model. These results highlight the robustness of the approach and underscore the advantages of performing inference in GCM within a variational framework. Importantly, the scheme supports continuous adaptation of both model parameters and uncertainty estimates without requiring batch optimisation or retrospective smoothing.

Our results further demonstrate that ODEM provides an effective operationalisation of online generalised predictive coding. In particular, the scheme allows fast hidden states to be accurately tracked while parameter and uncertainty estimates progressively stabilise over time. More broadly, ODEM provides a bridge between predictive coding, variational inference, and online dynamical systems modelling, offering a scalable framework for studying adaptive inference in both biological and artificial agents.

A natural extension of ODEM is the incorporation of hierarchical generative models, allowing inference across multiple temporal and representational scales. Throughout, we assumed a smooth Gaussian process with a ``squared-exponential'' covariance kernel for the generative models considered. This was key in obtaining approximate marginal distribution for the generalised coordinate vectors. However, since the truncation of the GC order is already done, one could employ other kernels such as the Mat\'ern \citep[e.g.][]{WiAd2013}. We note that, whilst initially enticing, a version of ODEM using non-stationary kernels could prove challenging, particularly when deriving the approximate marginal distribution as in Section.~\ref{sec:gcm}.

So far, we have treated the ``smoothness'' matrix, $S_k(\sigma^2)$---whose elements correspond to covariances among all derivatives of the respective state or observation noise processes at stationarity, up to order $k$---as fixed. We will explore the effect of parametrising this matrix and making its elements learnable, which will directly influence the generalised precision terms $\Pi_\gcx$ and $\Pi_\gcy$. This can turn ODEM into a more robust scheme, which will have the flexibility to deal with observation time-series of varying degrees of smoothness.


We will also develop a ``hierarchical'' formulation of online generalised PC, in which the GM constitutes a deep architecture for modelling a real-world GP exhibiting a separation of temporal scales. The class of dynamics considered is assumed to belong to a universal analytic family that admits a description in terms of ``Renormalisation Group (RG)'' formalisms. Under this assumption, the GM is endowed with prior beliefs that the underlying data-generating dynamics can be systematically coarse-grained and characterised through RG transformations. This induces a deep hierarchical structure in the GM, wherein each level encodes dynamics operating at distinct temporal scales. Crucially, the separation of temporal scales is not implemented through explicit parameter updates governing slow processes. Instead, slower dynamics are represented implicitly as latent states at higher levels of the hierarchy. In this way, what would conventionally be treated as slowly evolving parameters are reinterpreted as hidden states at coarser temporal resolutions, enabling a unified state-based treatment of inference across scales.

Finally, dimensionality reduction and inflation techniques will be explored, so that ODEM can work with high-dimensional observations and track a high-dimensional GP using an underlying GM with lower dimensionality. Specifically, Physics-inspired course-graining techniques, such as renormalisation group used in creating Renormalising Generative Models (RGMs) \citep{friston2025pixels} will be explored. Beyond methodological development, future applications may explore whether ODEM can serve as a mechanistic model of online belief updating in biological systems, particularly in settings involving learning, attention, and adaptive behaviour.

\section*{Acknowledgments}
Mehran Hossein Zadeh Bazargani is supported under the European Union’s Horizon 2020 research and innovation programme under the Marie Skłodowska-Curie grant agreement No.~101034252. Adeel Razi, Mehran Hossein Zadeh Bazargani and Karl Friston are supported by the Australian Research Council (Refs: FT250100563 and DP260104251). Adeel Razi is also funded by the Australian National Health and Medical Research Council (Investigator Grant 1194910). Adeel Razi is a CIFAR Azrieli Global Scholar in the Brain, Mind \& Consciousness Program. Thomas Brendan Murphy is supported by funding from  Taighde \'{E}ireann -- Research Ireland grant (RI/12/RC/2289$\_$P2). Karl Friston is supported by funding from the Wellcome Trust (Ref: 226793/Z/22/Z). This research was supported by Monash eResearch capabilities, including HPC (M3 / MASSIVE).

\bibliographystyle{apalike}
\bibliography{references}

\clearpage
\FloatBarrier
\appendix

\section{Hierarchical generalised predictive coding}\label{app:Hierarchical generalised predictive coding}
A PC model can be modelled as a hierarchical state space model (SSM), where each of the $L$ layers of the hierarchy represents a level of abstraction:
\begin{align}
\begin{array}{cc}
    \tfrac{\dd}{\dd t}\bfx_t^1 &= \bff_1\left(\bfx_t^1, \bmtheta_t^1\right) + \bmomega_{x,1}(t) \\
                \bfy_t &= \bfg_1\left(\bfx_t^1, \bmtheta_t^1\right) + \bmomega_{y,1}(t) 
\end{array}~~~\text{and}~~~
\begin{array}{cc}
    \tfrac{\dd}{\dd t}\bfx_t^\ell &= \bff_\ell\left(\bfx_t^\ell, \bmtheta_t^\ell\right) + \bmomega_{x,\ell}(t), \\
                \bmtheta^{\ell-1}_t &= \bfg_\ell\left(\bfx_t^\ell, \bmtheta_t^\ell\right) + \bmomega_{\theta,\ell}(t), 
\end{array}\label{eq:ssm_hierarchical}
\end{align}

for $\ell=2,\ldots,L$; here, $\tfrac{\dd}{\dd t}\bfx_t^1$ is the first-order time derivative of the hidden state in layer $\ell$ at time $t$, and it represents the rate of change (i.e.,\ velocity) of the hidden state. Additionally, $\bmomega_{x,\ell}(t)$ and $\bmomega_{\theta,\ell}(t)$, are the random fluctuation processes in layer $\ell$, which if we assume to be zero-mean Gaussian processes, lead to approximately Gaussian conditional distributions: $\tfrac{\dd}{\dd t}\bfx_t^1 | (\bfx_t^1, \bmtheta_t^\ell)\sim \mathsf{N}(\bff_\ell(\bfx_t^\ell, \bmtheta_t^\ell), \Pi_{x^\ell}^{-1})$ and $\bmtheta^{\ell-1}_t| (\bfx_t^1, \bmtheta_t^\ell)\sim \mathsf{N}(\bfg_\ell(\bfx_t^\ell, \bmtheta_t^\ell), \Pi_{\theta^\ell}^{-1})$, where precision terms $\Pi_{x^\ell}$ and $\Pi_{\theta^\ell}$ are based on the assumed variance structure of the random fluctuations. Indeed, layer $\ell$ infers the most likely distribution over both the position and the velocity of the hidden states in layer $(\ell-1)$ by minimising VFE (in a purely local and Hebbian sense between layer $(\ell-1)$ and layer $\ell$). We can concatenate these position and velocity estimates as one variable $\gcx^{\ell}_t=	\{\bfx_t^\ell,\frac{\dd}{\dd t} \bfx_t^\ell\}$, which is called the \emph{generalised} hidden states at layer $\ell$. A generalised hidden states encapsulates not only the hidden state of interest but also the dynamics of the hidden state. In its simplest form, a generalised state consists of the position of the state and its velocity, denoting two orders of motion but as we will see in the next section, one can expand the GM to higher orders of motion. Thus, the main goal of layer $\ell$ is to infer $\gcx^\ell_t$ at layer $(\ell-1)$.
At the bottom of the hierarchy, the first layer plays the role of sensory epithelia, tasked with inferring the hidden states (and their dynamics) of the external world based on noisy sensory signals.  Crucially, the same VFE minimisation approach takes place between each pair of consecutive layers in a local fashion. As each pair of neighbouring layers minimises its own VFE independently, the entire hierarchical GM is effectively inverted, achieving hierarchical inference of the hidden states that cause sensory observations—this is essentially the process of perception through hierarchical generalised PC.
The communication between the layers relies on the parameters $\bmtheta$ and $\bfg(\,\cdot\,)$, which together enable inter-layer communication. To reduce notational burden, $\bmtheta$ represents the collection of separate parameters for each part of the model; the parameters are sometimes referred to as \textit{causes}\citep{adams2013computational}. This hierarchical message-passing scheme reflects the brain's ability to integrate and process information across different levels of abstraction.

\section{Generalised prediction errors}\label{app:errors_example}
The objective function in the ODEM scheme is the VFE under the Laplace approximation
\begin{align*}
    \vfe_L&=\tfrac{1}{2}\left(\gceps^\top\Pi_\gceps\gceps+\bmeps_\theta^\top\Pi_\theta\bmeps_\theta+\bmeps_\lambda^\top\Pi_\lambda\bmeps_\lambda -\log|\Pi_\gceps||\Pi_\theta||\Pi_\lambda||\Sigma_x||\Sigma_\theta||\Sigma_\lambda|+d_yk_y\log2\pi\right),
\end{align*}     
where $\gceps = \left(\gceps_y,\gceps_x\right)^\top = \left(\gcy - \tilde{\bfg}\left(\widetilde\bmmu_{x}, \bmmu_\theta\right),D\widetilde\bmmu_{x} - \tilde{\bff}\left(\widetilde\bmmu_{x}, \bmmu_\theta\right)\right)^\top$ is the vector of generalised prediction errors. As an example, the generalised prediction errors for $k_x=2$ orders of motion are
\begin{align*}
    \begin{bmatrix}
        \gceps_y\\\gceps_x
    \end{bmatrix}
    =
    \begin{bmatrix}
        \gcy - \tilde{\bfg}\left(\widetilde\bmmu_{x}, \bmmu_\theta\right)\\D\widetilde\bmmu_{x} - \tilde{\bff}\left(\widetilde\bmmu_{x}, \bmmu_\theta\right)
    \end{bmatrix}
    =
    \begin{bmatrix}
        \bfy-\bfg(\bmmu_x,\bmmu_\theta)\\
        \bmmu_{x'}-{\bff}\left(\bmmu_{x}, \bmmu_\theta\right)
    \end{bmatrix},
\end{align*}
and for $k_x=3$ orders of motion, they are
\begin{align*}
    \begin{bmatrix}
        \gceps_y\\\gceps_x
    \end{bmatrix}
    =
    \begin{bmatrix}
        \gcy - \tilde{\bfg}\left(\widetilde\bmmu_{x}, \bmmu_\theta\right)\\D\widetilde\bmmu_{x} - \tilde{\bff}\left(\widetilde\bmmu_{x}, \bmmu_\theta\right)
    \end{bmatrix}
    =
    \begin{bmatrix}
        \bfy-\bfg(\bmmu_x,\bmmu_\theta)\\
        \bfy' -\nabla\bfg(\bmmu_x,\bmmu_\theta)\bmmu_{x'}\\
        \bmmu_{x'}-{\bff}\left(\bmmu_{x}, \bmmu_\theta\right)\\
        \bmmu_{x''}-\nabla \bff \left(\bmmu_{x}, \bmmu_\theta\right)\bmmu_{x'}
    \end{bmatrix}.
\end{align*}

\section{GLV-GP}\label{app:glv_gp}
We have chosen \(A\) to be \emph{anti-symmetric}, i.e.,\ \(A = -A^\top\), and set the growth vector to \(\boldsymbol{r} = (0,0,0)^\top\), simplifying the GLV-GP process to $\dot{\boldsymbol{x}} = \boldsymbol{x} \odot A\boldsymbol{x}$. Specifically, the $A$ matrix in the GLV-GP is selected to be:
\[
A = 
\begin{bmatrix}
0.0 & 0.2 & -0.4 \\
-0.2 & 0.0 & 0.1 \\
0.4 & -0.1 & 0.0
\end{bmatrix}.
\]

Substituting these into the standard GLV formulation yields the final GLV-GP model following system of coupled differential equations:
\[
\begin{aligned}
\dot{x}_0 &= x_0 \left( 0.2\,x_1 - 0.4\,x_2 \right), \\
\dot{x}_1 &= x_1 \left( -0.2\,x_0 + 0.1\,x_2 \right), \\
\dot{x}_2 &= x_2 \left( 0.4\,x_0 - 0.1\,x_1 \right).
\end{aligned}
\]

\section{\texorpdfstring{Inferred states $\mu_x$}{Inferred states mu x}}
\label{app:x}

The following demonstrates the inferred states for $k_x=2$ (Appendix.~\ref{app:x_kx=2}), and $k_x=3$ (Appendix.~\ref{app:x_kx=3}) orders of motion.

\subsection{\texorpdfstring{$k_x = 2$  }{k x = 2}}
\label{app:x_kx=2}

In Fig.~\ref{fig:xhat_kx2_all_ratios}, for scenario-different (left panel), it seems that 2 orders of motion (i.e.,~ $k_x=2$) are not sufficient for a Lorenz-GM to track the GLV-GP; nevertheless, the tracking improves as the precision prior ratio gets larger (Fig.~\ref{fig:x_scenario=1_kx=2_ratio=0.02} to {Fig.~\ref{fig:x_scenario=1_kx=2_ratio=50}}). On the other hand, in scenario-same (right panel), 2 orders of motion seem sufficient to enable the GLV-GM to track GLV-GP reasonably well (albeit with a small lag), as seen in Fig.~\ref{fig:x_scenario=2_kx=2_ratio=0.02}. As we move from the lowest precision prior to the highest, the tracking improves (Fig.~\ref{fig:x_scenario=2_kx=2_ratio=0.02} to {Fig.~\ref{fig:x_scenario=2_kx=2_ratio=50}}). This is due the fact that if the underlying assumptions of the state dynamics in the GM match that of the GP, then the GM can track the GP nearly effortlessly and would not even require higher orders of motion nor would it need a careful precision prior tuning.

\begin{figure}[H]
    \centering
    \begin{subfigure}[t]{0.48\textwidth}
        \centering
        \includegraphics[width=\linewidth]{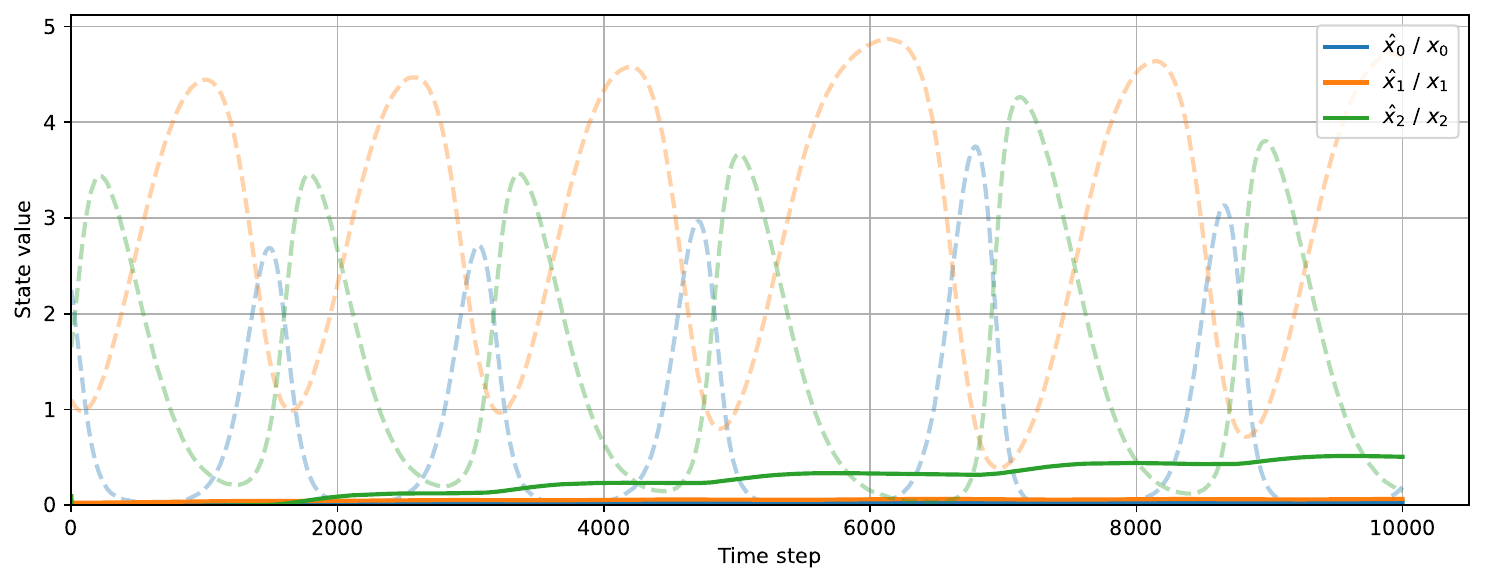}
        \caption{$\frac{E_{\Pi_y}}{E_{\Pi_x}}=0.02$}
        \label{fig:x_scenario=1_kx=2_ratio=0.02}
    \end{subfigure}\hfill
    \begin{subfigure}[t]{0.48\textwidth}
        \centering
        \includegraphics[width=\linewidth]{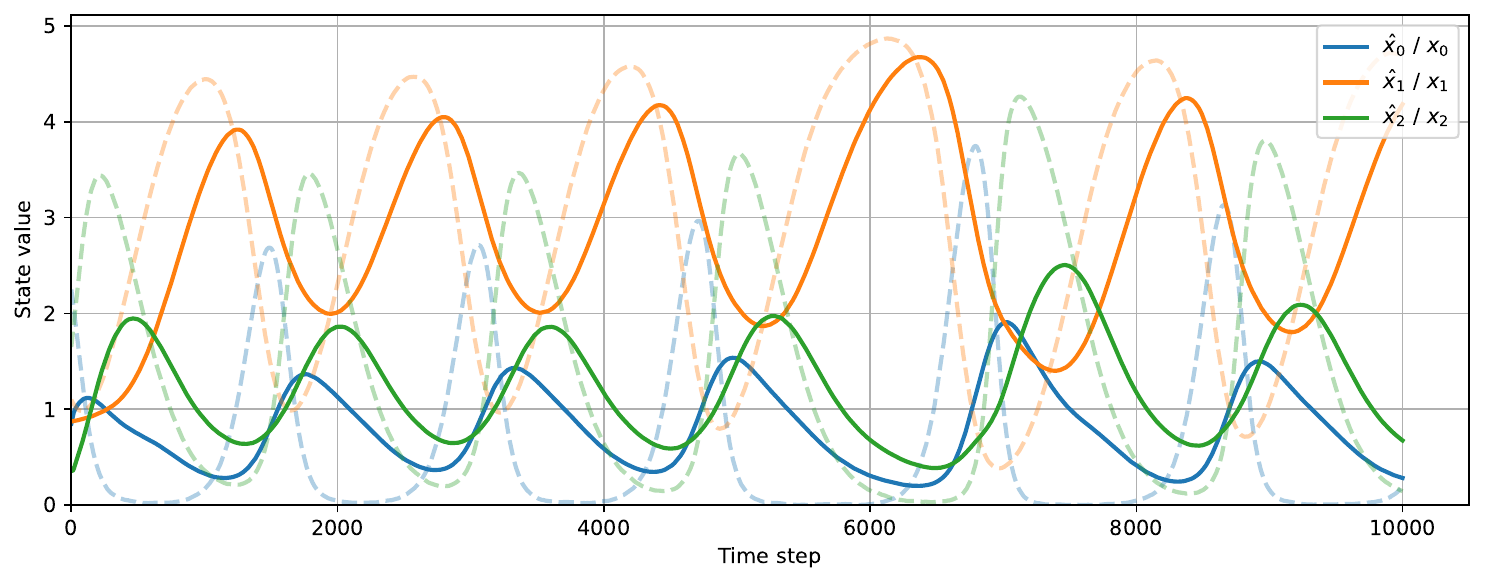}
        \caption{$\frac{E_{\Pi_y}}{E_{\Pi_x}}=0.02$}
        \label{fig:x_scenario=2_kx=2_ratio=0.02}
    \end{subfigure}

    \vspace{0.3cm}

    \begin{subfigure}[t]{0.48\textwidth}
        \centering
        \includegraphics[width=\linewidth]{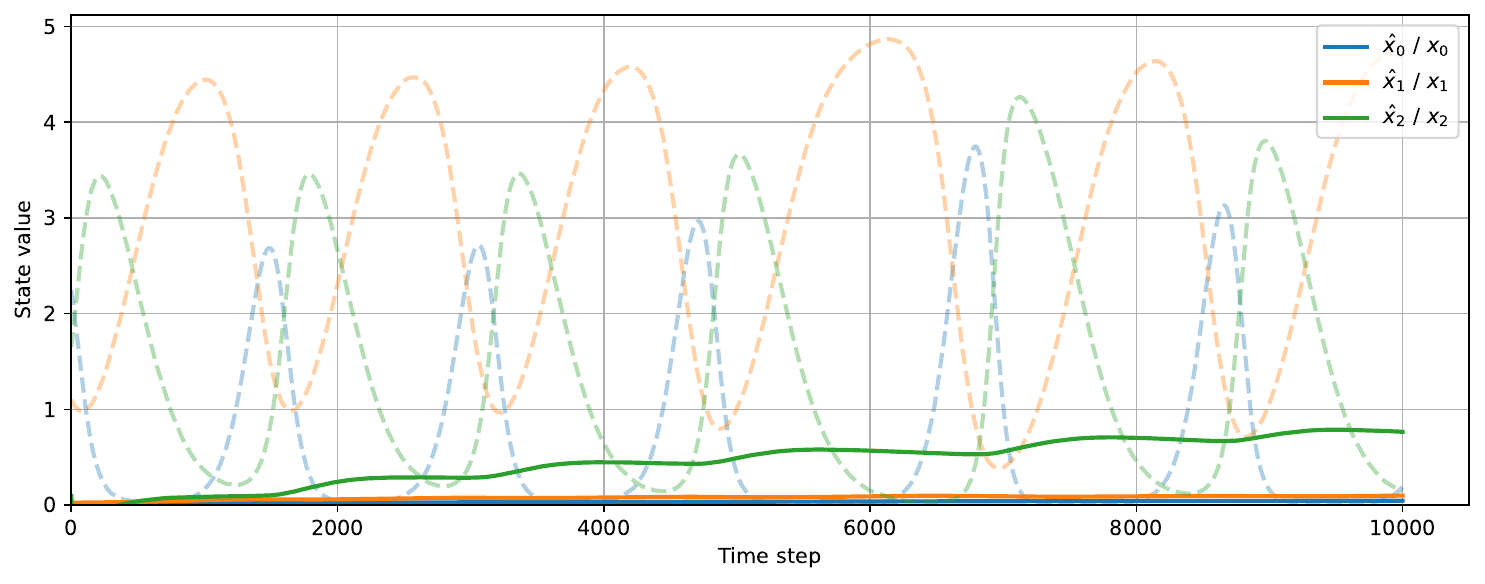}
        \caption{$\frac{E_{\Pi_y}}{E_{\Pi_x}}=0.04$}
        \label{fig:x_scenario=1_kx=2_ratio=0.04}
    \end{subfigure}\hfill
    \begin{subfigure}[t]{0.48\textwidth}
        \centering
        \includegraphics[width=\linewidth]{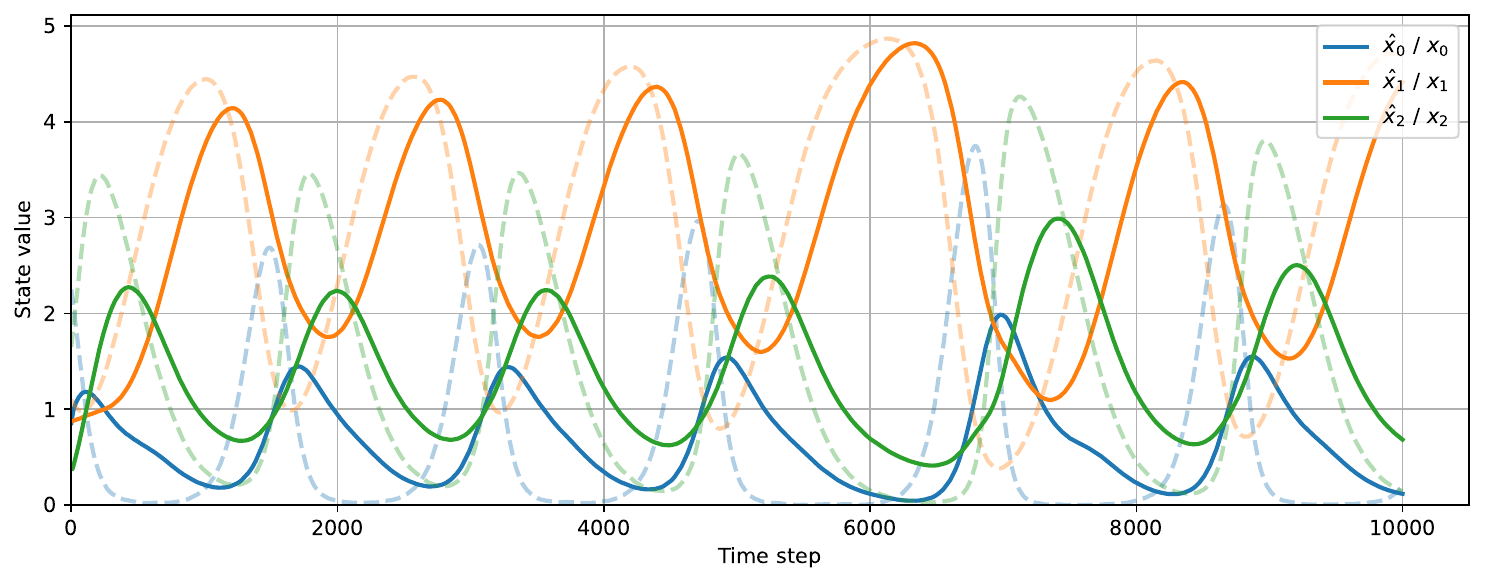}
        \caption{$\frac{E_{\Pi_y}}{E_{\Pi_x}}=0.04$}
        \label{fig:x_scenario=2_kx=2_ratio=0.04}
    \end{subfigure}

    \vspace{0.3cm}

    \begin{subfigure}[t]{0.48\textwidth}
        \centering
        \includegraphics[width=\linewidth]{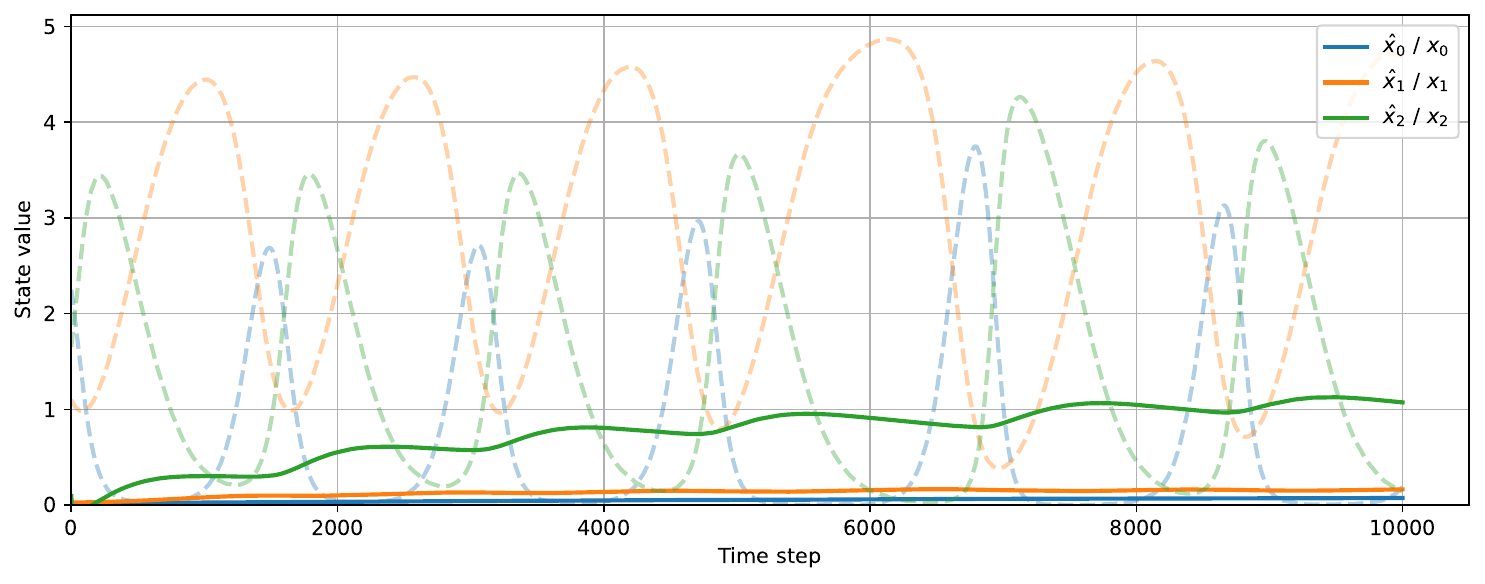}
        \caption{$\frac{E_{\Pi_y}}{E_{\Pi_x}}=0.1$}
        \label{fig:x_scenario=1_kx=2_ratio=0.1}
    \end{subfigure}\hfill
    \begin{subfigure}[t]{0.48\textwidth}
        \centering
        \includegraphics[width=\linewidth]{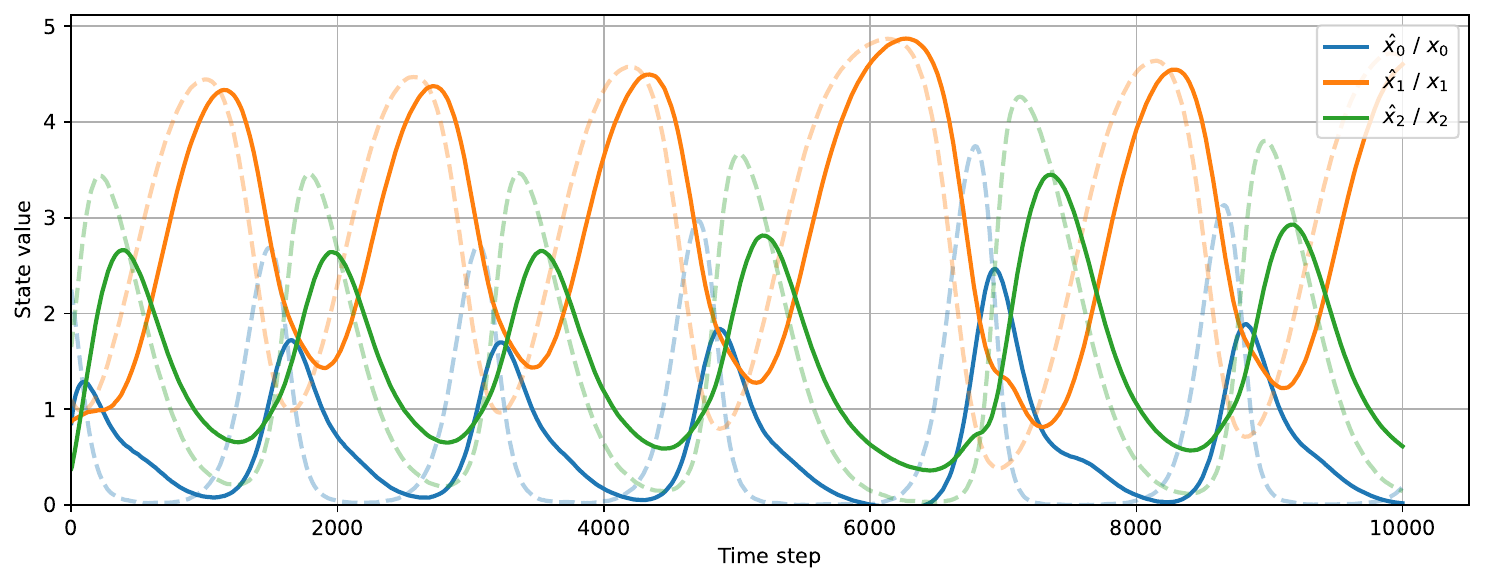}
        \caption{$\frac{E_{\Pi_y}}{E_{\Pi_x}}=0.1$}
        \label{fig:x_scenario=2_kx=2_ratio=0.1}
    \end{subfigure}

    \vspace{0.3cm}

    \begin{subfigure}[t]{0.48\textwidth}
        \centering
        \includegraphics[width=\linewidth]{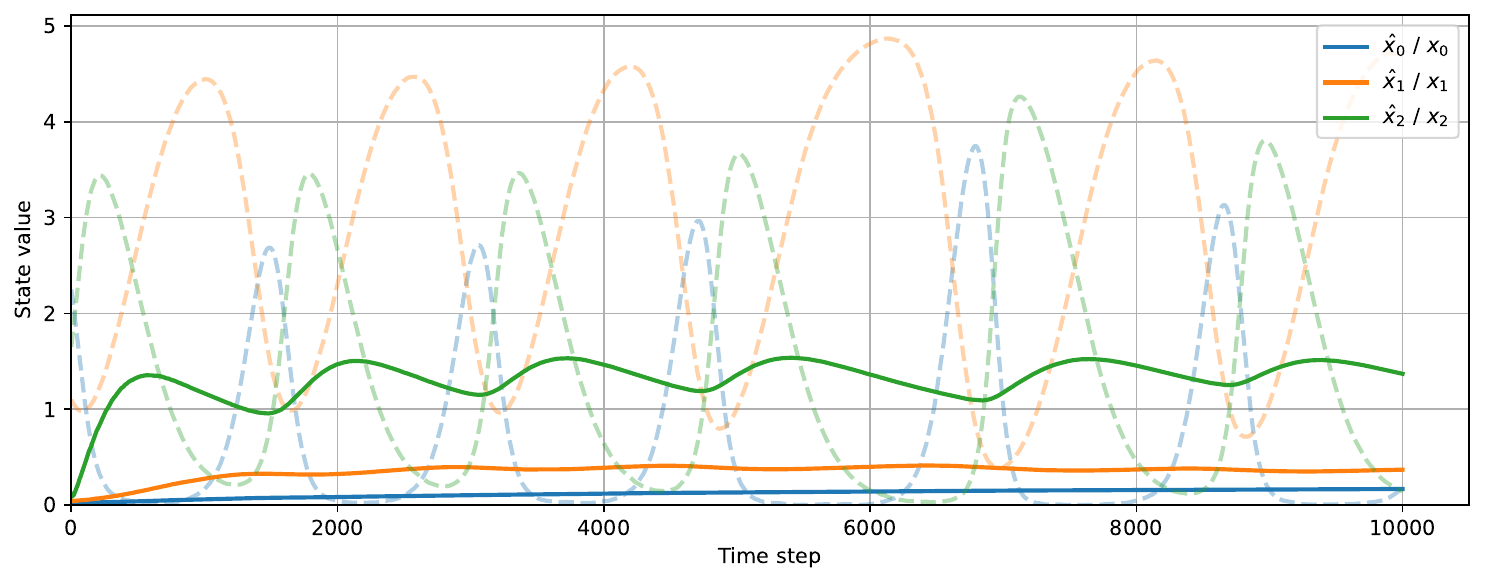}
        \caption{$\frac{E_{\Pi_y}}{E_{\Pi_x}}=1$}
        \label{fig:x_scenario=1_kx=2_ratio=1}
    \end{subfigure}\hfill
    \begin{subfigure}[t]{0.48\textwidth}
        \centering
        \includegraphics[width=\linewidth]{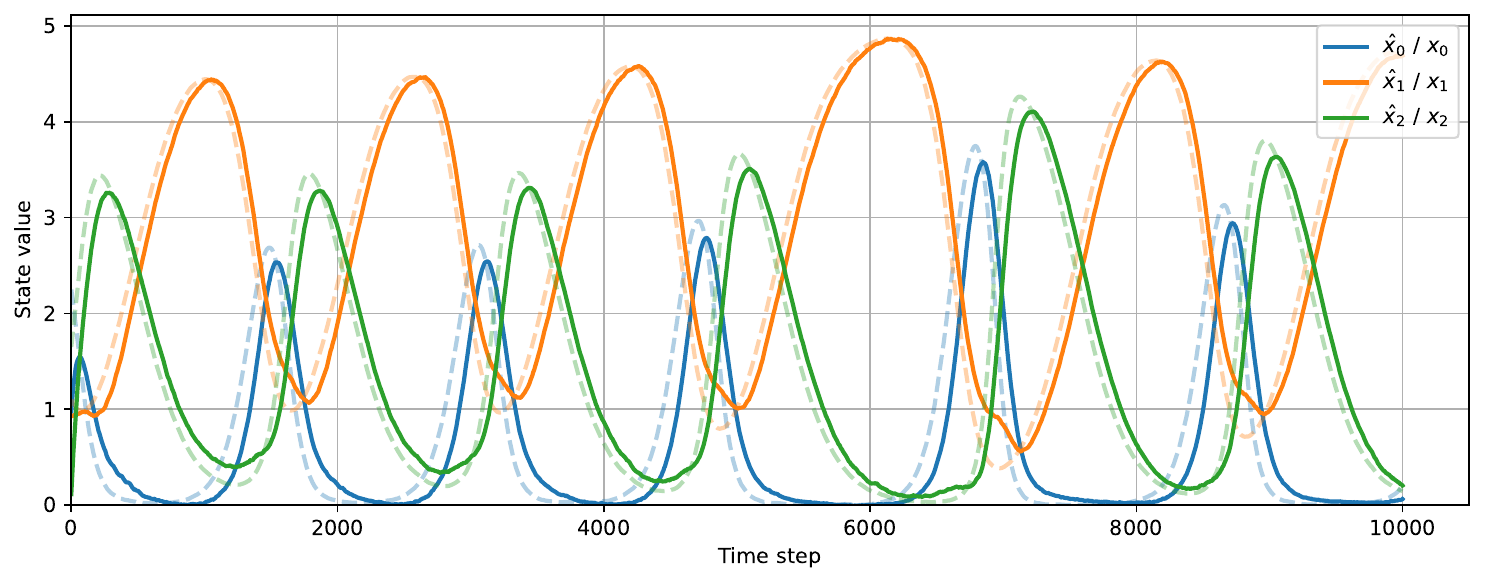}
        \caption{$\frac{E_{\Pi_y}}{E_{\Pi_x}}=1$}
        \label{fig:x_scenario=2_kx=2_ratio=1}
    \end{subfigure}
\end{figure}

\begin{figure}[H]\ContinuedFloat
    \centering

    \begin{subfigure}[t]{0.48\textwidth}
        \centering
        \includegraphics[width=\linewidth]{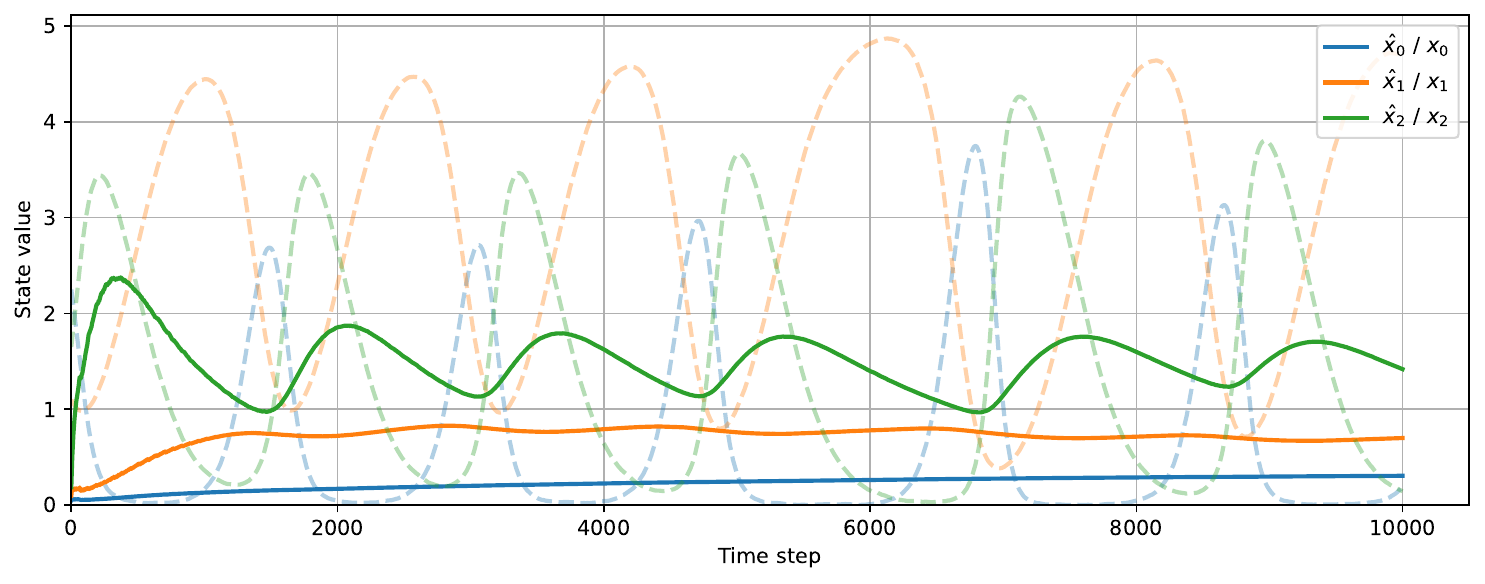}
        \caption{$\frac{E_{\Pi_y}}{E_{\Pi_x}}=10$}
        \label{fig:x_scenario=1_kx=2_ratio=10}
    \end{subfigure}\hfill
    \begin{subfigure}[t]{0.48\textwidth}
        \centering
        \includegraphics[width=\linewidth]{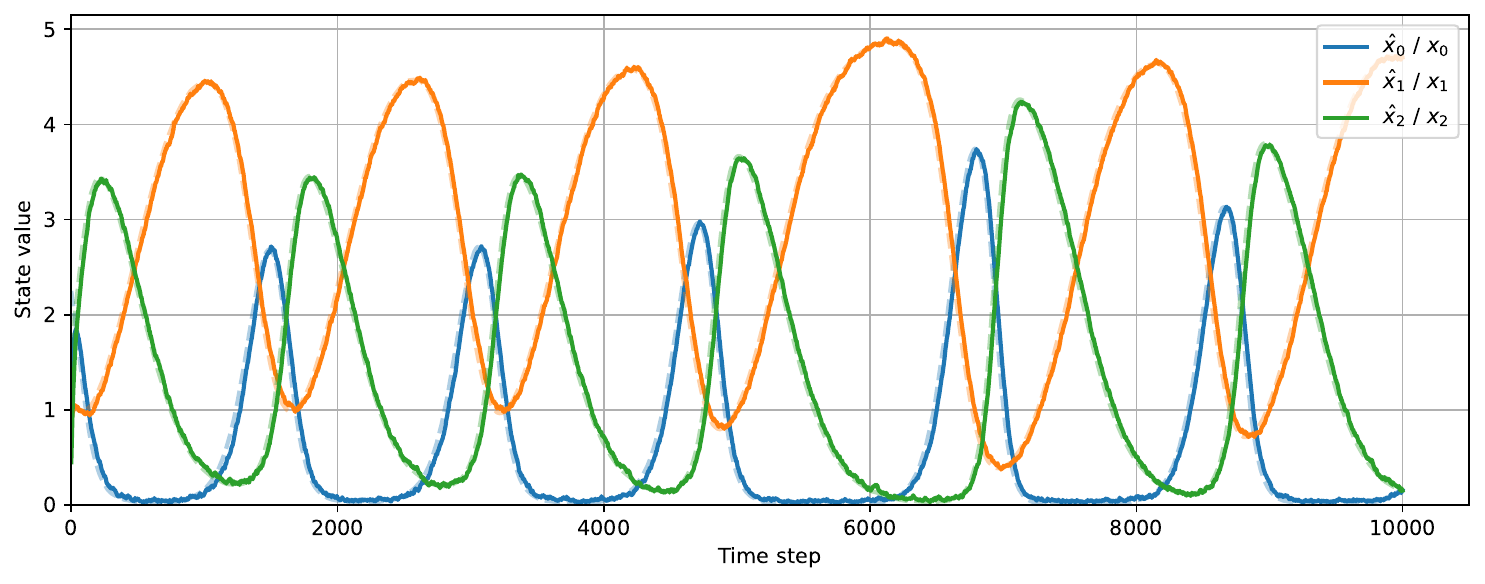}
        \caption{$\frac{E_{\Pi_y}}{E_{\Pi_x}}=10$}
        \label{fig:x_scenario=2_kx=2_ratio=10}
    \end{subfigure}

    \vspace{0.3cm}

    \begin{subfigure}[t]{0.48\textwidth}
        \centering
        \includegraphics[width=\linewidth]{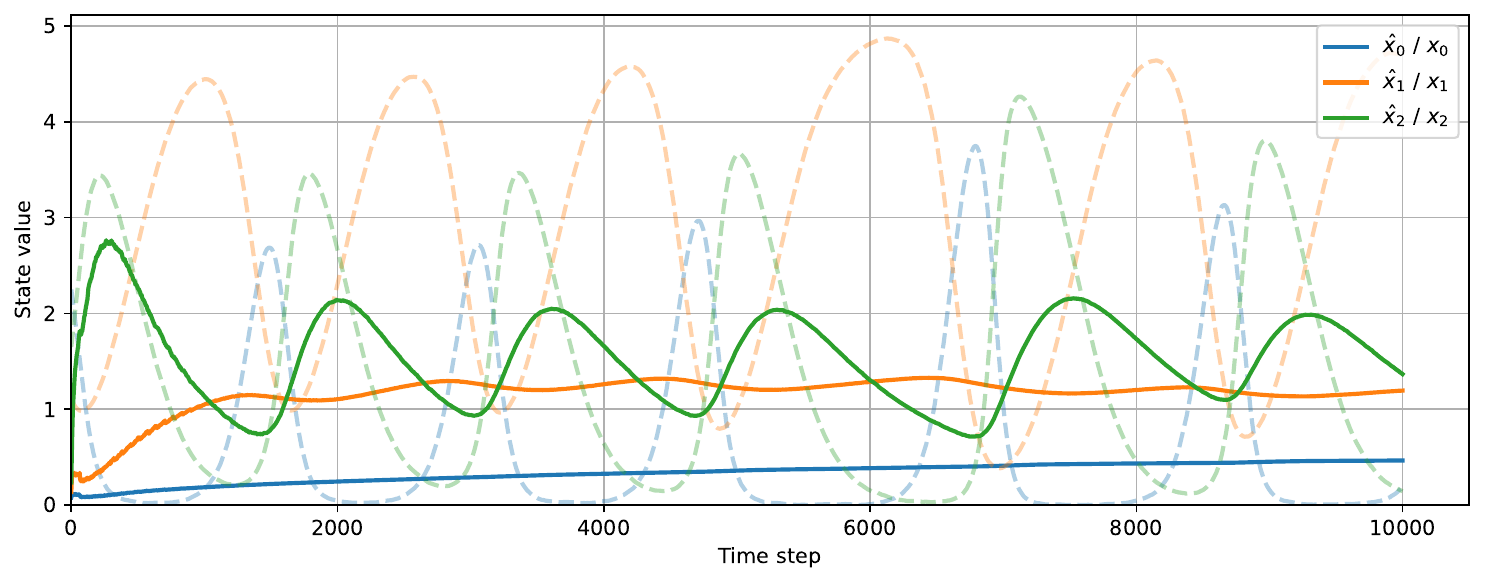}
        \caption{$\frac{E_{\Pi_y}}{E_{\Pi_x}}=25$}
        \label{fig:x_scenario=1_kx=2_ratio=25}
    \end{subfigure}\hfill
    \begin{subfigure}[t]{0.48\textwidth}
        \centering
        \includegraphics[width=\linewidth]{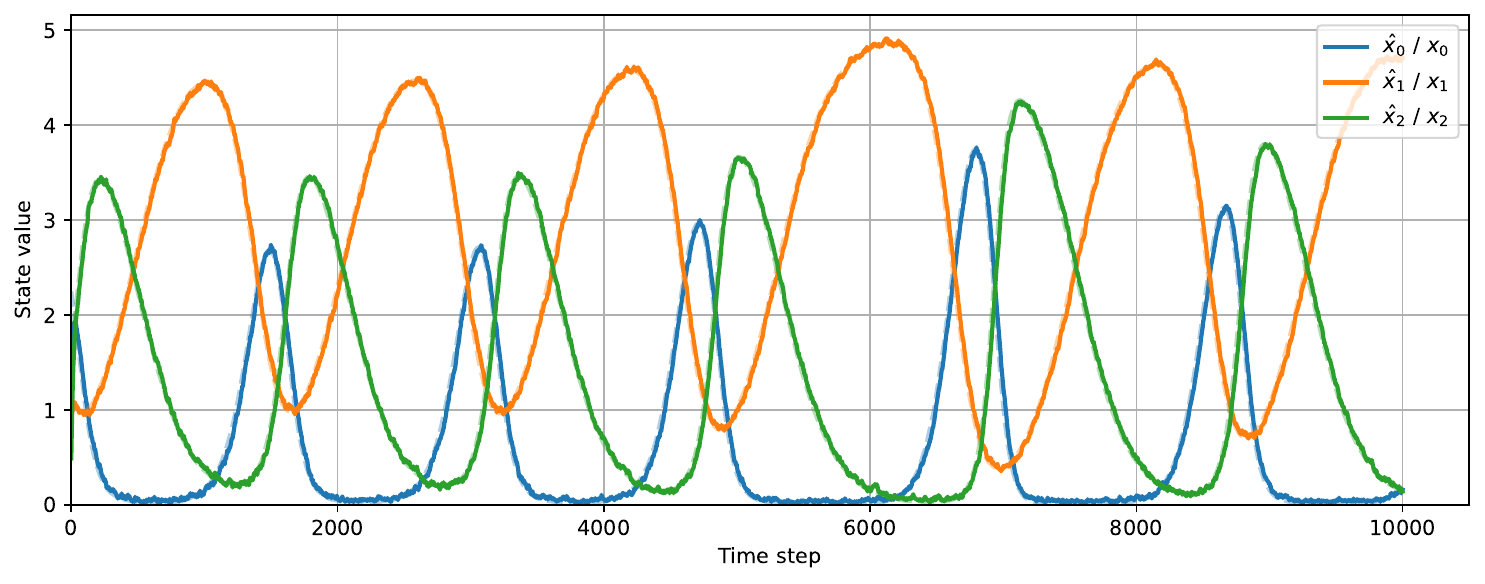}
        \caption{$\frac{E_{\Pi_y}}{E_{\Pi_x}}=25$}
        \label{fig:x_scenario=2_kx=2_ratio=25}
    \end{subfigure}

    \vspace{0.3cm}

    \begin{subfigure}[t]{0.48\textwidth}
        \centering
        \includegraphics[width=\linewidth]{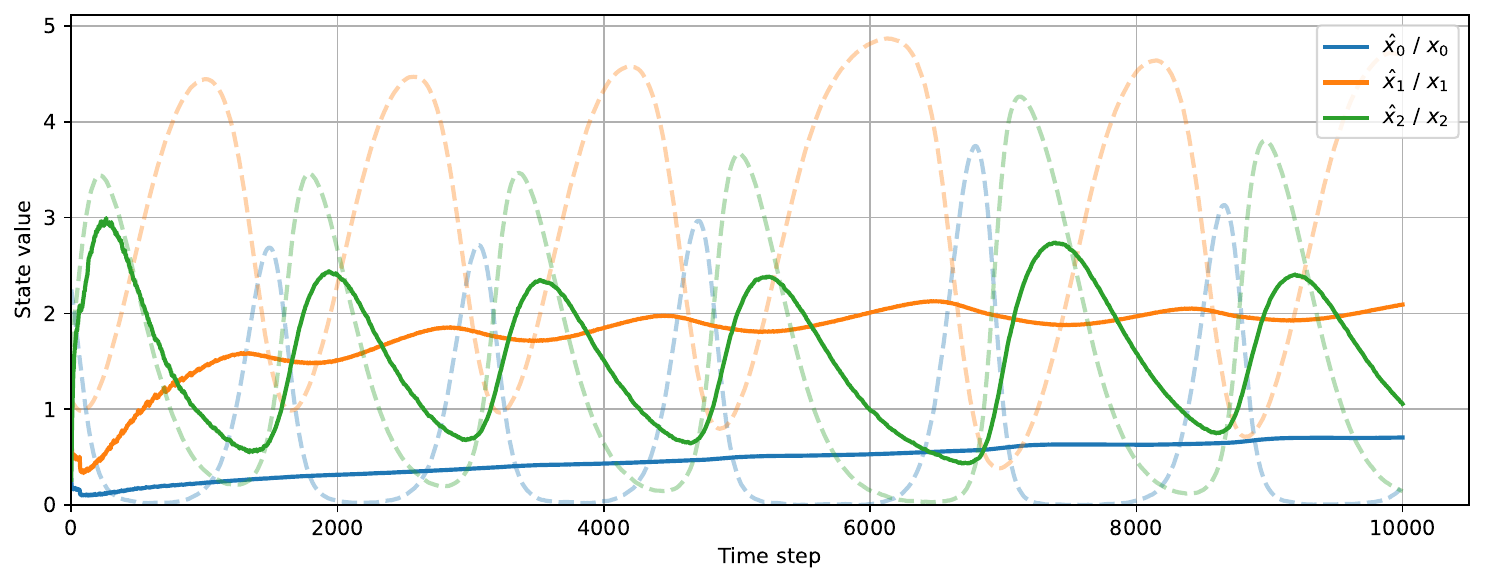}
        \caption{$\frac{E_{\Pi_y}}{E_{\Pi_x}}=50$}
        \label{fig:x_scenario=1_kx=2_ratio=50}
    \end{subfigure}\hfill
    \begin{subfigure}[t]{0.48\textwidth}
        \centering
        \includegraphics[width=\linewidth]{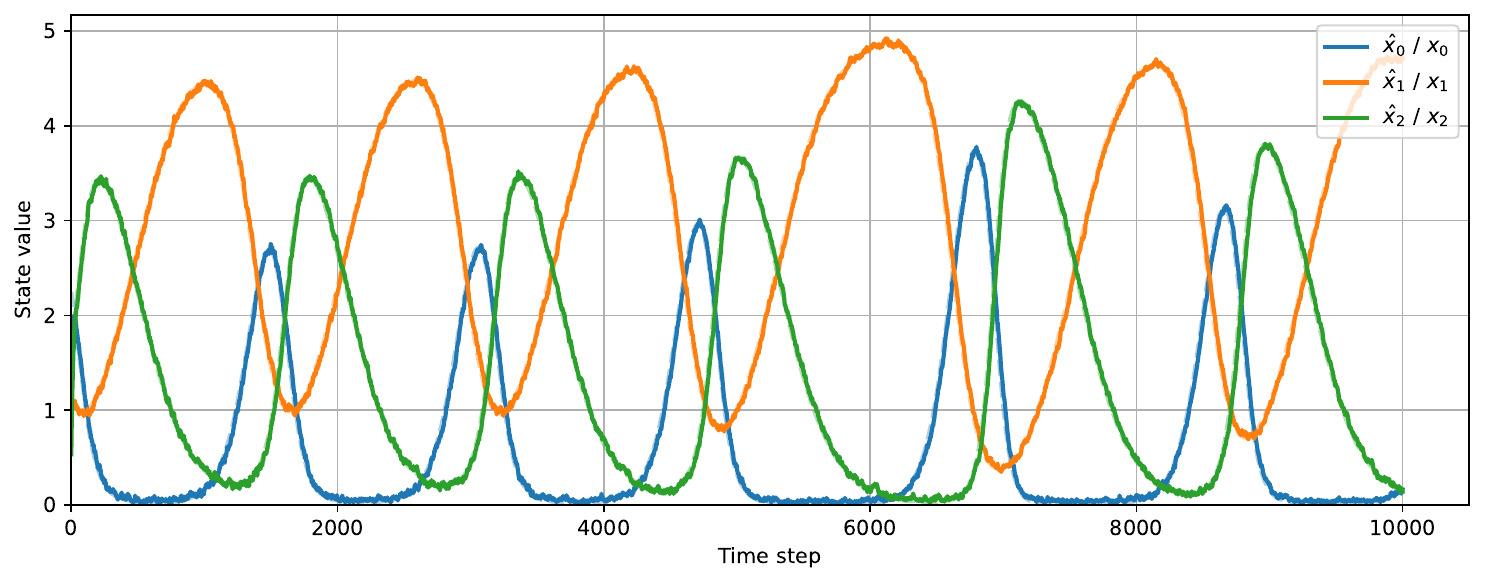}
        \caption{$\frac{E_{\Pi_y}}{E_{\Pi_x}}=50$}
        \label{fig:x_scenario=2_kx=2_ratio=50}
    \end{subfigure}

    \caption[]{ The inferred states for $k_x=2$ orders of motion. The left panel is for scenario-different: Lorenz-GM vs. GLV-GP, and the right panel is for scenario-same: GLV-GM vs. GLV-GP. Each row corresponds to a different precision prior ratio.}
    \label{fig:xhat_kx2_all_ratios}

\end{figure}

\subsection{\texorpdfstring{$k_x = 3$}{k x = 3}}
\label{app:x_kx=3}

In Fig.~\ref{fig:xhat_kx3_all_ratios}, for scenario-different (left panel) there is already an improvement in the performance of the Lorenz-GM under 3 orders of motion. Specifically, when $\frac{E_{\Pi_y}}{E_{\Pi_x}}=10$ in Fig.~\ref{fig:x_scenario=1_kx=3_ratio=10}, we can see a decent tracking of the GLV-GP, where the inferred states are not too noisy. As we increase the prior precision ratio, the inferred states become noisier, since the Lorenz-GM is attending more towards prediction error minimisation than tracking the dynamics (Fig.~\ref{fig:x_scenario=1_kx=3_ratio=10} to {Fig.~\ref{fig:x_scenario=1_kx=3_ratio=50}}). In Fig.~\ref{fig:xhat_kx3_all_ratios}, for scenario-same in (right panel), it seems that having 3 orders of motion has already enabled the GLV-GM to track the GLV-GP accurately, even under the lowest precision prior ratio. For instance, the lag we have previously observed in Fig.~\ref{fig:x_scenario=2_kx=2_ratio=0.02} is no longer happening in Fig.~\ref{fig:x_scenario=2_kx=3_ratio=0.02}, which highlights the value of higher orders of motion. Interestingly, as we increase the prior precision ratio beyond 1.0 (Fig.~\ref{fig:x_scenario=2_kx=3_ratio=1}), 
the inferred states become visibly noisier (i.e., less filtering) to the point that the dynamics are completely ignored in Fig.~\ref{fig:x_scenario=2_kx=3_ratio=50}. It is crucial to always experiment with different orders of motion and different precision prior ratios to find the GM with the best performance, that is, the GM with the lowest FA.

\begin{figure}[H]
    \centering

    \begin{subfigure}[t]{0.48\textwidth}
        \centering
        \includegraphics[width=\linewidth]{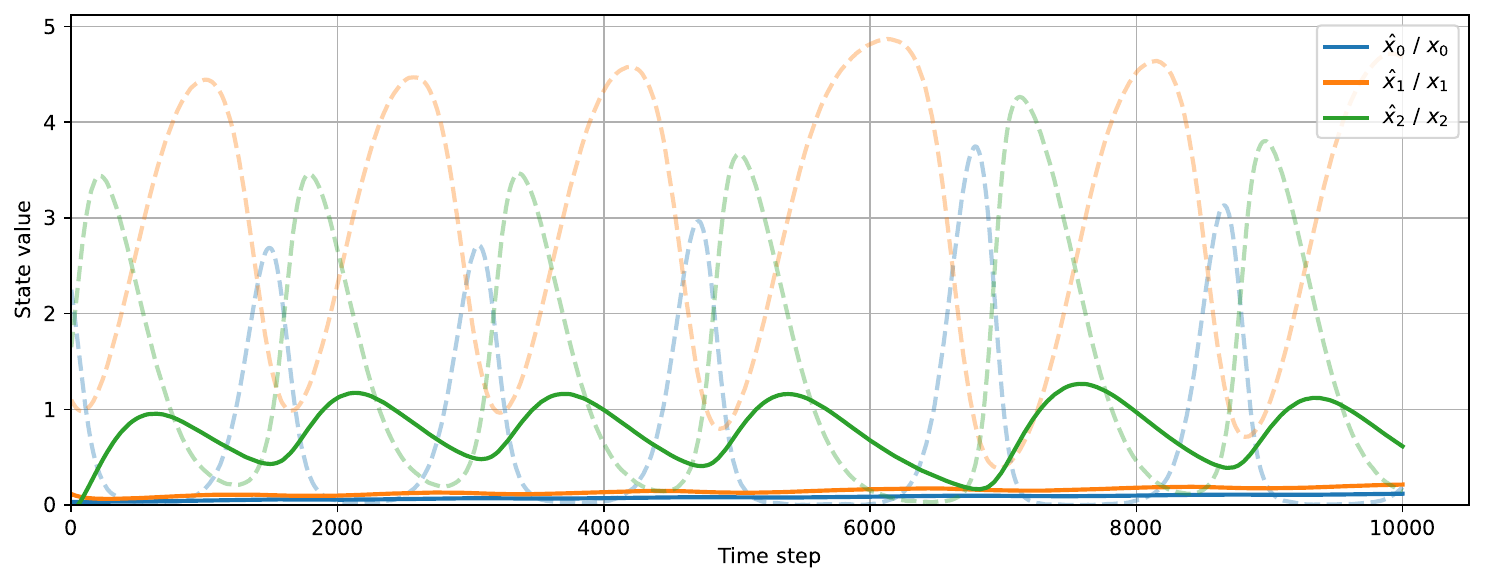}
        \caption{$\frac{E_{\Pi_y}}{E_{\Pi_x}}= 0.02$}
        \label{fig:x_scenario=1_kx=3_ratio=0.02}
    \end{subfigure}\hfill
    \begin{subfigure}[t]{0.48\textwidth}
        \centering
        \includegraphics[width=\linewidth]{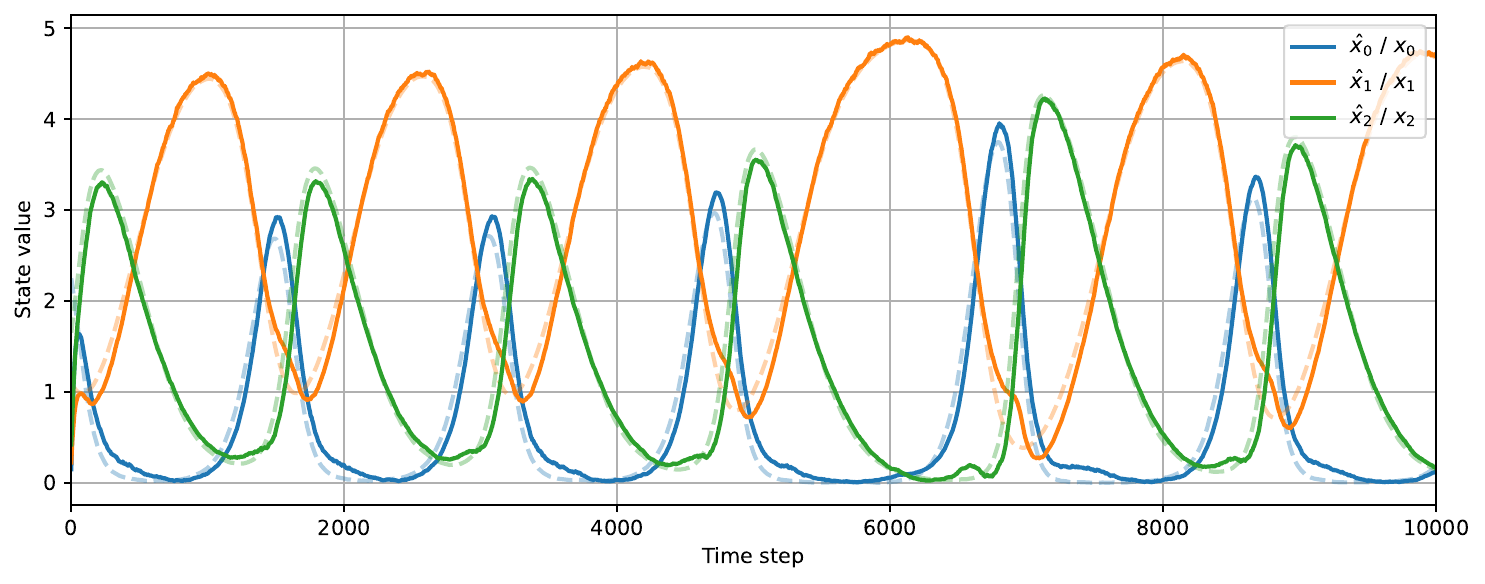}
        \caption{$\frac{E_{\Pi_y}}{E_{\Pi_x}}=0.02$}
        \label{fig:x_scenario=2_kx=3_ratio=0.02}
    \end{subfigure}

    \vspace{0.3cm}

    \begin{subfigure}[t]{0.48\textwidth}
        \centering
        \includegraphics[width=\linewidth]{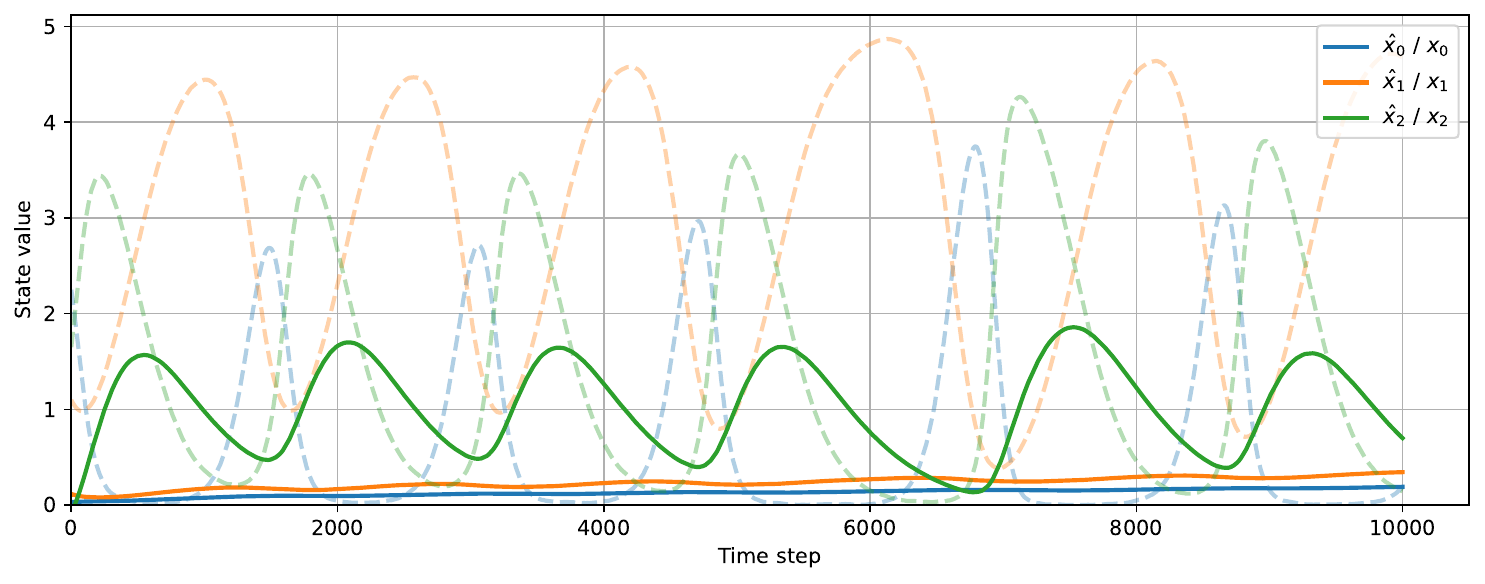}
        \caption{$\frac{E_{\Pi_y}}{E_{\Pi_x}}=0.04$}
        \label{fig:x_scenario=1_kx=3_ratio=0.04}
    \end{subfigure}\hfill
    \begin{subfigure}[t]{0.48\textwidth}
        \centering
        \includegraphics[width=\linewidth]{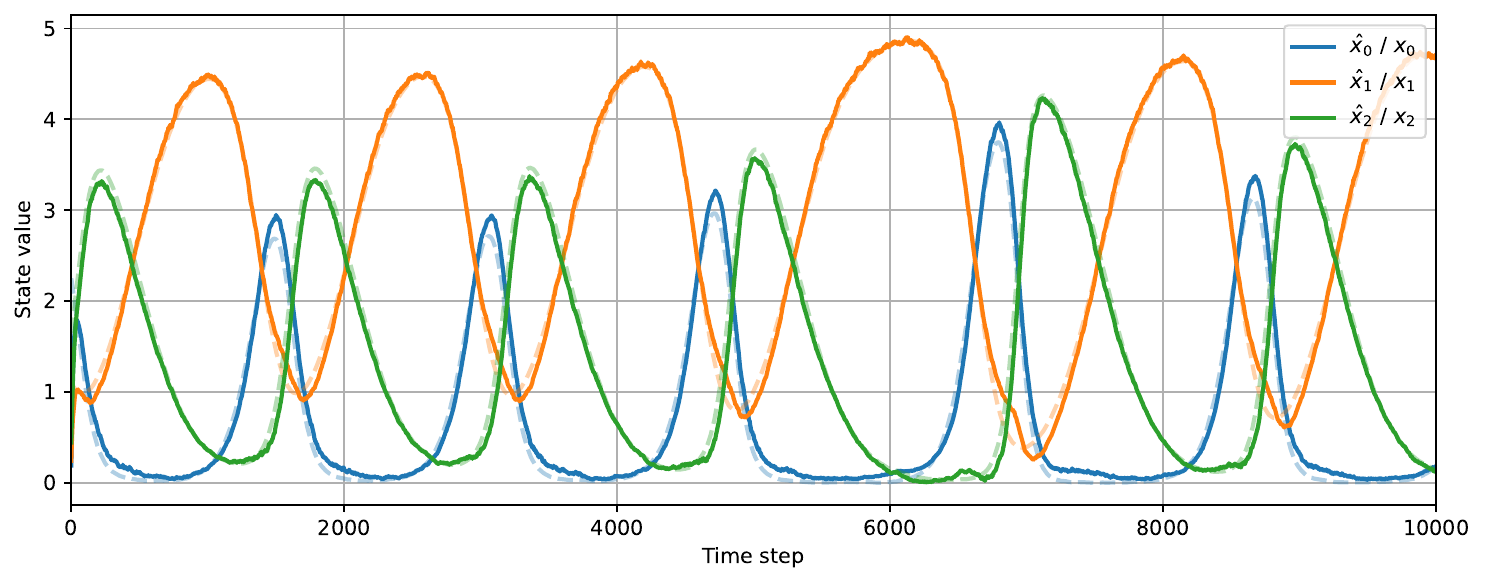}
        \caption{$\frac{E_{\Pi_y}}{E_{\Pi_x}}=0.04$}
        \label{fig:x_scenario=2_kx=3_ratio=0.04}
    \end{subfigure}

    \vspace{0.3cm}

    \begin{subfigure}[t]{0.48\textwidth}
        \centering
        \includegraphics[width=\linewidth]{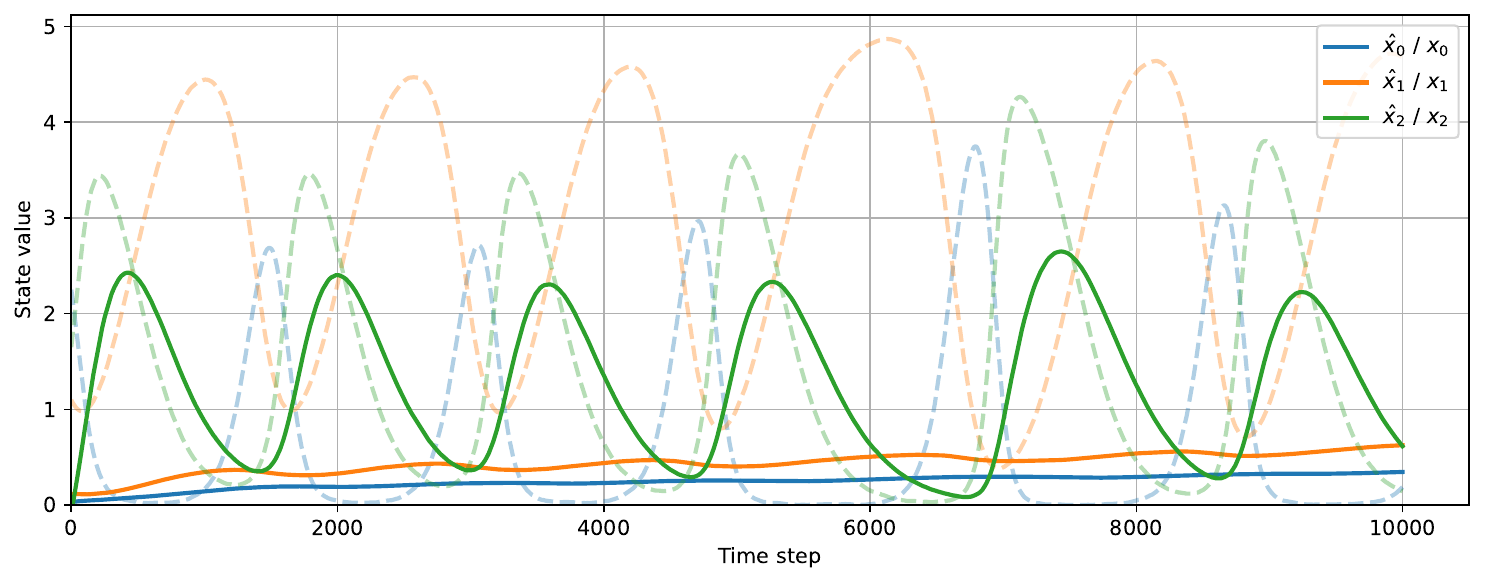}
        \caption{$\frac{E_{\Pi_y}}{E_{\Pi_x}}=0.1$}
        \label{fig:x_scenario=1_kx=3_ratio=0.1}
    \end{subfigure}\hfill
    \begin{subfigure}[t]{0.48\textwidth}
        \centering
        \includegraphics[width=\linewidth]{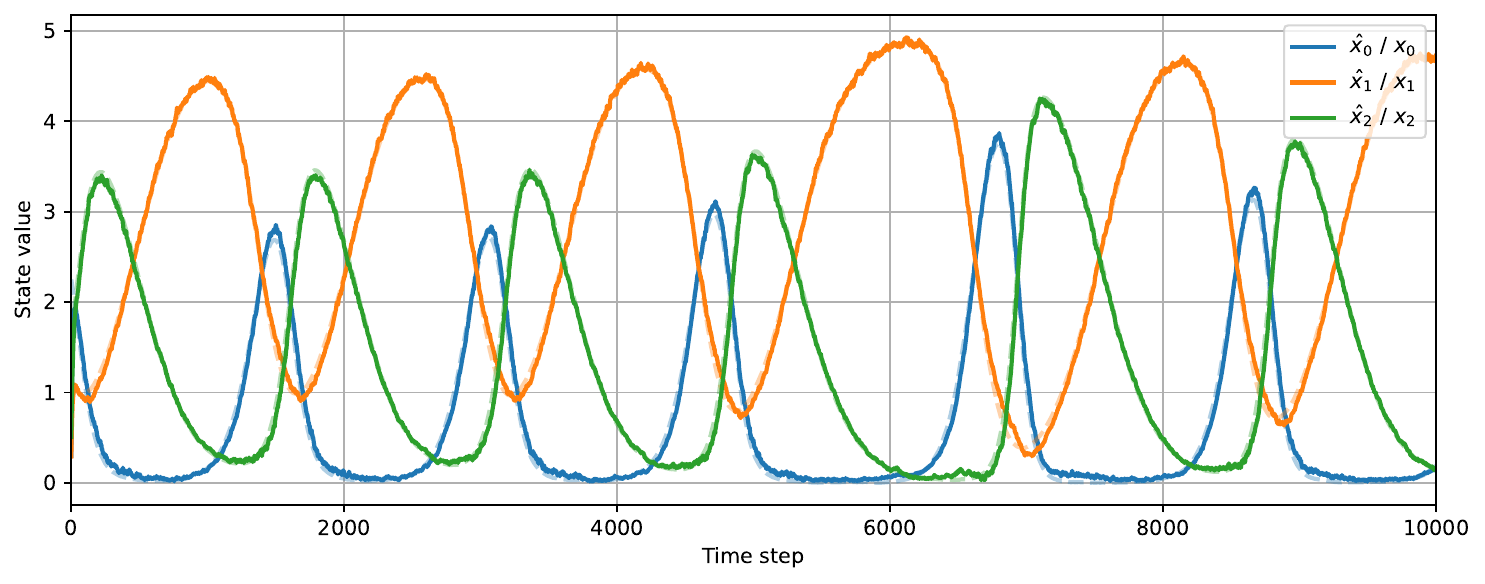}
        \caption{$\frac{E_{\Pi_y}}{E_{\Pi_x}}=0.1$}
        \label{fig:x_scenario=2_kx=3_ratio=0.1}
    \end{subfigure}

    \vspace{0.3cm}

    \begin{subfigure}[t]{0.48\textwidth}
        \centering
        \includegraphics[width=\linewidth]{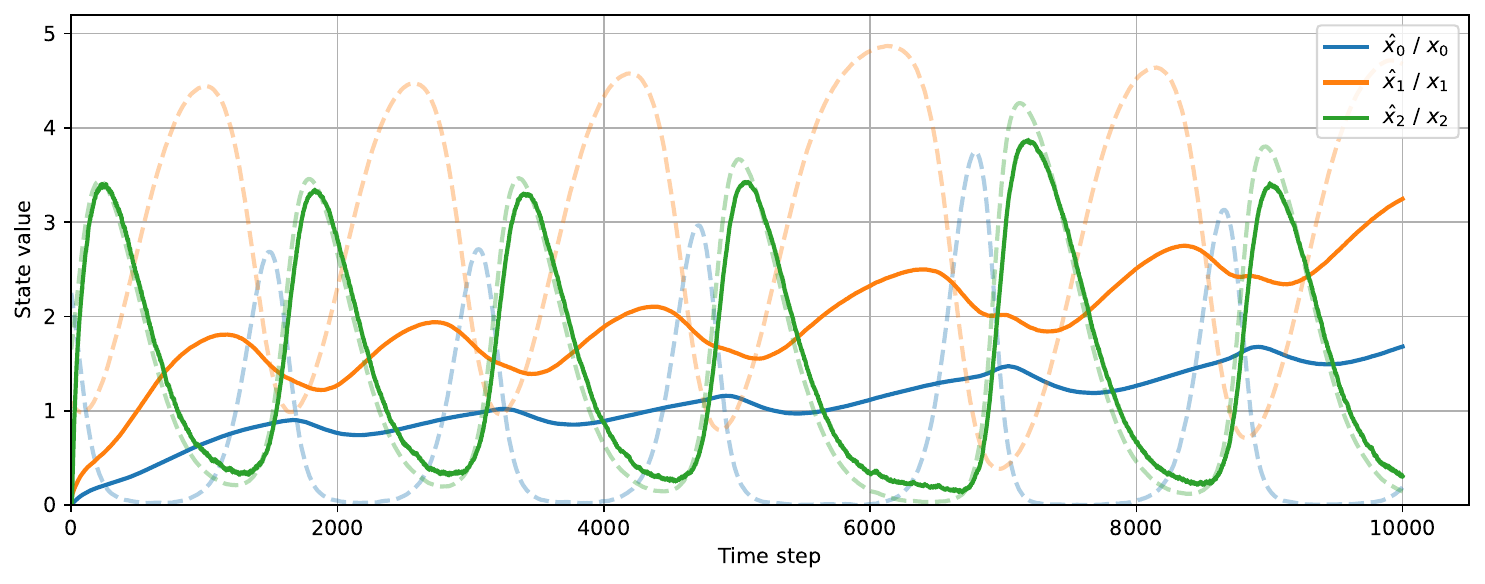}
        \caption{$\frac{E_{\Pi_y}}{E_{\Pi_x}}=1$}
        \label{fig:x_scenario=1_kx=3_ratio=1}
    \end{subfigure}\hfill
    \begin{subfigure}[t]{0.48\textwidth}
        \centering
        \includegraphics[width=\linewidth]{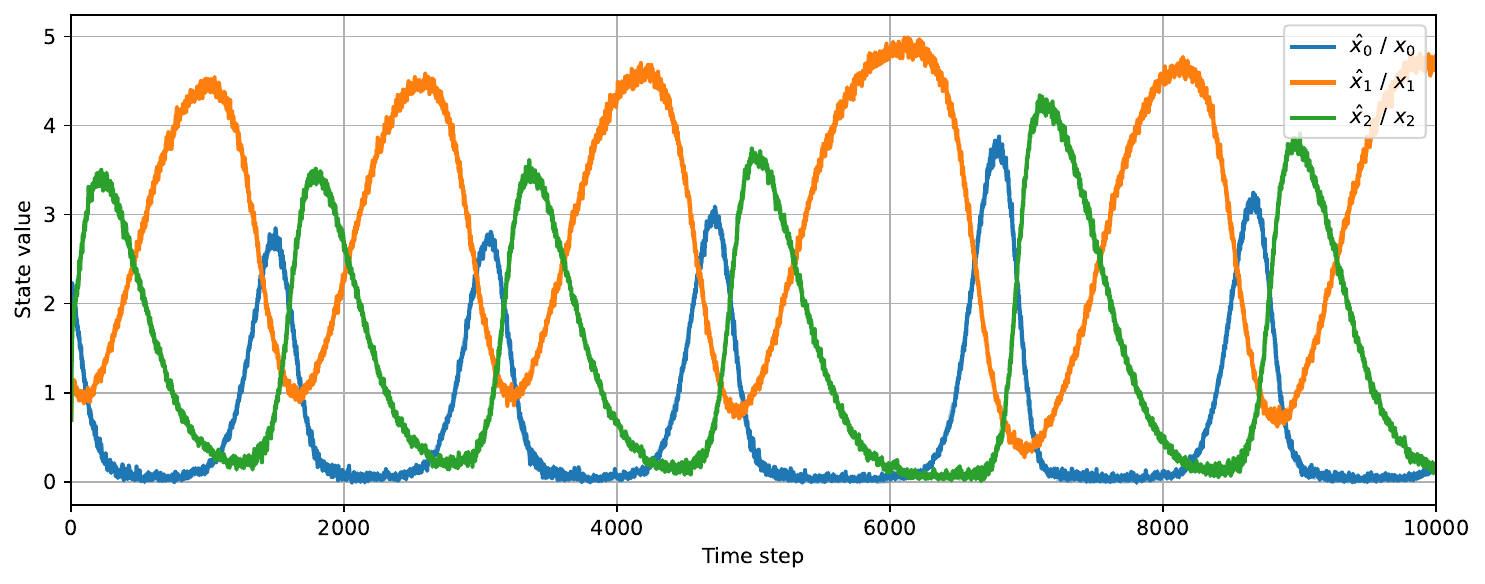}
        \caption{$\frac{E_{\Pi_y}}{E_{\Pi_x}}=1$}
        \label{fig:x_scenario=2_kx=3_ratio=1}
    \end{subfigure}


\end{figure}

\begin{figure}[H]\ContinuedFloat
    \centering

    \begin{subfigure}[t]{0.48\textwidth}
        \centering
        \includegraphics[width=\linewidth]{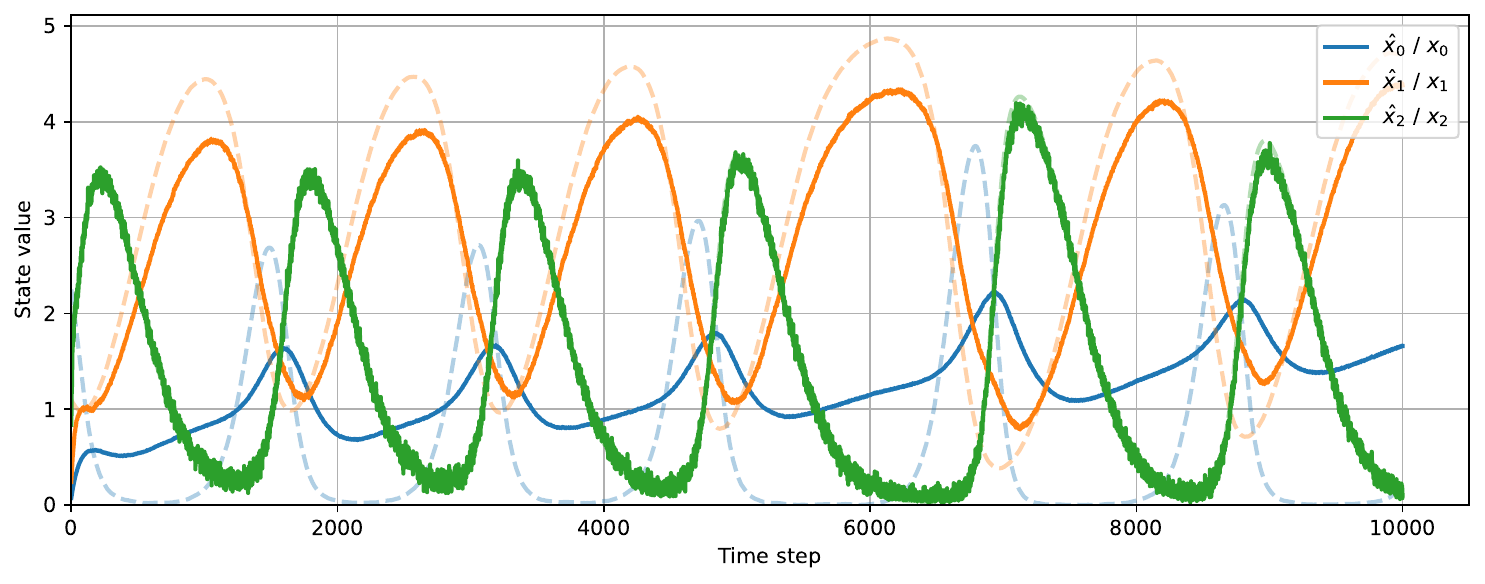}
        \caption{$\frac{E_{\Pi_y}}{E_{\Pi_x}}=10$}
        \label{fig:x_scenario=1_kx=3_ratio=10}
    \end{subfigure}\hfill
    \begin{subfigure}[t]{0.48\textwidth}
        \centering
        \includegraphics[width=\linewidth]{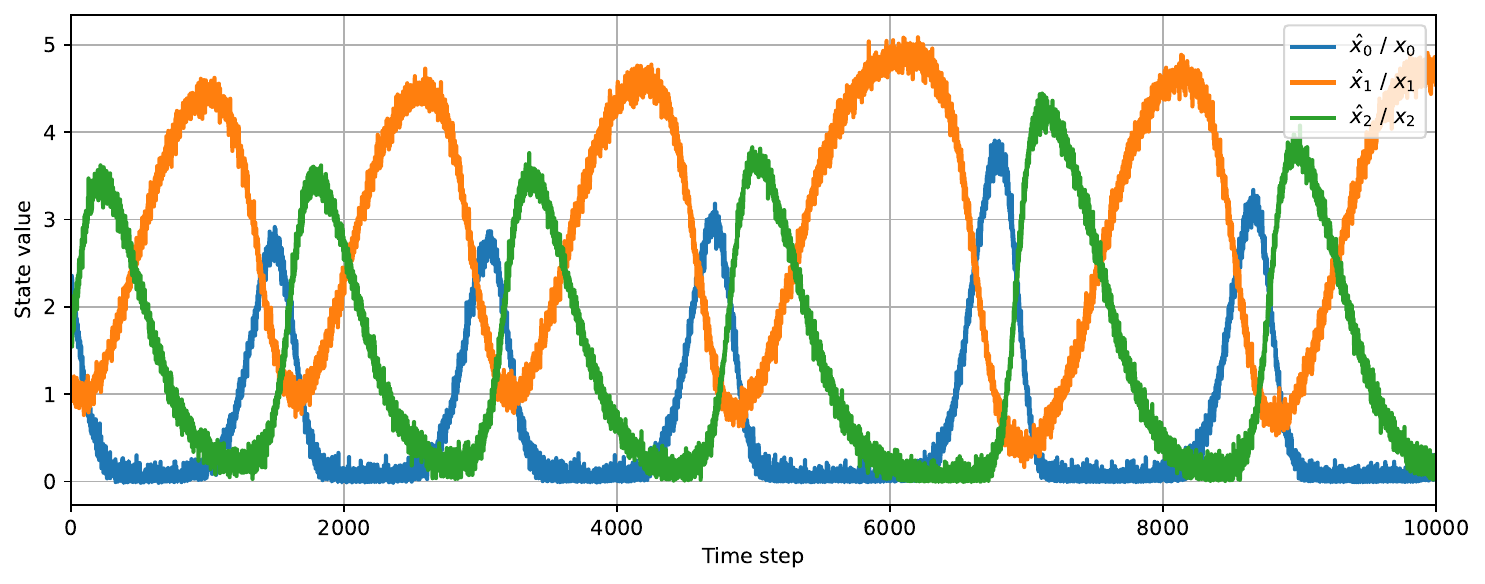}
        \caption{$\frac{E_{\Pi_y}}{E_{\Pi_x}}=10$}
        \label{fig:x_scenario=2_kx=3_ratio=10}
    \end{subfigure}

    \vspace{0.3cm}

    \begin{subfigure}[t]{0.48\textwidth}
        \centering
        \includegraphics[width=\linewidth]{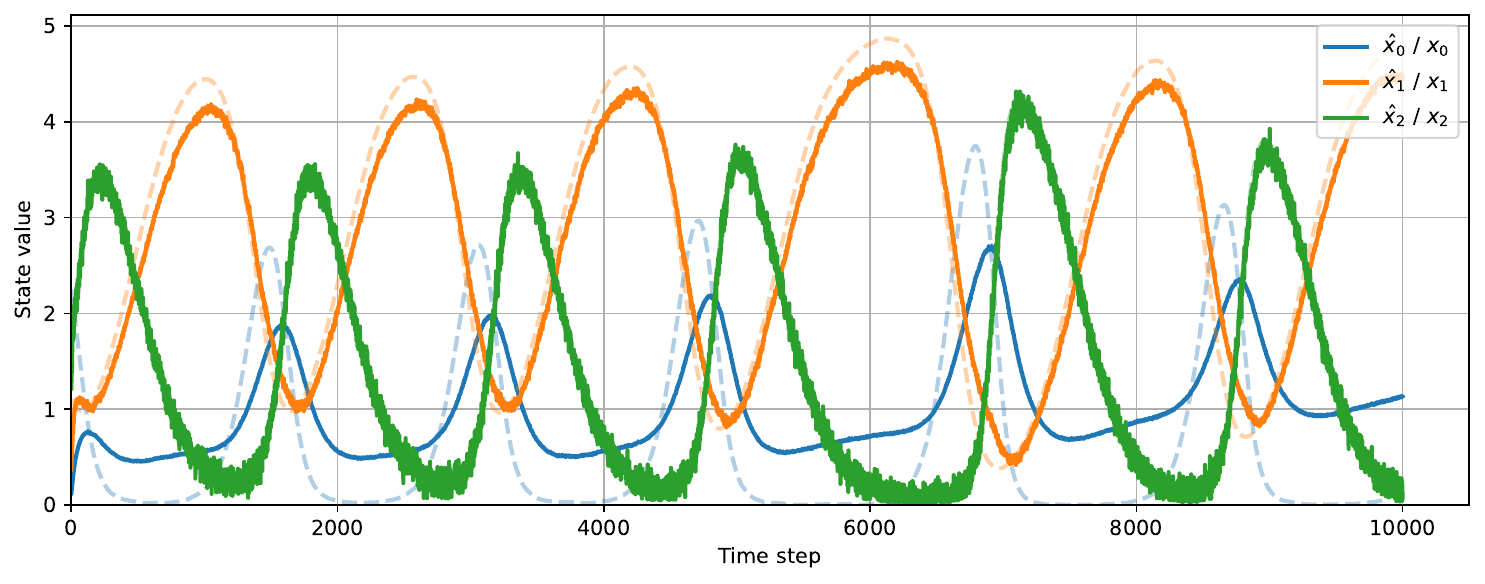}
        \caption{$\frac{E_{\Pi_y}}{E_{\Pi_x}}=25$}
        \label{fig:x_scenario=1_kx=3_ratio=25}
    \end{subfigure}\hfill
    \begin{subfigure}[t]{0.48\textwidth}
        \centering
        \includegraphics[width=\linewidth]{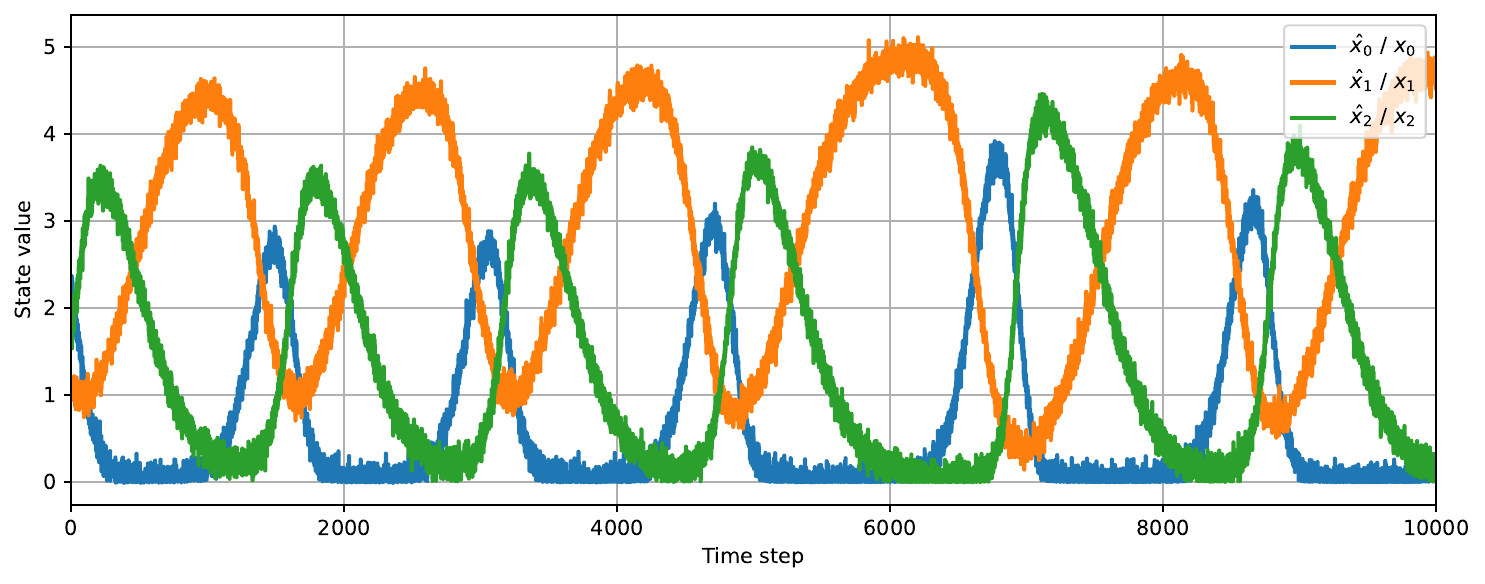}
        \caption{$\frac{E_{\Pi_y}}{E_{\Pi_x}}=25$}
        \label{fig:x_scenario=2_kx=3_ratio=25}
    \end{subfigure}

    \vspace{0.3cm}

    \begin{subfigure}[t]{0.48\textwidth}
        \centering
        \includegraphics[width=\linewidth]{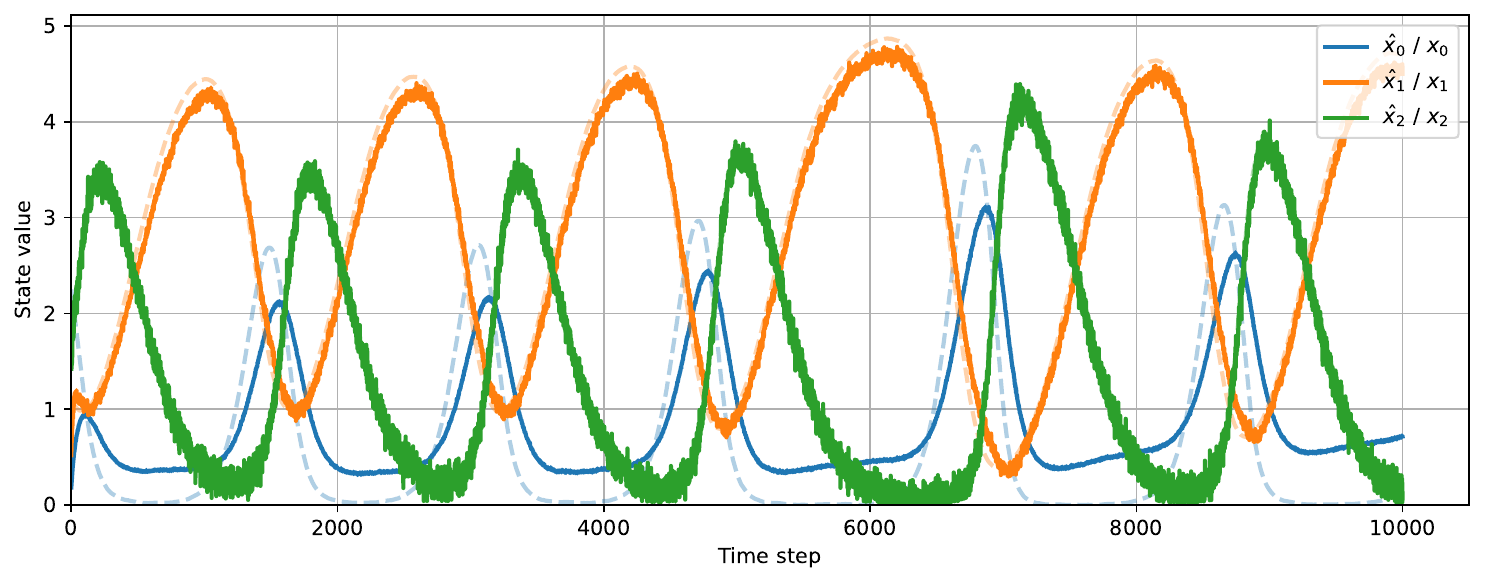}
        \caption{$\frac{E_{\Pi_y}}{E_{\Pi_x}}=50$}
        \label{fig:x_scenario=1_kx=3_ratio=50}
    \end{subfigure}\hfill
    \begin{subfigure}[t]{0.48\textwidth}
        \centering
        \includegraphics[width=\linewidth]{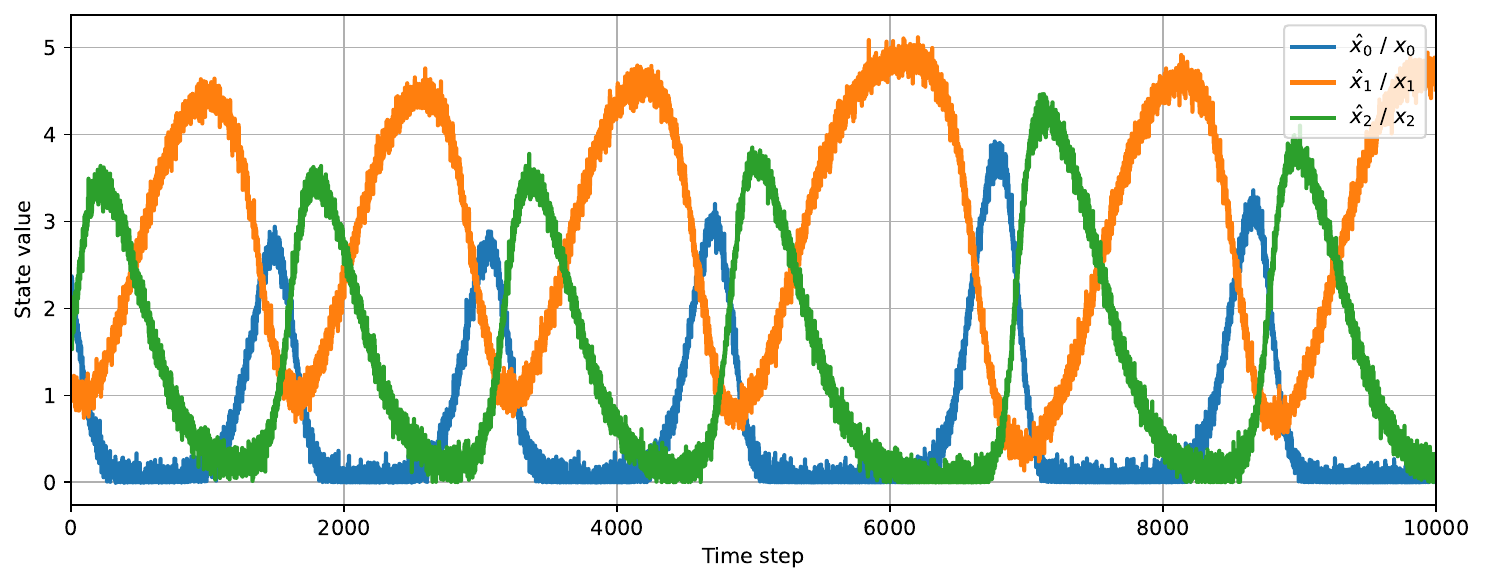}
        \caption{$\frac{E_{\Pi_y}}{E_{\Pi_x}}=50$}
        \label{fig:x_scenario=2_kx=3_ratio=50}
    \end{subfigure}

    \caption[]{The inferred states for $k_x=3$ orders of motion. The left panel is for scenario-different: Lorenz-GM vs. GLV-GP, and the right panel is for scenario-same: GLV-GM vs. GLV-GP. Each row corresponds to a different precision prior ratio.}
    \label{fig:xhat_kx3_all_ratios}

\end{figure}

\section{\texorpdfstring{Predicted sensations $\hat{y}$}{Predicted sensations y-hat}}
\label{app:y}
The following demonstrates the predicted sensations, $\hat{\bfy}_t$, for $k_x=2$ (Appendix.~\ref{app:y_kx=2}), and $k_x=3$ (Appendix.~\ref{app:y_kx=3}) orders of motion. In all figures in this section, the transparent curves correspond to the true noisy sensations, $\bfy_t$, while the solid lines correspond to the generated sensations, $\hat{\bfy}_t$. Different colours correspond to different dimensions of the observations. Similar to Appendix.~\ref{app:x}, the horizontal axis is time. The left panel corresponds to scenario-different: Lorenz-GM vs. GLV-GP, while the right panel corresponds to scenario-same: GLV-GM vs. GLV-GP. Finally, each row corresponds to a specific precision prior ratio and for any particular ratio, the trajectories belonging to the GM with lowest FA has been reported. The solid curves are the predictions and the transparent curves are the actual sensations.

\subsection{\texorpdfstring{$k_x = 2$}{k x = 2}}
\label{app:y_kx=2}
The following demonstrates the predicted sensations $\hat{\bfy}_t$ for $k_x=2$ orders of motion. 
In the left panel, we can see that as the precision prior ratio increases, the predicted sensations improve (i.e., moving from Fig.~\ref{fig:y_scenario=1_kx=2_ratio=0.02} to Fig.~\ref{fig:y_scenario=1_kx=2_ratio=50}). This is expected as by increasing the ratio, the GM begins to attend more and more towards minimising the prediction error, however, even when at its highest value of 50 in Fig.~\ref{fig:y_scenario=1_kx=2_ratio=50}, the Lorenz-GM fails terribly in predicting the correct sensations. As we will see in the next section, an increase in the orders of motion can remedy this for the Lorenz-GM.
In the right panel, since the functional form of the GM and GP is identical, we can already see that even using $k_x=2$ orders of motion and the lowest precision prior ratio of 0.02 (Fig.~\ref{fig:y_scenario=2_kx=2_ratio=0.02}), the predicted sensations are reasonable, especially when compared to a Lorenz GM under the same prior ratio value in Fig.~\ref{fig:y_scenario=1_kx=2_ratio=0.02}. Indeed, by increasing the ratio, we can see a clear improvement in the predicted sensations, especially at the $\frac{E_{\Pi_y}}{E_{\Pi_x}}=1$ ratio.
\begin{figure}[H]
    \centering

    \begin{subfigure}[t]{0.48\textwidth}
        \centering
        \includegraphics[width=\linewidth]{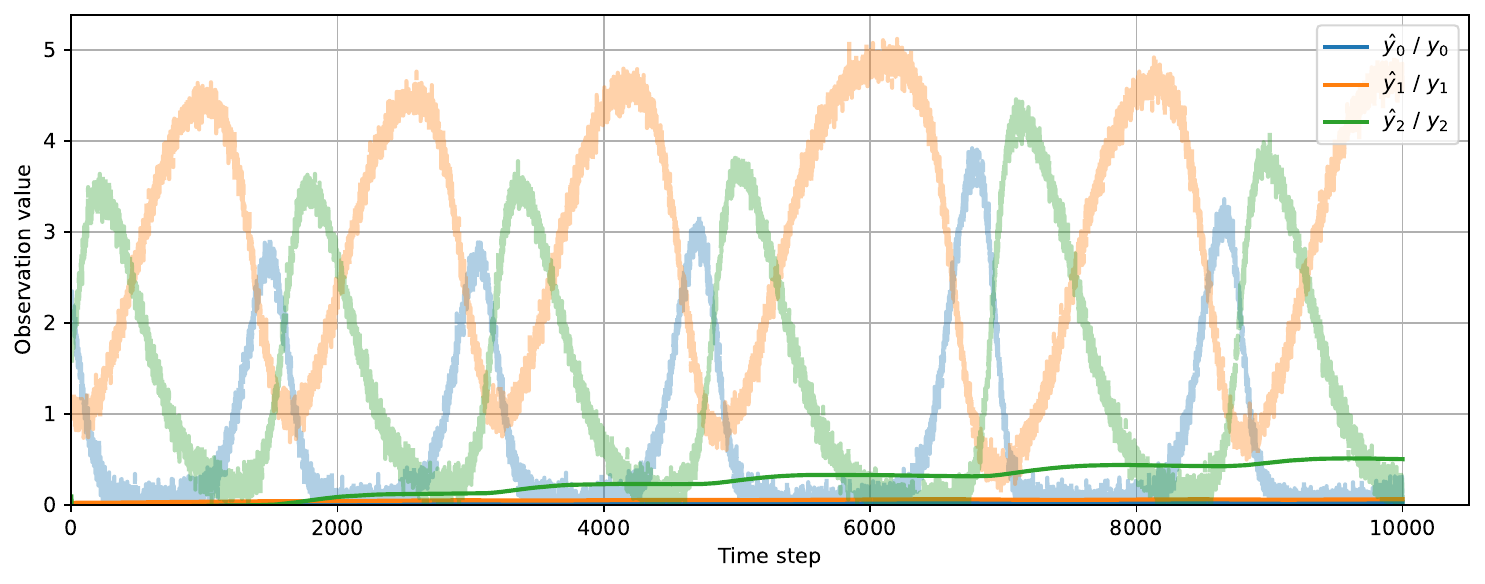}
        \caption{$\frac{E_{\Pi_y}}{E_{\Pi_x}}=0.02$}
        \label{fig:y_scenario=1_kx=2_ratio=0.02}
    \end{subfigure}\hfill
    \begin{subfigure}[t]{0.48\textwidth}
        \centering
        \includegraphics[width=\linewidth]{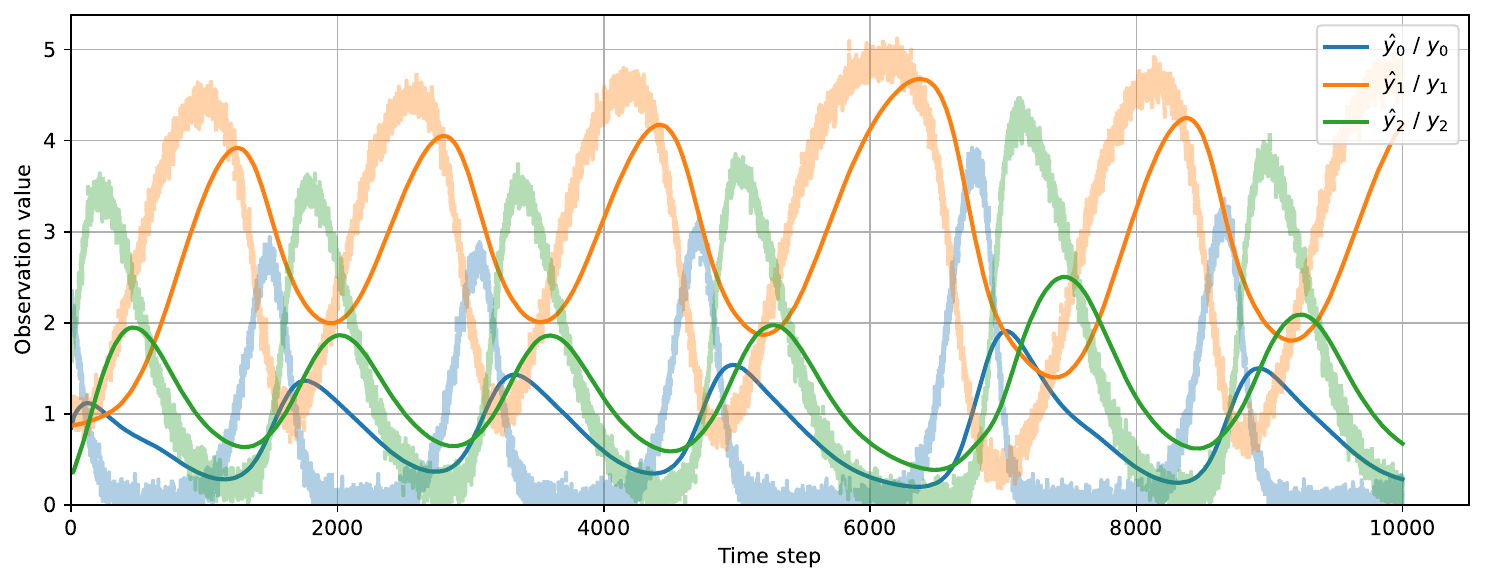}
        \caption{$\frac{E_{\Pi_y}}{E_{\Pi_x}}=0.02$}
        \label{fig:y_scenario=2_kx=2_ratio=0.02}
    \end{subfigure}

    \vspace{0.3cm}

    \begin{subfigure}[t]{0.48\textwidth}
        \centering
        \includegraphics[width=\linewidth]{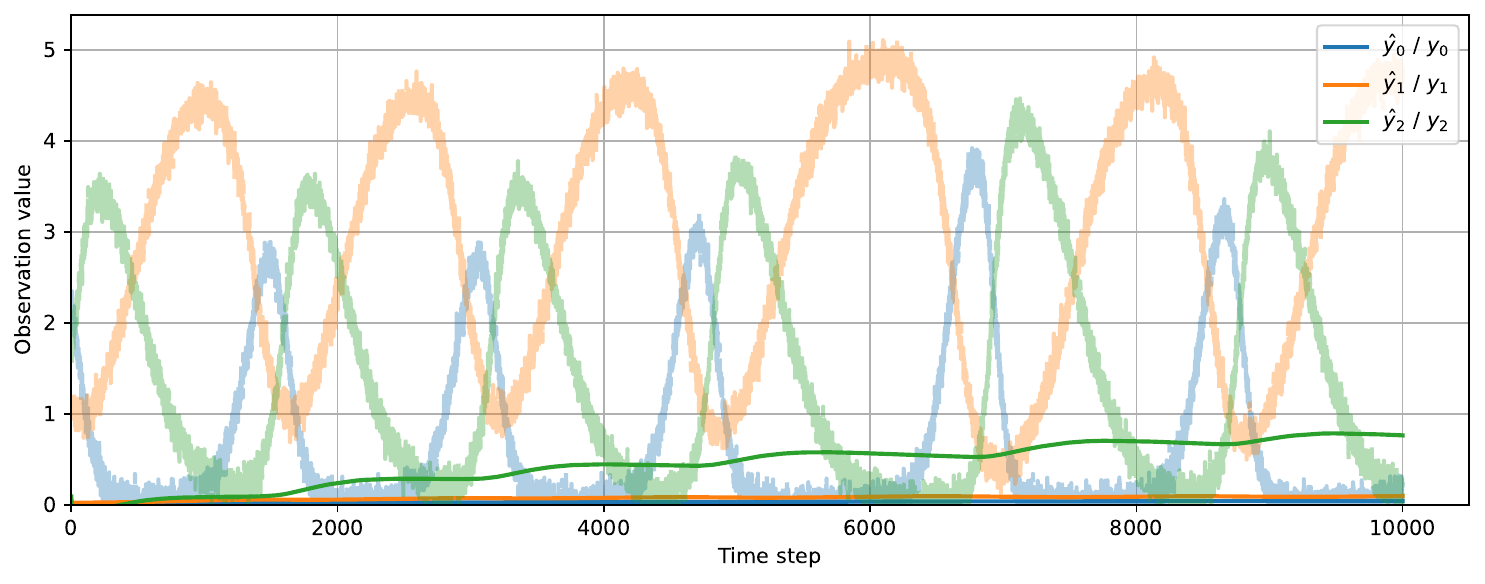}
        \caption{$\frac{E_{\Pi_y}}{E_{\Pi_x}}=0.04$}
        \label{fig:y_scenario=1_kx=2_ratio=0.04}
    \end{subfigure}\hfill
    \begin{subfigure}[t]{0.48\textwidth}
        \centering
        \includegraphics[width=\linewidth]{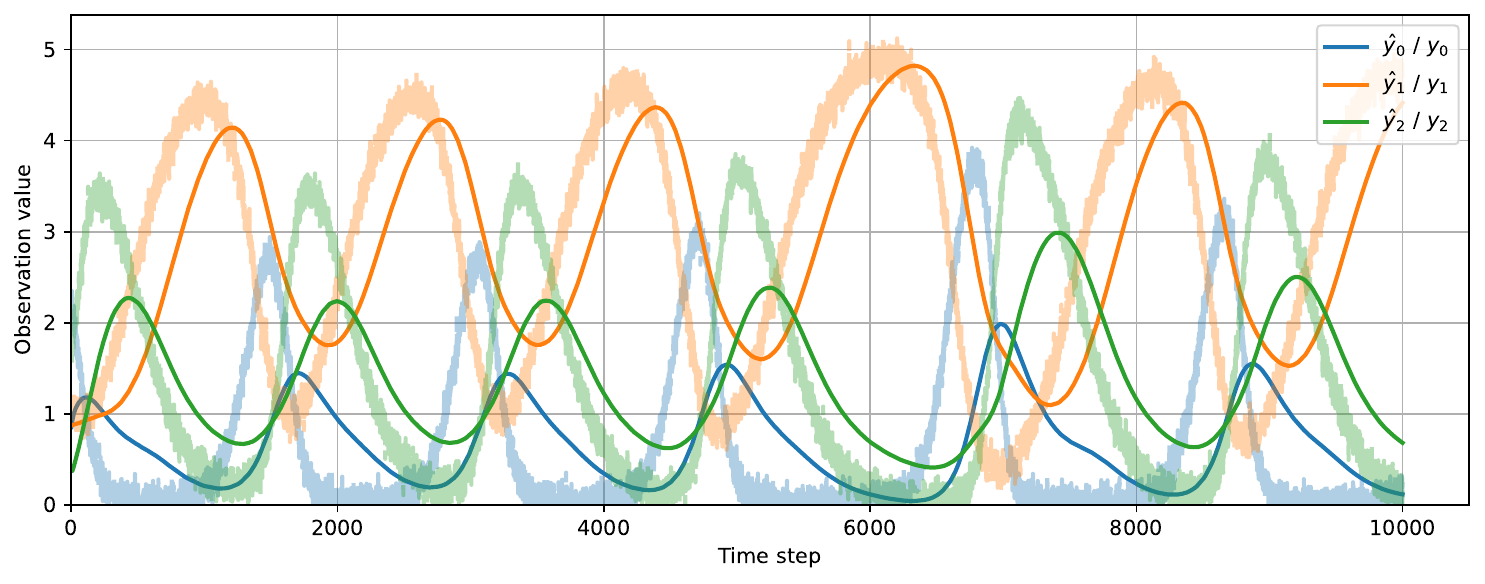}
        \caption{$\frac{E_{\Pi_y}}{E_{\Pi_x}}=0.04$}
        \label{fig:y_scenario=2_kx=2_ratio=0.04}
    \end{subfigure}

    \vspace{0.3cm}

    \begin{subfigure}[t]{0.48\textwidth}
        \centering
        \includegraphics[width=\linewidth]{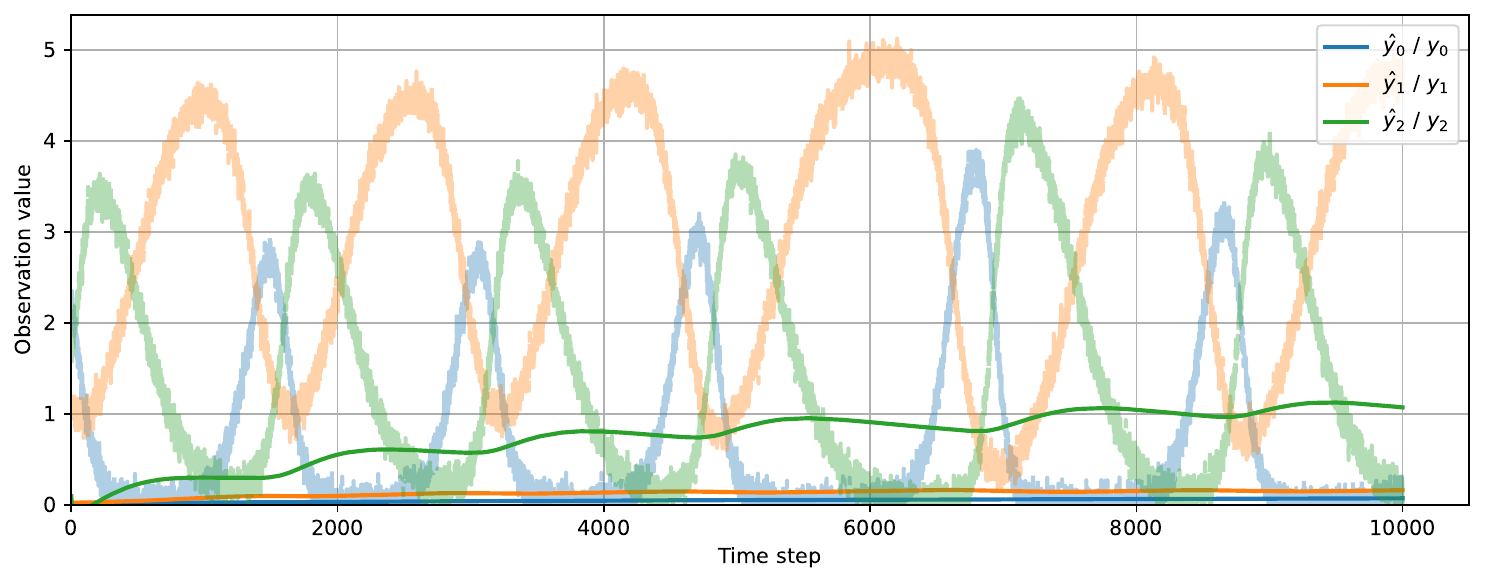}
        \caption{$\frac{E_{\Pi_y}}{E_{\Pi_x}}=0.1$}
        \label{fig:y_scenario=1_kx=2_ratio=0.1}
    \end{subfigure}\hfill
    \begin{subfigure}[t]{0.48\textwidth}
        \centering
        \includegraphics[width=\linewidth]{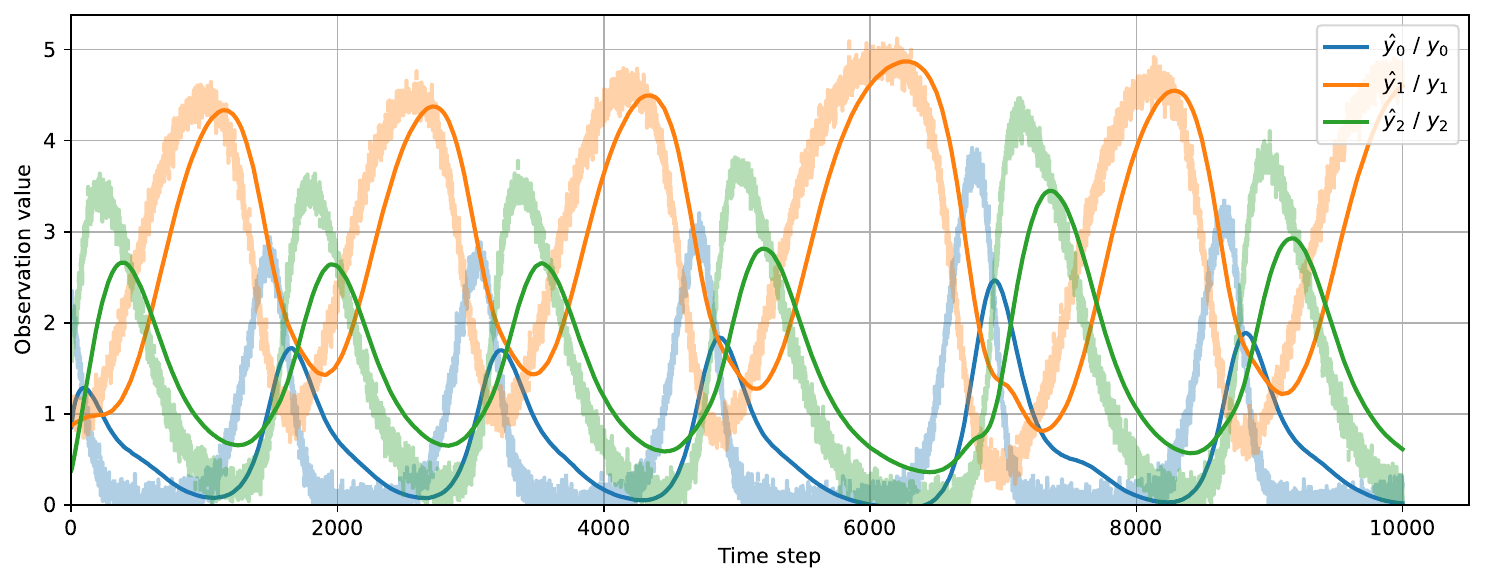}
        \caption{$\frac{E_{\Pi_y}}{E_{\Pi_x}}=0.1$}
        \label{fig:y_scenario=2_kx=2_ratio=0.1}
    \end{subfigure}

    \vspace{0.3cm}

    \begin{subfigure}[t]{0.48\textwidth}
        \centering
        \includegraphics[width=\linewidth]{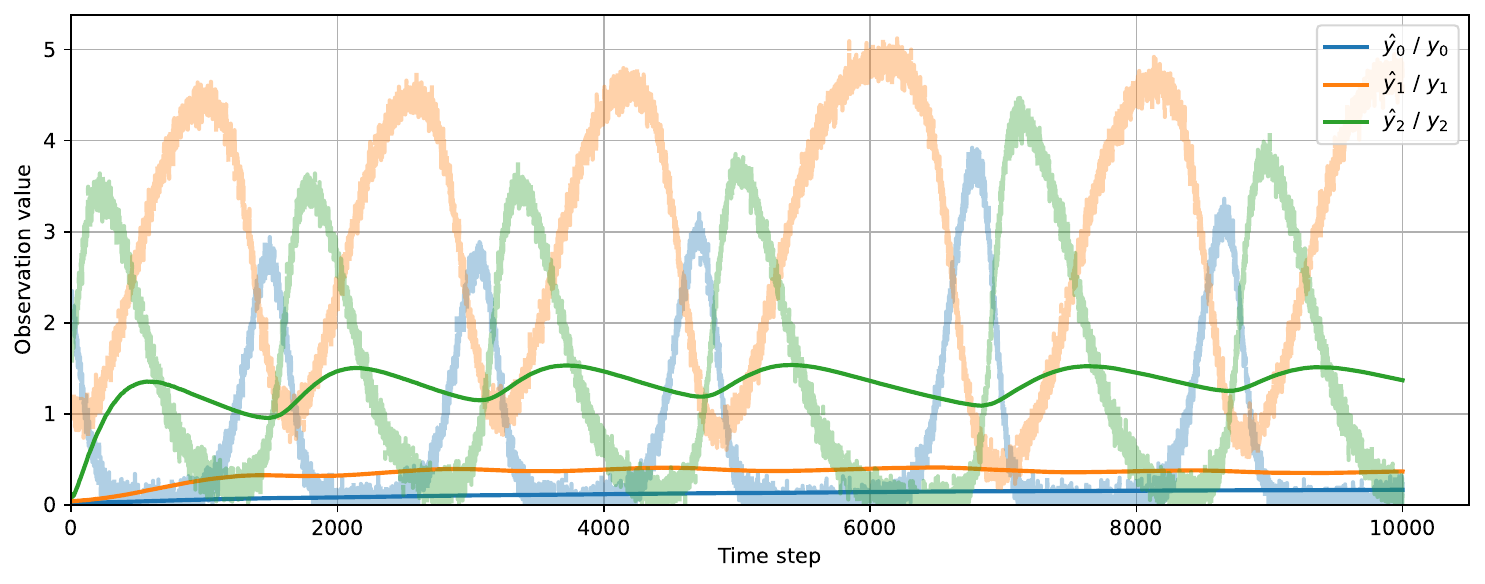}
        \caption{$\frac{E_{\Pi_y}}{E_{\Pi_x}}=1$}
        \label{fig:y_scenario=1_kx=2_ratio=1}
    \end{subfigure}\hfill
    \begin{subfigure}[t]{0.48\textwidth}
        \centering
        \includegraphics[width=\linewidth]{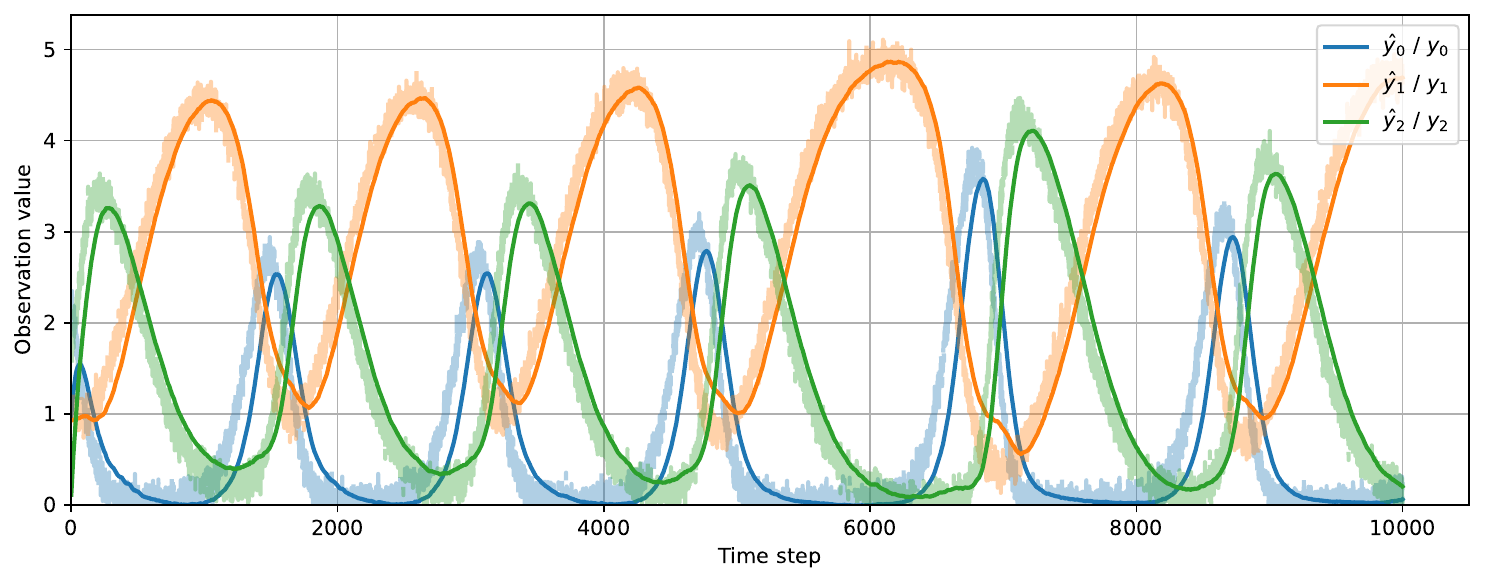}
        \caption{$\frac{E_{\Pi_y}}{E_{\Pi_x}}=1$}
        \label{fig:y_scenario=2_kx=2_ratio=1}
    \end{subfigure}


\end{figure}
\begin{figure}[H]\ContinuedFloat
    \centering

    \begin{subfigure}[t]{0.48\textwidth}
        \centering
        \includegraphics[width=\linewidth]{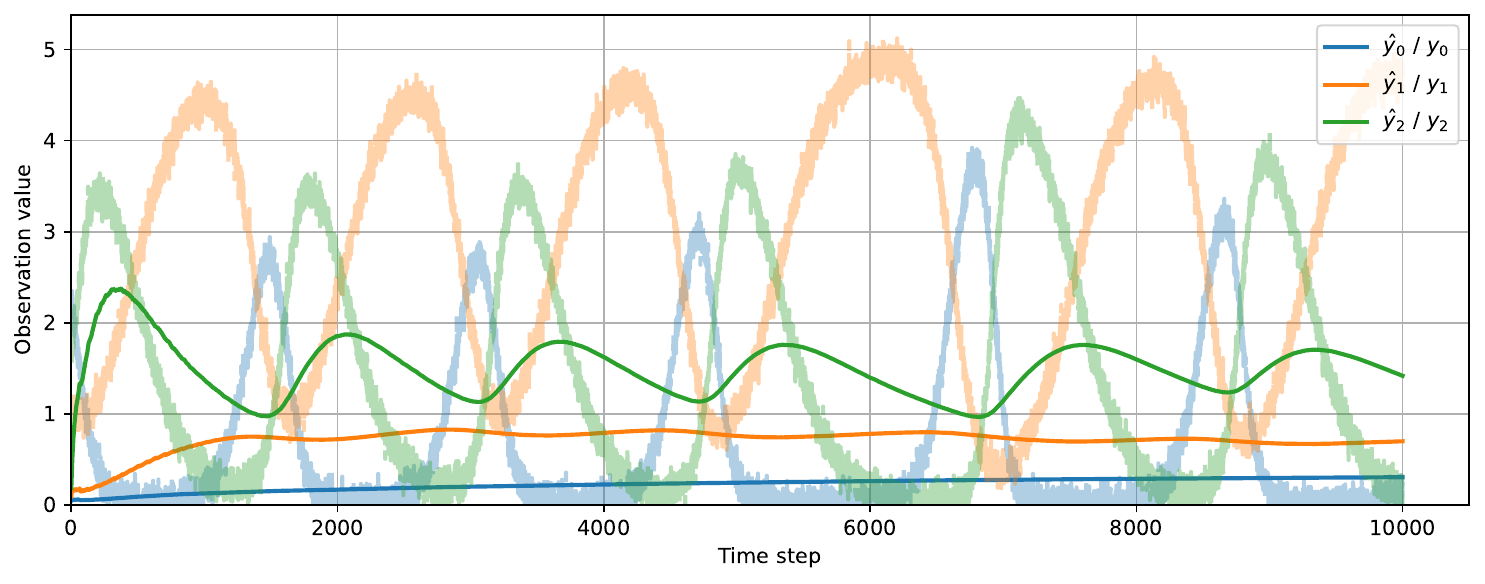}
        \caption{$\frac{E_{\Pi_y}}{E_{\Pi_x}}=10$}
        \label{fig:y_scenario=1_kx=2_ratio=10}
    \end{subfigure}\hfill
    \begin{subfigure}[t]{0.48\textwidth}
        \centering
        \includegraphics[width=\linewidth]{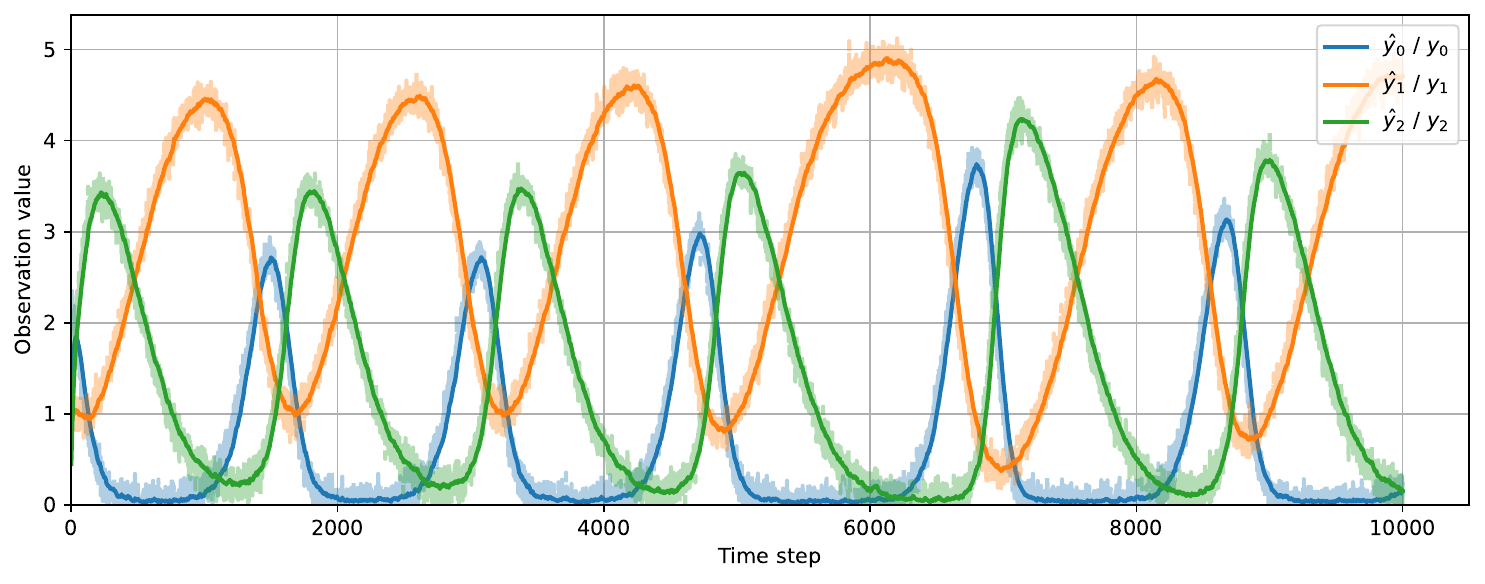}
        \caption{$\frac{E_{\Pi_y}}{E_{\Pi_x}}=10$}
        \label{fig:y_scenario=2_kx=2_ratio=10}
    \end{subfigure}

    \vspace{0.3cm}

    \begin{subfigure}[t]{0.48\textwidth}
        \centering
        \includegraphics[width=\linewidth]{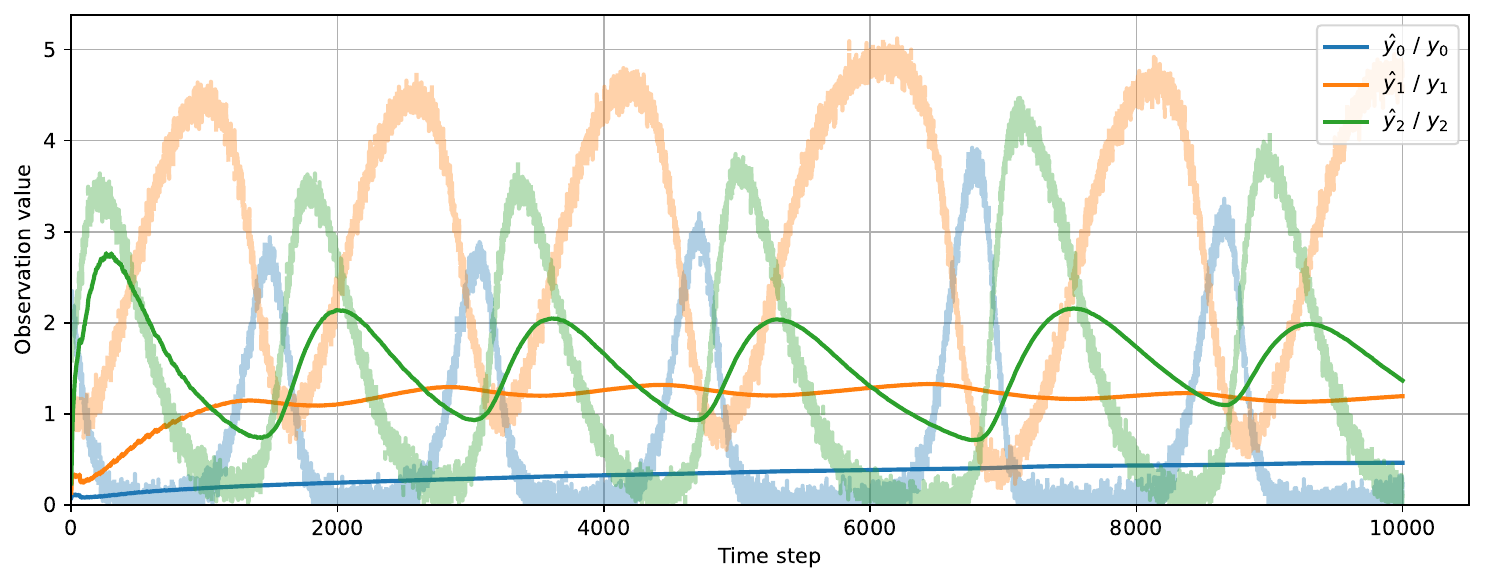}
        \caption{$\frac{E_{\Pi_y}}{E_{\Pi_x}}=25$}
        \label{fig:y_scenario=1_kx=2_ratio=25}
    \end{subfigure}\hfill
    \begin{subfigure}[t]{0.48\textwidth}
        \centering
        \includegraphics[width=\linewidth]{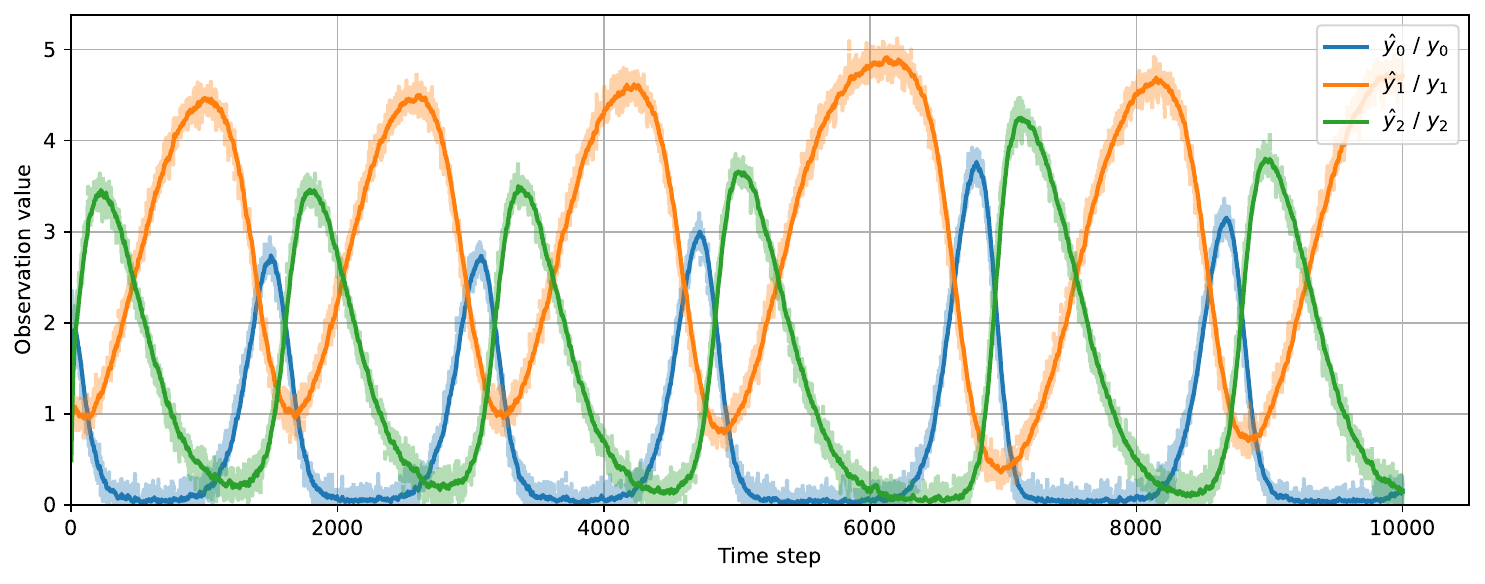}
        \caption{$\frac{E_{\Pi_y}}{E_{\Pi_x}}=25$}
        \label{fig:y_scenario=2_kx=2_ratio=25}
    \end{subfigure}

    \vspace{0.3cm}

    \begin{subfigure}[t]{0.48\textwidth}
        \centering
        \includegraphics[width=\linewidth]{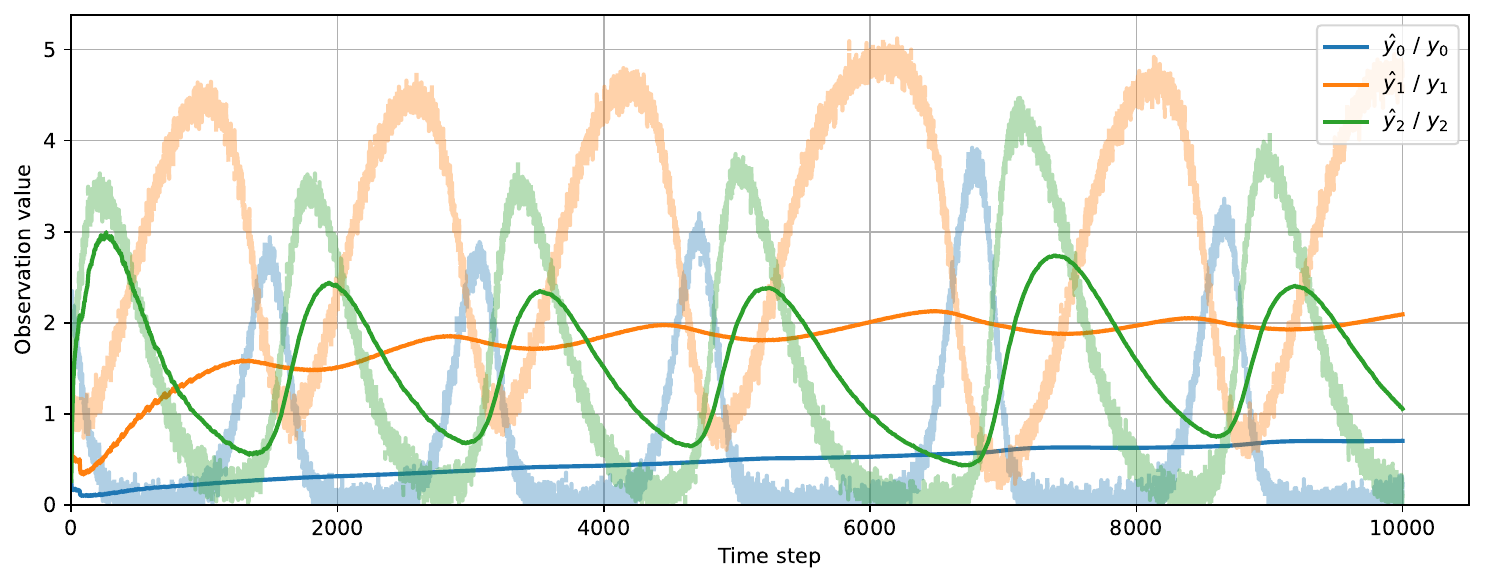}
        \caption{$\frac{E_{\Pi_y}}{E_{\Pi_x}}=50$}
        \label{fig:y_scenario=1_kx=2_ratio=50}
    \end{subfigure}\hfill
    \begin{subfigure}[t]{0.48\textwidth}
        \centering
        \includegraphics[width=\linewidth]{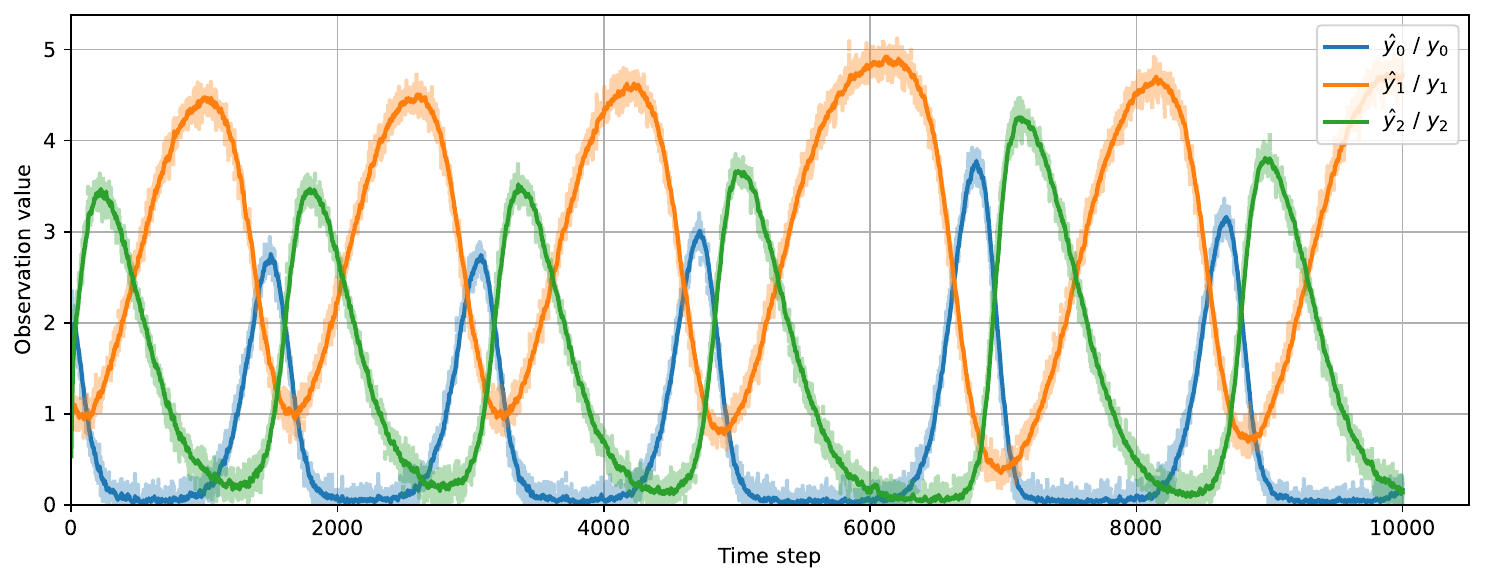}
        \caption{$\frac{E_{\Pi_y}}{E_{\Pi_x}}=50$}
        \label{fig:y_scenario=2_kx=2_ratio=50}
    \end{subfigure}

    \caption[]{The predicted sensations, $\hat{y}$, for $k_x=2$ orders of motion. The left panel is for scenario-different: Lorenz-GM vs. GLV-GP, and the right panel is for scenario-same: GLV-GM vs. GLV-GP. Each row corresponds to a different precision prior ratio.}
\end{figure}

\subsection{\texorpdfstring{$k_x = 3$}{k x = 3}}
\label{app:y_kx=3}
The following demonstrates the predicted sensations $\hat{\gcy}_t$ for $k_x=3$ orders of motion. In the left panel: Lorenz-GM vs. GLV-GP, we can see that as the precision prior ratio increases, the predicted sensations improve (i.e., moving from Fig.~\ref{fig:y_scenario=1_kx=3_ratio=0.02} to Fig.~\ref{fig:y_scenario=1_kx=3_ratio=50}). The Lorenz-GM becomes much better in predicting the sensations, especially if we compare these figures with Fig.~\ref{fig:y_scenario=1_kx=2_ratio=0.02} to Fig.~\ref{fig:y_scenario=1_kx=2_ratio=50}, the added value of having $k_x=3$ orders of motion becomes much more apparent. The increase number of generalised coordinates, allows the Lorenz-GM to compensate for the fact that it has a completely different functional form than the GLV-GP.

Let us now consider the right panel: GLV-GM vs. GLV-GP. Since the functional form of the GM and GP is identical, we can see immediately that having $k_x=3$ orders of motion even under the lowest precision prior ratio 0.02 in Fig.~\ref{fig:y_scenario=2_kx=3_ratio=0.02} is performing much better than the previous case with $k_x=2$ orders of motion with precision prior ratio 0.02 in Fig.~\ref{fig:y_scenario=2_kx=2_ratio=0.02}, highlighting the value of generalised coordinates. However, we can also see that as we move beyond $\frac{E_{\Pi_y}}{E_{\Pi_x}}=1$ in Fig.~\ref{fig:y_scenario=2_kx=3_ratio=1} to $\frac{E_{\Pi_y}}{E_{\Pi_x}}=50$ in Fig.~\ref{fig:y_scenario=2_kx=3_ratio=50}, the GLV-GM begins to completely ignore the dynamics and focus on minimising the sensation prediction error. Note that this was not the case for $k_x=2$ in Fig.~\ref{fig:y_scenario=2_kx=2_ratio=1} to Fig.~\ref{fig:y_scenario=2_kx=2_ratio=50}. 

\begin{figure}[H]
    \centering

    \begin{subfigure}[t]{0.48\textwidth}
        \centering
        \includegraphics[width=\linewidth]{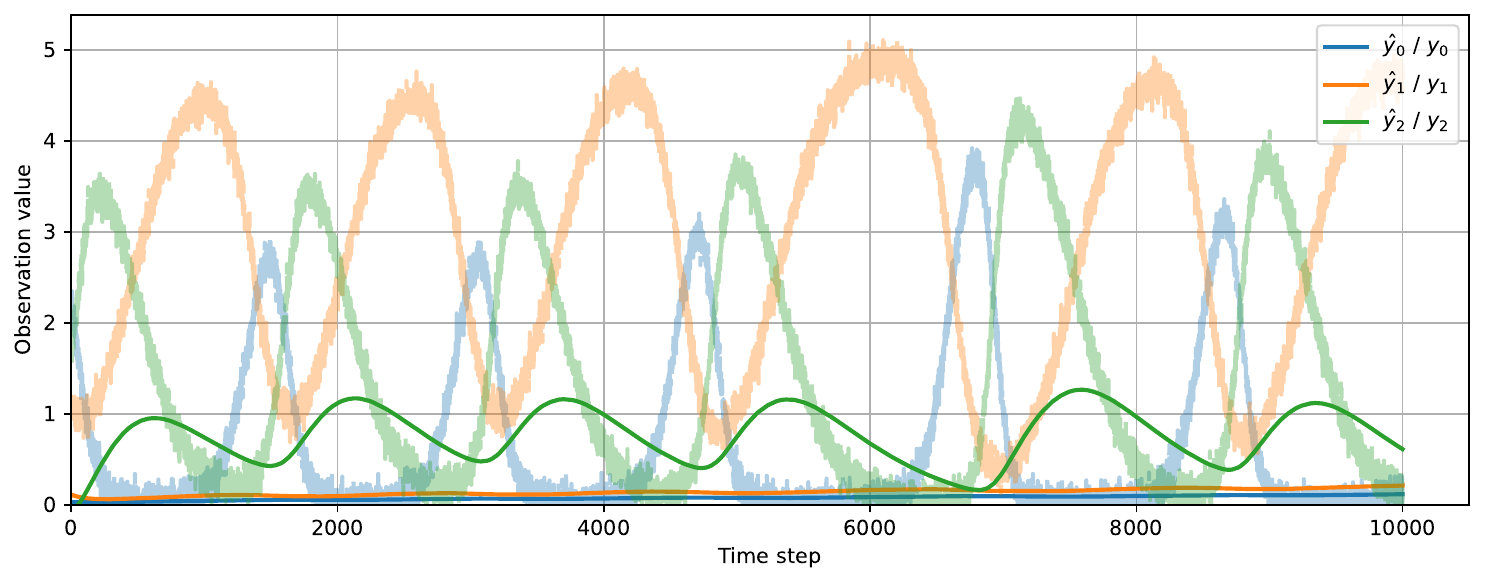}
        \caption{$\frac{E_{\Pi_y}}{E_{\Pi_x}}=0.02$}
        \label{fig:y_scenario=1_kx=3_ratio=0.02}
    \end{subfigure}\hfill
    \begin{subfigure}[t]{0.48\textwidth}
        \centering
        \includegraphics[width=\linewidth]{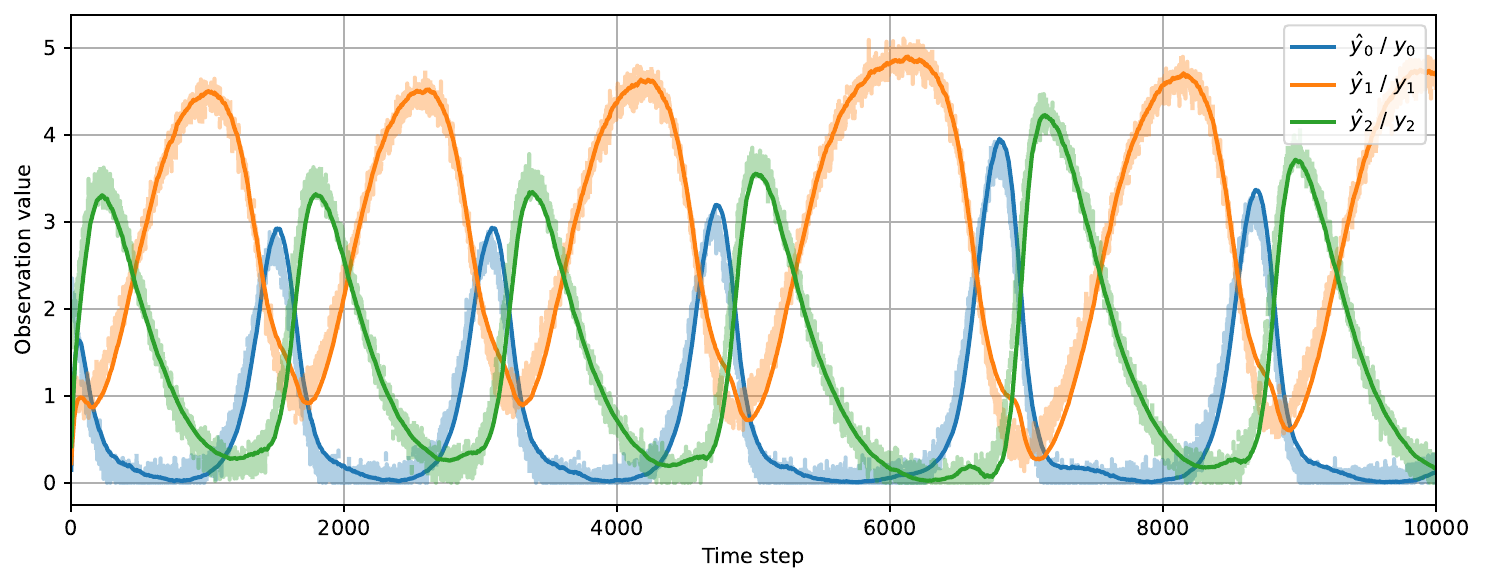}
        \caption{$\frac{E_{\Pi_y}}{E_{\Pi_x}}=0.02$}
        \label{fig:y_scenario=2_kx=3_ratio=0.02}
    \end{subfigure}

    \vspace{0.3cm}

    \begin{subfigure}[t]{0.48\textwidth}
        \centering
        \includegraphics[width=\linewidth]{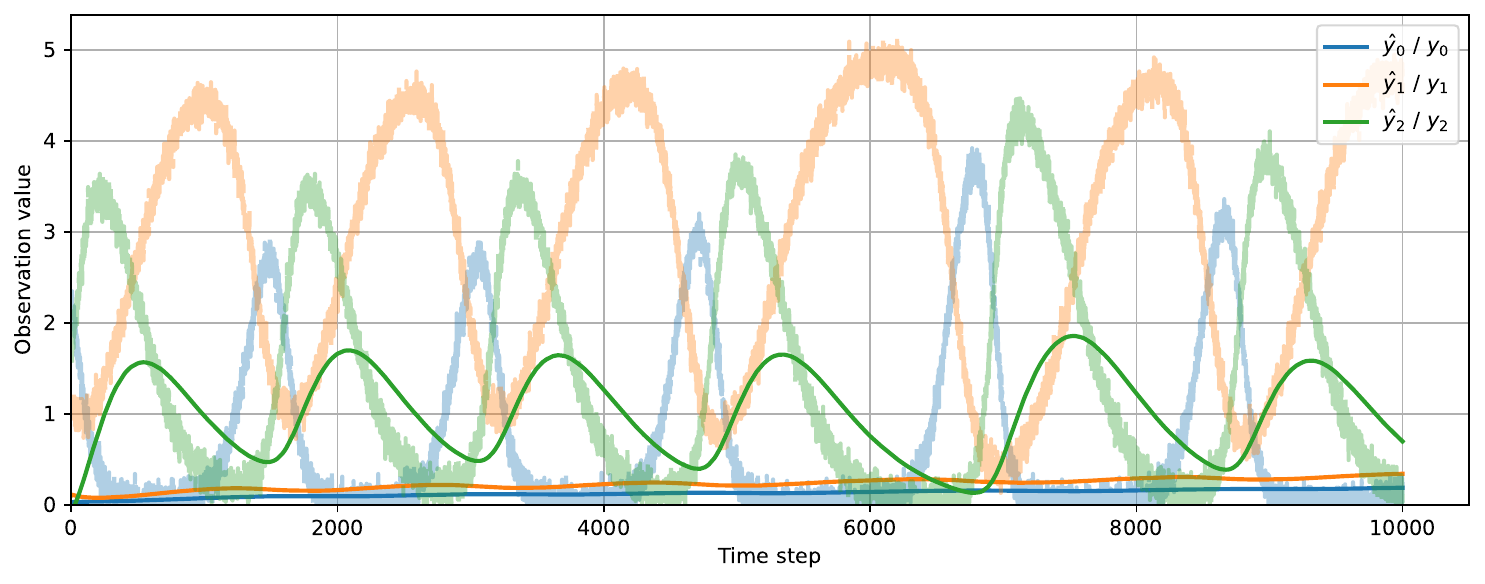}
        \caption{$\frac{E_{\Pi_y}}{E_{\Pi_x}}=0.04$}
        \label{fig:y_scenario=1_kx=3_ratio=0.04}
    \end{subfigure}\hfill
    \begin{subfigure}[t]{0.48\textwidth}
        \centering
        \includegraphics[width=\linewidth]{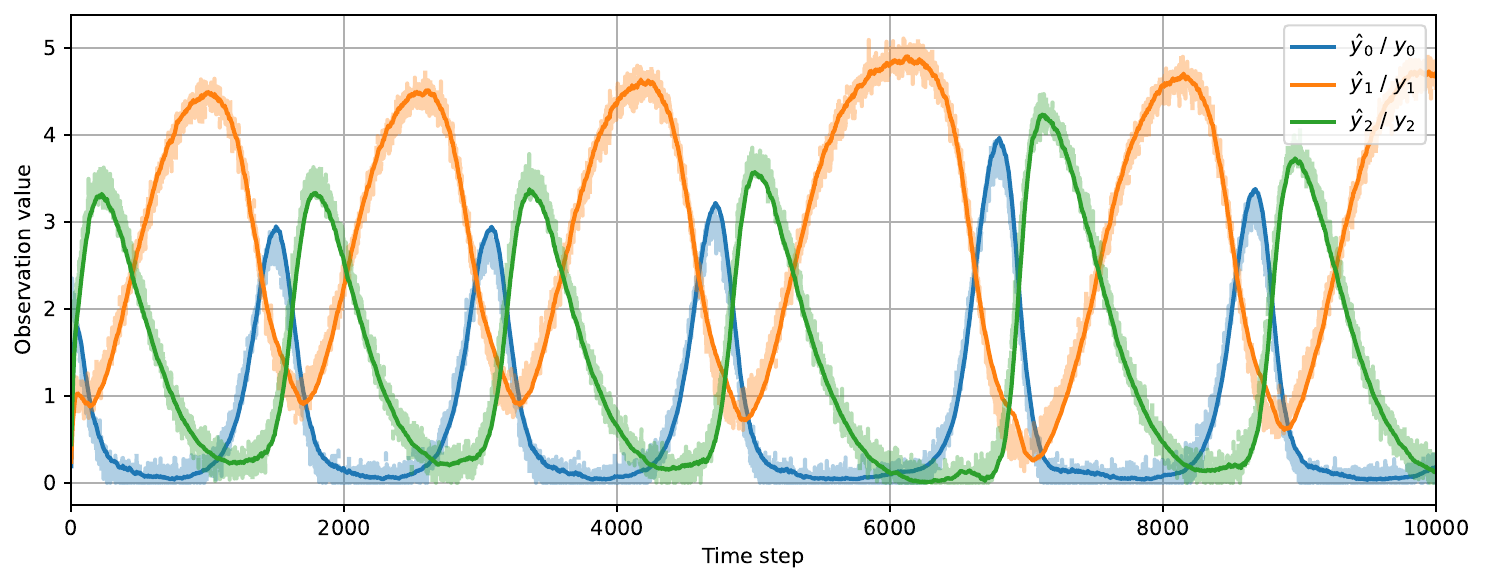}
        \caption{$\frac{E_{\Pi_y}}{E_{\Pi_x}}=0.04$}
        \label{fig:y_scenario=2_kx=3_ratio=0.04}
    \end{subfigure}

    \vspace{0.3cm}

    \begin{subfigure}[t]{0.48\textwidth}
        \centering
        \includegraphics[width=\linewidth]{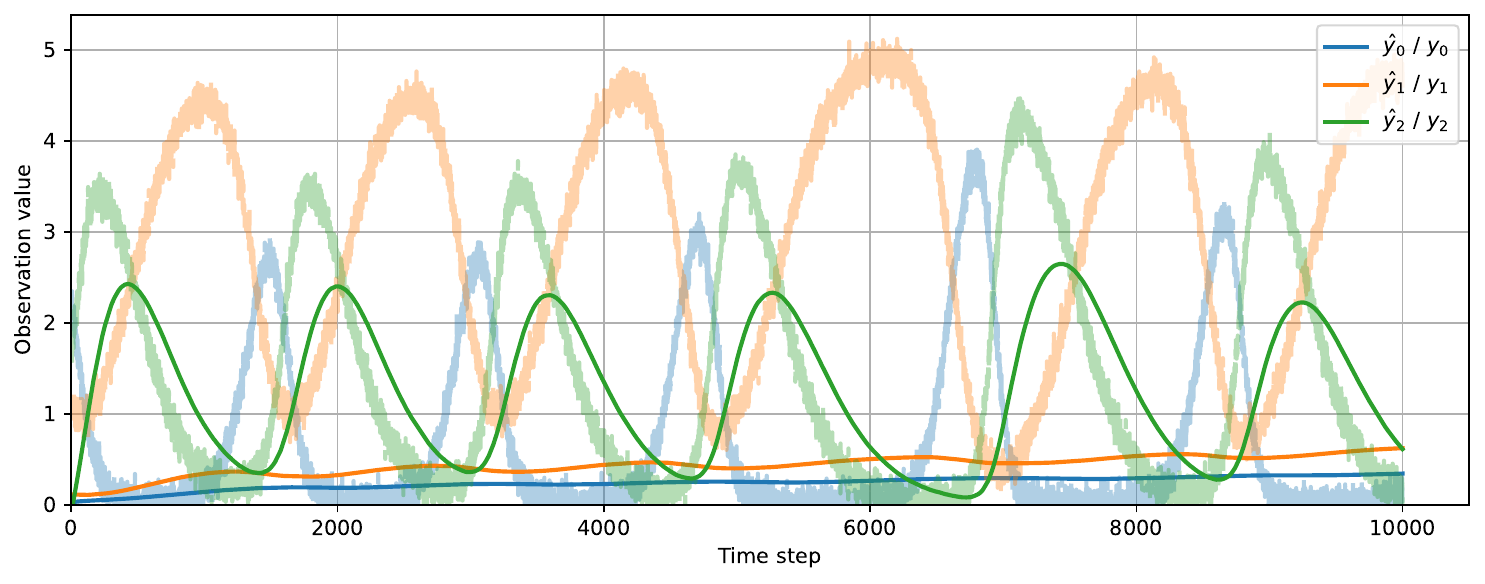}
        \caption{$\frac{E_{\Pi_y}}{E_{\Pi_x}}=0.1$}
        \label{fig:y_scenario=1_kx=3_ratio=0.1}
    \end{subfigure}\hfill
    \begin{subfigure}[t]{0.48\textwidth}
        \centering
        \includegraphics[width=\linewidth]{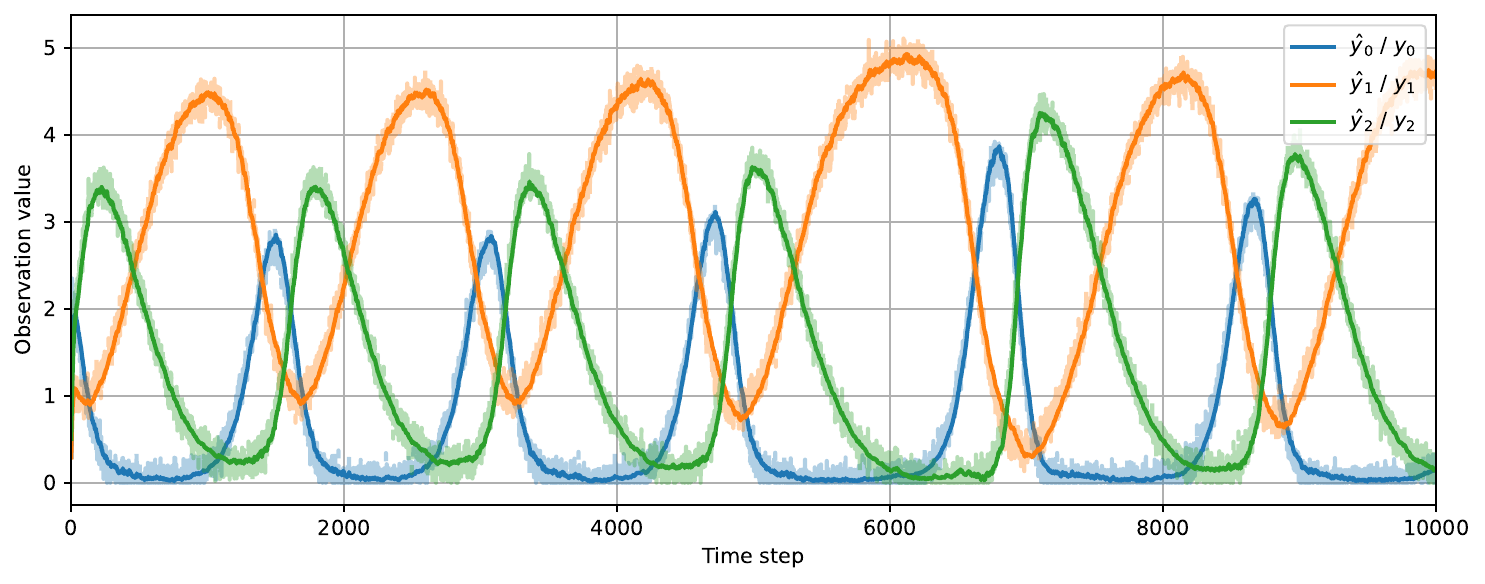}
        \caption{$\frac{E_{\Pi_y}}{E_{\Pi_x}}=0.1$}
        \label{fig:y_scenario=2_kx=3_ratio=0.1}
    \end{subfigure}

    \vspace{0.3cm}

    \begin{subfigure}[t]{0.48\textwidth}
        \centering
        \includegraphics[width=\linewidth]{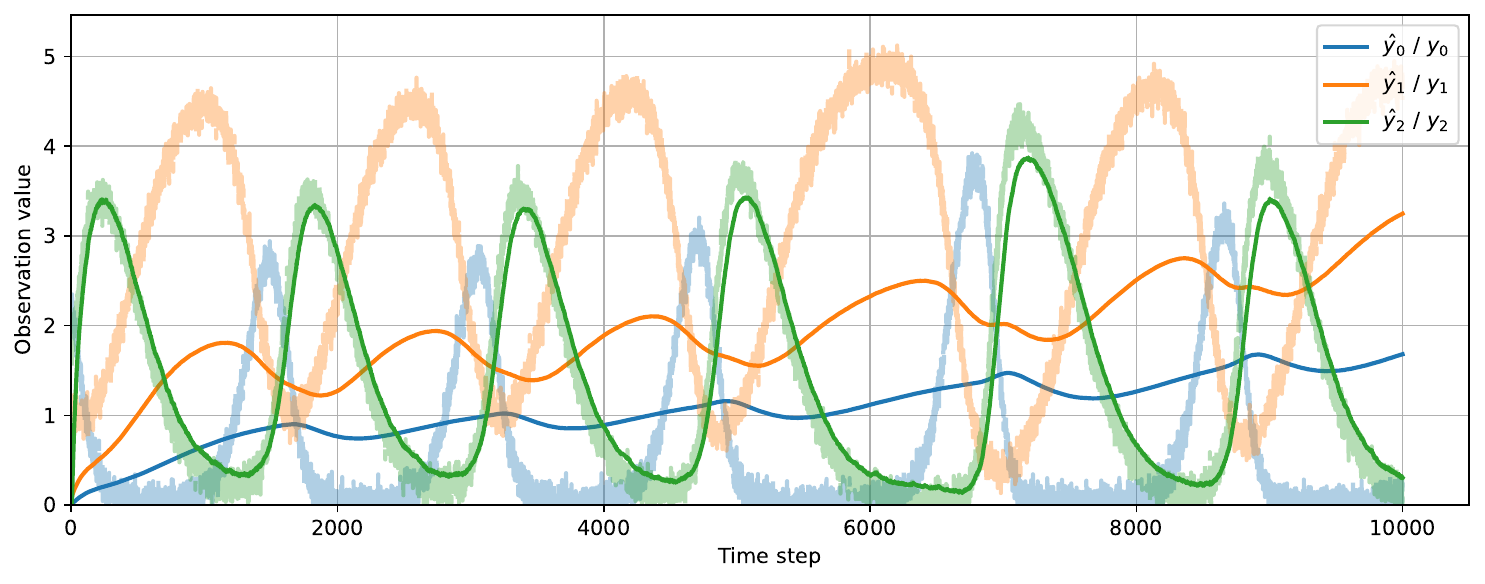}
        \caption{$\frac{E_{\Pi_y}}{E_{\Pi_x}}=1$}
        \label{fig:y_scenario=1_kx=3_ratio=1}
    \end{subfigure}\hfill
    \begin{subfigure}[t]{0.48\textwidth}
        \centering
        \includegraphics[width=\linewidth]{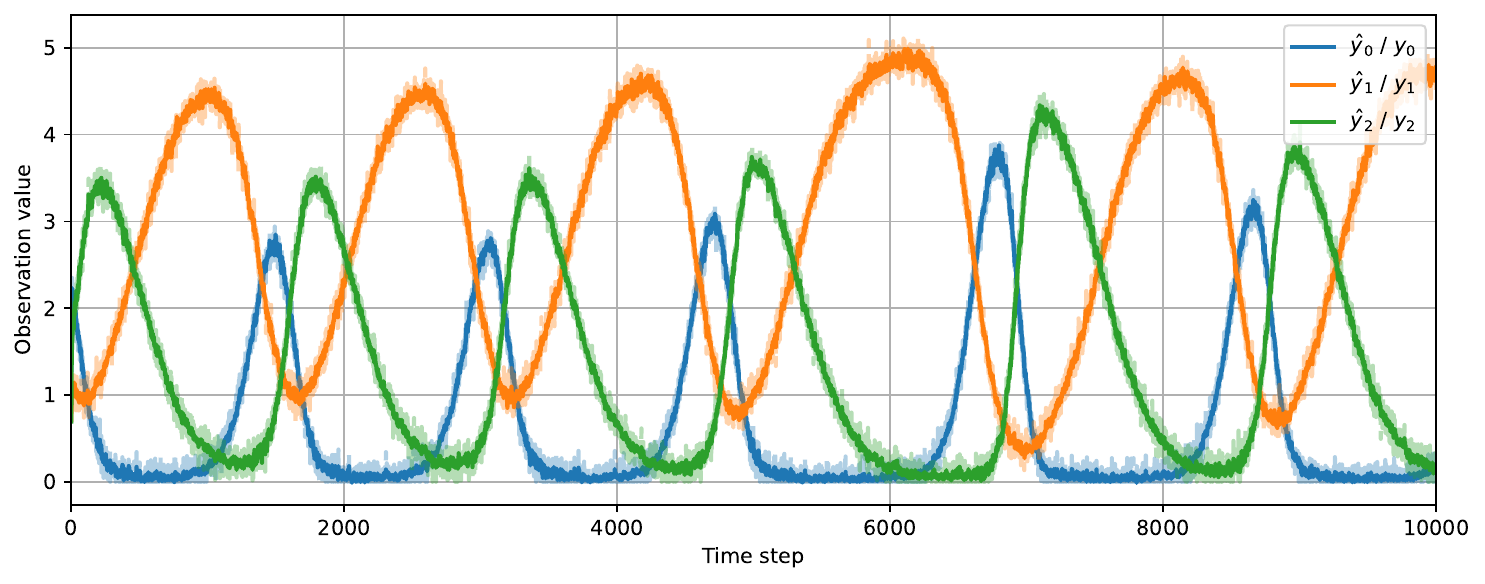}
        \caption{$\frac{E_{\Pi_y}}{E_{\Pi_x}}=1$}
        \label{fig:y_scenario=2_kx=3_ratio=1}
    \end{subfigure}

\end{figure}

\begin{figure}[H]\ContinuedFloat
    \centering

    \begin{subfigure}[t]{0.48\textwidth}
        \centering
        \includegraphics[width=\linewidth]{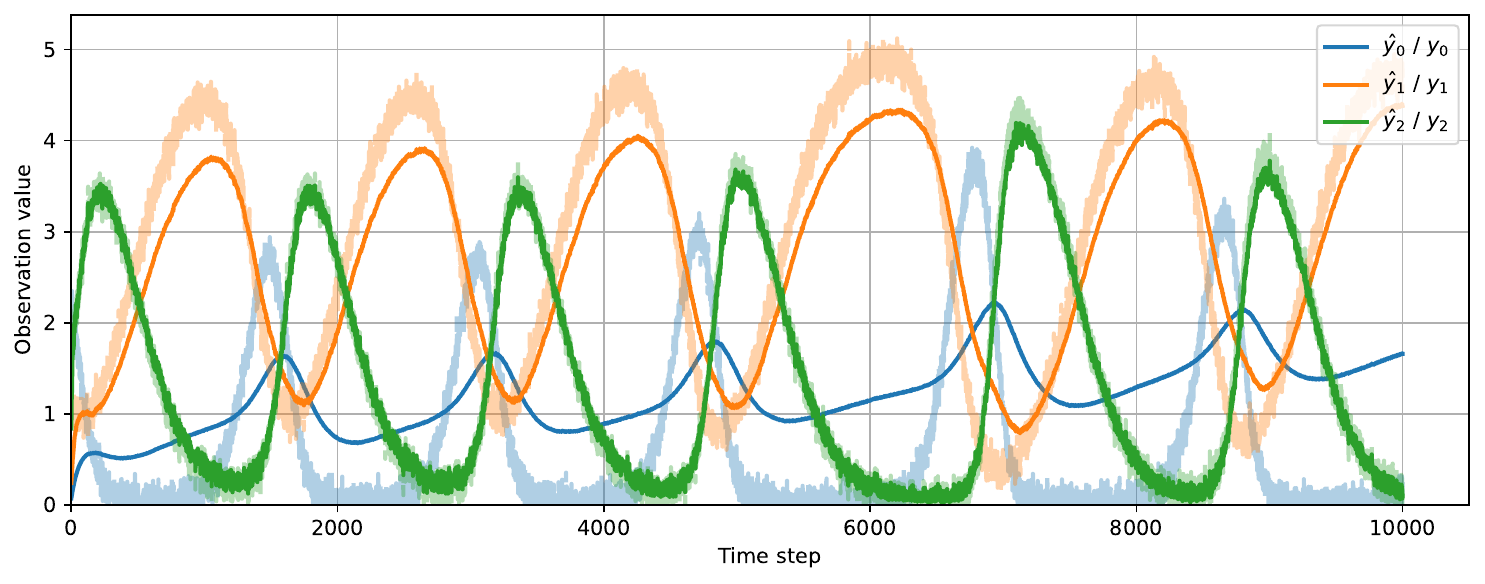}
        \caption{$\frac{E_{\Pi_y}}{E_{\Pi_x}}=10$}
        \label{fig:y_scenario=1_kx=3_ratio=10}
    \end{subfigure}\hfill
    \begin{subfigure}[t]{0.48\textwidth}
        \centering
        \includegraphics[width=\linewidth]{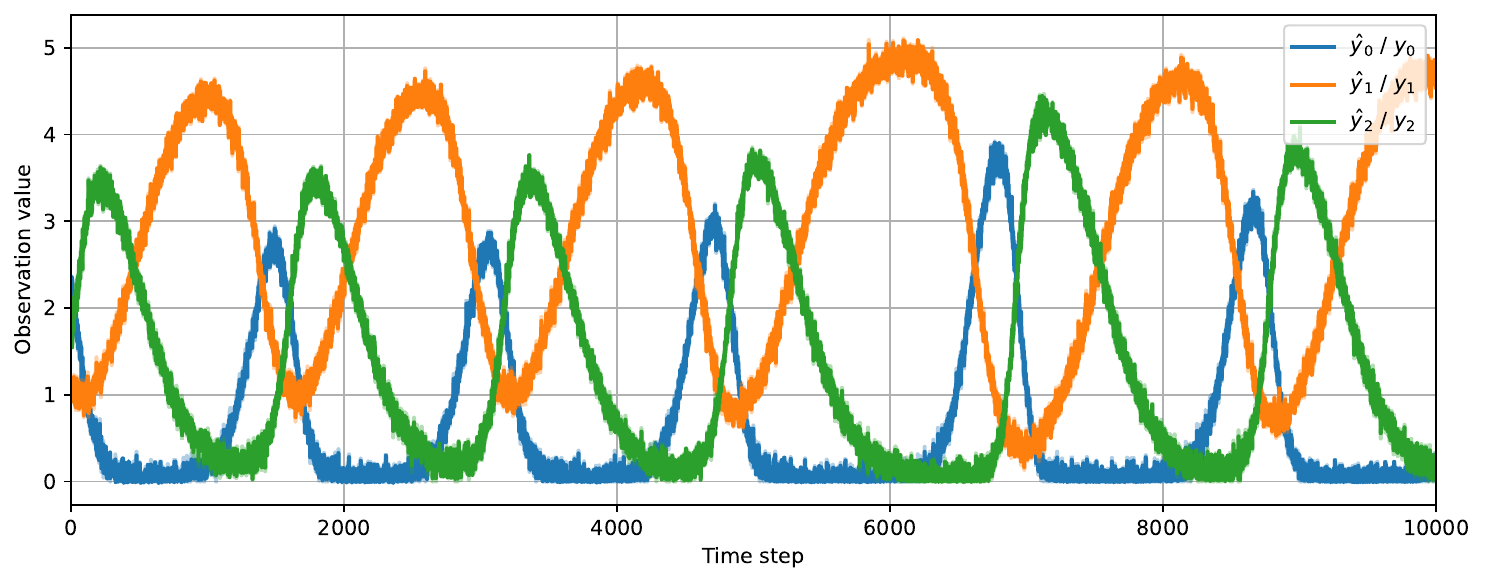}
        \caption{$\frac{E_{\Pi_y}}{E_{\Pi_x}}=10$}
        \label{fig:y_scenario=2_kx=3_ratio=10}
    \end{subfigure}

    \vspace{0.3cm}

    \begin{subfigure}[t]{0.48\textwidth}
        \centering
        \includegraphics[width=\linewidth]{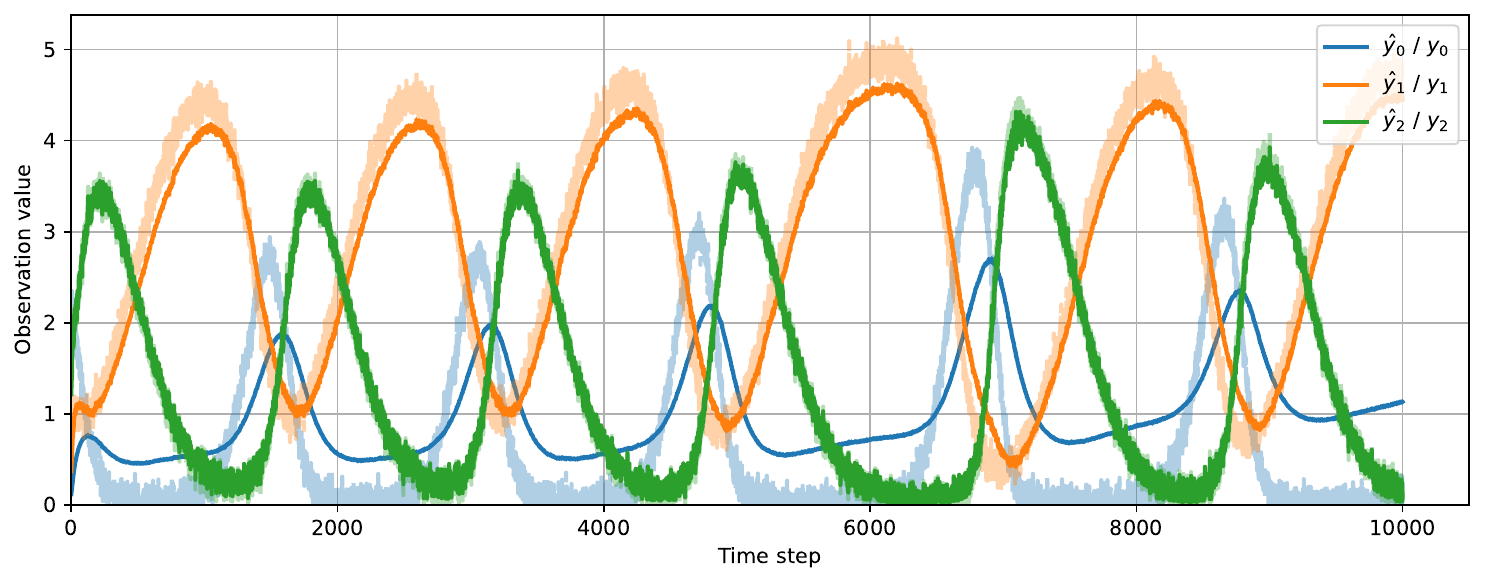}
        \caption{$\frac{E_{\Pi_y}}{E_{\Pi_x}}=25$}
        \label{fig:y_scenario=1_kx=3_ratio=25}
    \end{subfigure}\hfill
    \begin{subfigure}[t]{0.48\textwidth}
        \centering
        \includegraphics[width=\linewidth]{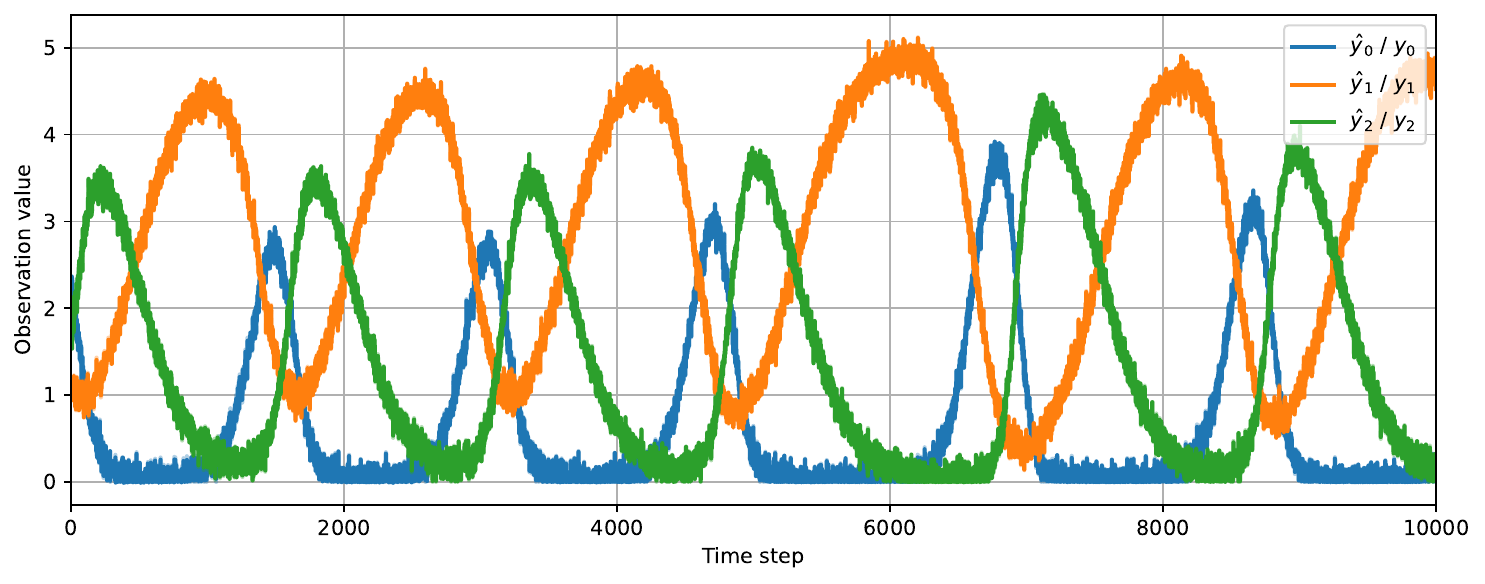}
        \caption{$\frac{E_{\Pi_y}}{E_{\Pi_x}}=25$}
        \label{fig:y_scenario=2_kx=3_ratio=25}
    \end{subfigure}

    \vspace{0.3cm}

    \begin{subfigure}[t]{0.48\textwidth}
        \centering
        \includegraphics[width=\linewidth]{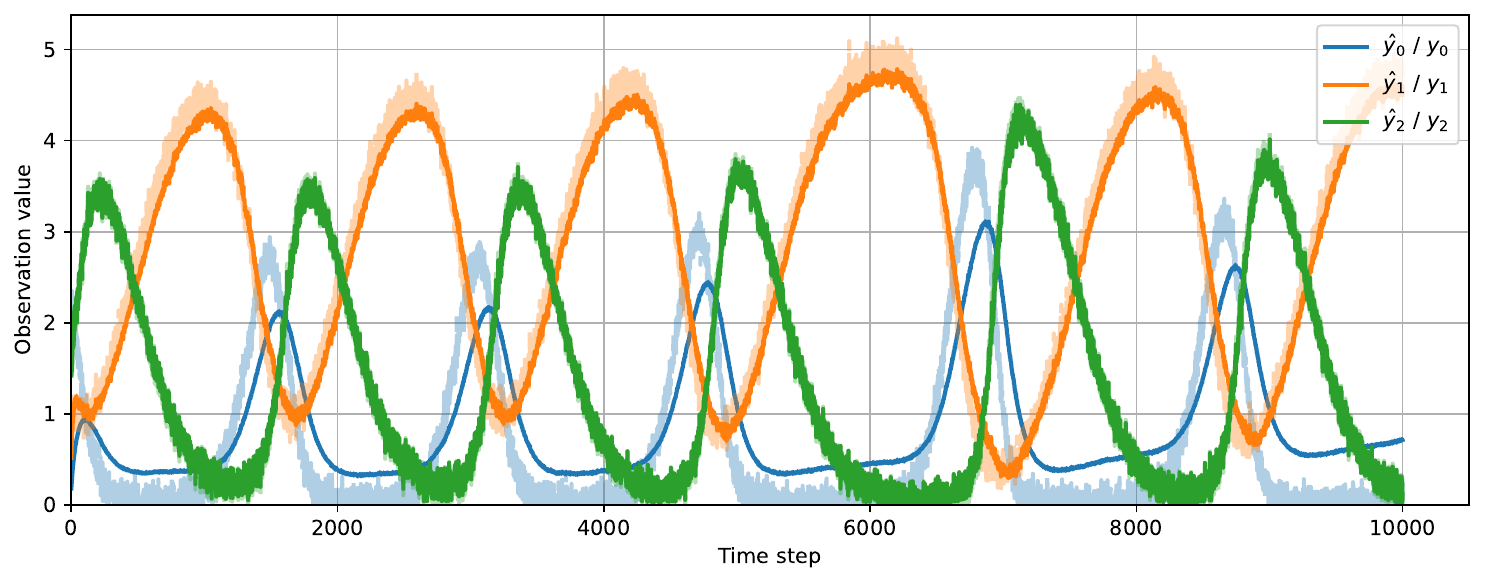}
        \caption{$\frac{E_{\Pi_y}}{E_{\Pi_x}}=50$}
        \label{fig:y_scenario=1_kx=3_ratio=50}
    \end{subfigure}\hfill
    \begin{subfigure}[t]{0.48\textwidth}
        \centering
        \includegraphics[width=\linewidth]{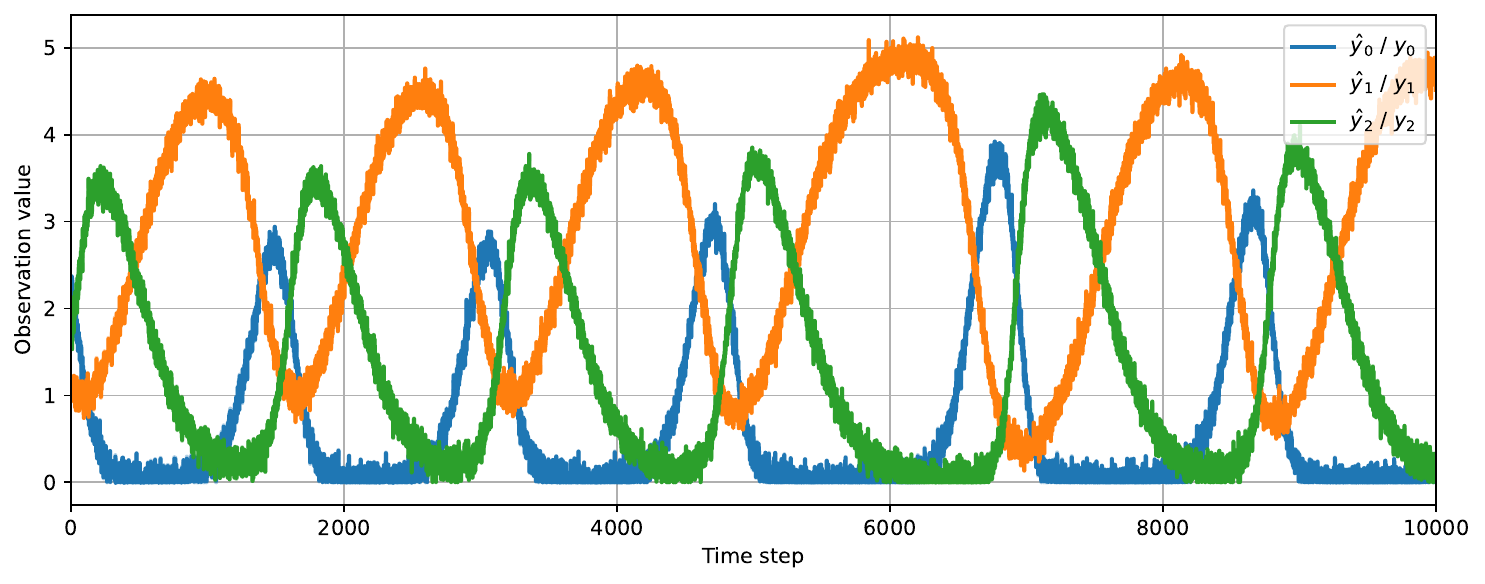}
        \caption{$\frac{E_{\Pi_y}}{E_{\Pi_x}}=50$}
        \label{fig:y_scenario=2_kx=3_ratio=50}
    \end{subfigure}

    \caption[]{The predicted sensations, $\hat{y}$, for $k_x=3$ orders of motion. The left panel is for scenario-different: Lorenz-GM vs. GLV-GP, and the right panel is for scenario-same: GLV-GM vs. GLV-GP. Each row corresponds to a different precision prior ratio.}
\end{figure}

\section{\texorpdfstring{Convergence of $\mu_{\lambda^y}$}{Convergence of mu-lambda-y}}
\label{app:lambda_y}
For a given any precision prior ratio (highlighted by different colour per figure), we show the behaviour of the approximate posterior means $\bmmu_{\lambda^y}$ for $k_x=2$ and $k_x=3$ orders of motion and for both scenarios: 1) Lorenz-GM vs. GLV-GP, and 2) GLV-GM vs. GLV-GP. For each precision prior precision, we plot the $\bmmu_{\lambda^y}$ components corresponding to the GM with the lowest FA. In the plots that follow, the shaded bands represent Bayesian credible intervals which correspond to the regions of the posterior within two standard deviations of the mean.

\subsection{\texorpdfstring{$k_x = 2$}{k x = 2}}
\begin{figure}[H]
    \centering

    \begin{subfigure}{0.9\textwidth}
        \centering
        \includegraphics[width=\linewidth]{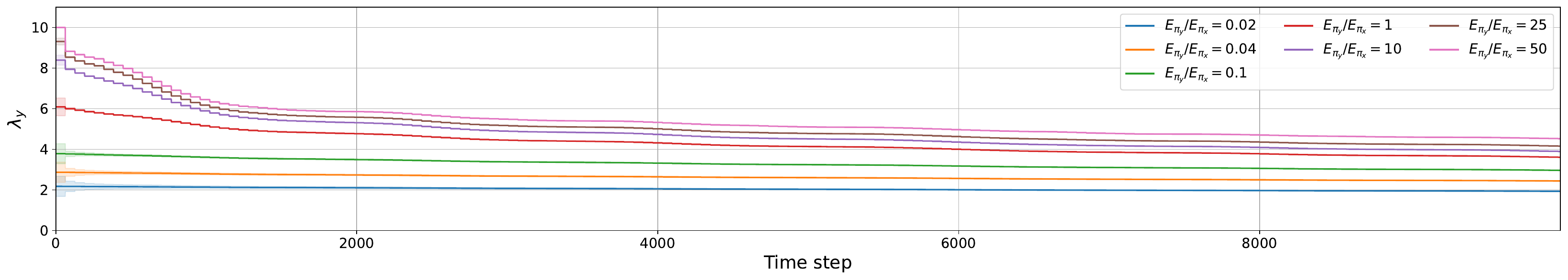}
        \caption{}
        \label{fig:lambda_y_scenario=1_kx=2}
    \end{subfigure}

    \vspace{0.4cm}

    \begin{subfigure}{0.9\textwidth}
        \centering
        \includegraphics[width=\linewidth]{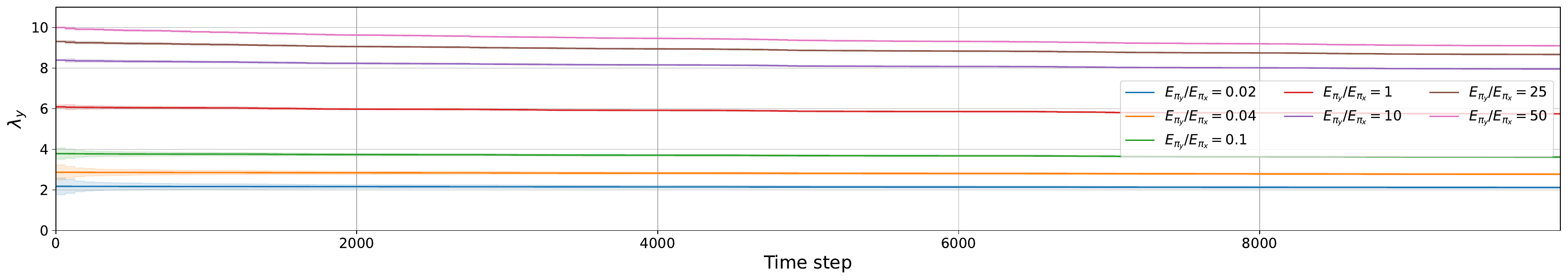}
        \caption{}
        \label{fig:lambda_y_scenario=2_kx=2}
    \end{subfigure}

    \caption{The evolution of $\mu_{\lambda^y}$ posterior estimate in a) scenario-different and b) scenario-same,  with $k_x=2$ orders of motion and along 7 precision prior ratios. The solid lines represent the posterior means at a given time, and the shaded bands represent credible regions within two standard deviations of the mean.}
    \label{fig:lambda_y_kx=2}
\end{figure}

In scenario-different, Fig.~\ref{fig:lambda_y_scenario=1_kx=2} shows that the ODEM scheme eventually stabilises the posterior estimate $\bmmu_{\lambda^y}$. Interestingly, if the $\frac{E_{\Pi_y}}{E_{\Pi_x}}\leq 1$, there are nearly no changes in the posterior estimates. We can also see that the initial uncertainty (i.e., the Bayesian credible interval plotted as the shaded area around the posterior estimates) shrinks very quickly early on. However, when we look at Fig.~\ref{fig:lambda_y_scenario=2_kx=2} for scenario-same, it is evident that the ODEM scheme does not even need the posterior updates for $\bmlambda^y$, since the GLV-GM has the identical state dynamics to that of the GLV-GP, which means the \emph{D}-step on its own is enough for minimising the VFE and tracking the GP.

\subsection{\texorpdfstring{$k_x = 3$}{k x = 3}}
\begin{figure}[H]
    \centering

    \begin{subfigure}{0.9\textwidth}
        \centering
        \includegraphics[width=\linewidth]{figures/summary_plots/lorenz-glv/lambda_y_kx3_d0.pdf}
        \caption{}
        \label{fig:lambda_y_scenario=1_kx=3}
    \end{subfigure}

    \vspace{0.4cm}

    \begin{subfigure}{0.9\textwidth}
        \centering
        \includegraphics[width=\linewidth]{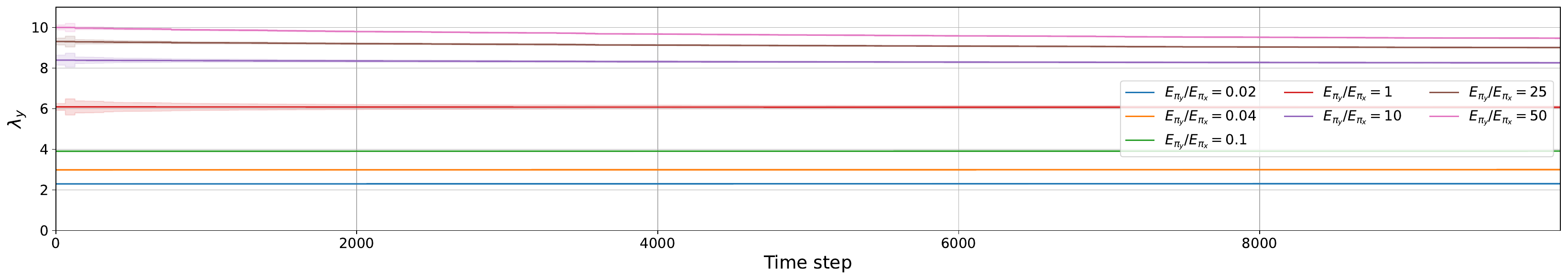}
        \caption{}
        \label{fig:lambda_y_scenario=2_kx=3}
    \end{subfigure}

    \caption{The evolution of $\mu_{\lambda^y}$ posterior estimate in a) scenario-different and b) scenario-same with $k_x=3$ orders of motion and along 7 precision prior ratios. The solid lines represent the posterior means at a given time, and the shaded bands represent credible regions within two standard deviations of the mean.}
    \label{fig:lambda_y_kx=3}
\end{figure}

We observe a similar behaviour to the previous section, where we had $k_x=2$ orders of motion. If we compare Fig.~\ref{fig:lambda_y_scenario=1_kx=3} with Fig.~\ref{fig:lambda_y_scenario=1_kx=2}, we note that since with a higher orders of motion, it looks as if the ODEM scheme needs even less changes to the posterior means  $\bmmu_{\lambda^y}$, but still needs to update them. Similar to Fig.~\ref{fig:lambda_y_scenario=2_kx=2}, again we see no much changes in the carves in Fig.~\ref{fig:lambda_y_scenario=2_kx=3}, since the \emph{D}-step is sufficient to minimise the VFE for the GLV-GM.  

\section{\texorpdfstring{Convergence of $\mu_{\lambda^x}$}{Convergence of mu-lambda-x}}
\label{app:lambda_x}
In this section, given any particular precision prior ratio, the evolution of the posterior estimate $\mu_{\lambda^x}$ for $k_x=2$ and $k_x=3$ orders of motion and for both scenarios: 1) Lorenz-GM vs. GLV-GP, and 2) GLV-GM vs. GLV-GP is demonstrated. For each precision prior ratio, we plot the $\bmmu_{\lambda^x}$ components corresponding with the GM with lowest FA. The shaded region around each curve denotes the Bayesian credible interval and corresponds to two standard deviations above and below the posterior mean. 

\subsection{\texorpdfstring{$k_x = 2$}{k x = 2}}
\begin{figure}[H]
    \centering

    \begin{subfigure}{0.9\textwidth}
        \centering
        \includegraphics[width=\linewidth]{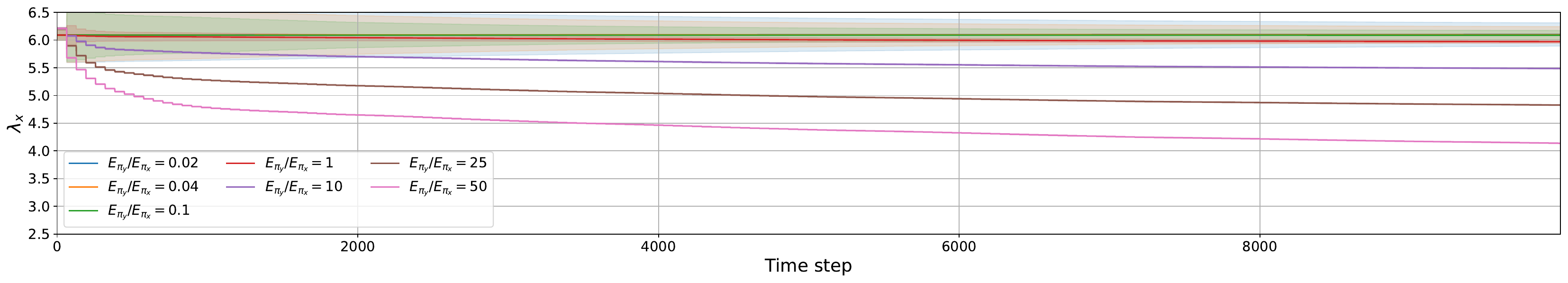}
        \caption{}
        \label{fig:lambda_x_scenario=1_kx=2}
    \end{subfigure}

    \vspace{0.4cm}

    \begin{subfigure}{0.9\textwidth}
        \centering
        \includegraphics[width=\linewidth]{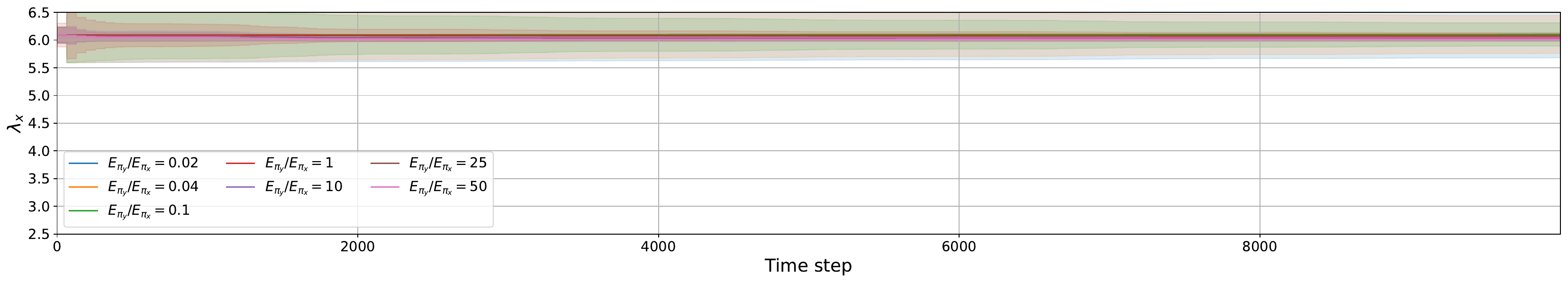}
        \caption{}
        \label{fig:lambda_x_scenario=2_kx=2}
    \end{subfigure}

    \caption{The evolution of $\mu_{\lambda^x}$ posterior estimate in a) scenario-different and b) scenario-same,  with $k_x=2$ orders of motion and along 7 precision prior ratios. The solid lines represent the posterior means at a given time, and the shaded bands represent credible regions within two standard deviations of the mean.}
    \label{fig:lambda_x_kx=2}
\end{figure}

Looking at Fig.~\ref{fig:lambda_x_scenario=1_kx=2}, we can see that the ODEM scheme stabilises the posterior estimate $\mu_{\lambda^x}$ as time goes by. Interestingly, if the $\frac{E_{\Pi_y}}{E_{\Pi_x}}\leq 1$, we see nearly no changes in either the posterior estimates or the uncertainty (i.e., Bayesian credible interval plotted as shaded area around the posterior estimates) around them. However, when we look at Fig.~\ref{fig:lambda_x_scenario=2_kx=2}, it is evident that the ODEM scheme does not even need to update the posterior estimates $\mu_{\lambda^x}$, since the GLV-GM has the identical state dynamics to that of the GLV-GP, which means the \emph{D}-step on its own is enough for minimising the VFE and track the GP.

\subsection{\texorpdfstring{$k_x = 3$}{k x = 3}}
For $k_x=3$ orders of motion, we see something very similar to the previous case with $k_x=2$ orders of motion.
\begin{figure}[H]
    \centering

    \begin{subfigure}{0.9\textwidth}
        \centering
        \includegraphics[width=\linewidth]{figures/summary_plots/lorenz-glv/lambda_x_kx3_d0.pdf}
        \caption{}
        \label{fig:lambda_x_scenario=1_kx=3}
    \end{subfigure}

    \vspace{0.4cm}

    \begin{subfigure}{0.9\textwidth}
        \centering
        \includegraphics[width=\linewidth]{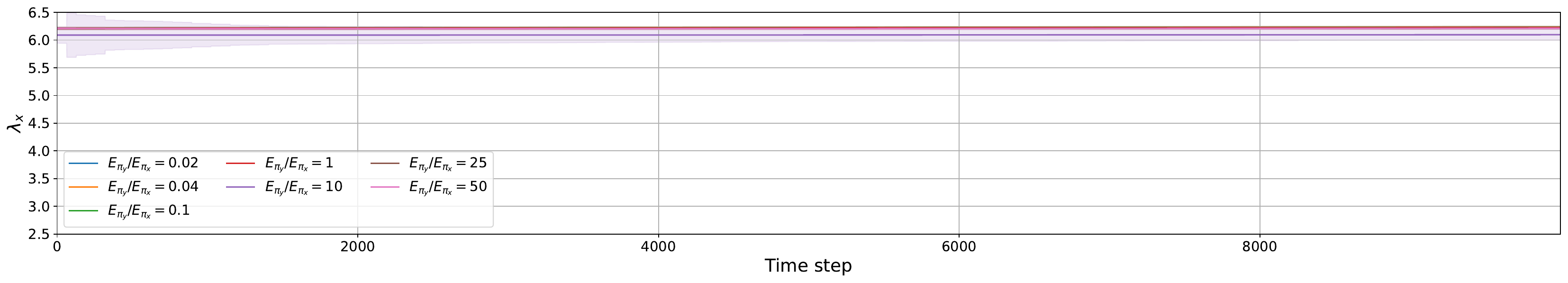}
        \caption{}
        \label{fig:lambda_x_scenario=2_kx=3}
    \end{subfigure}

    \caption{The evolution of $\mu_{\lambda^x}$ posterior estimate in a) scenario-different and b) scenario-same,  with $k_x=3$ orders of motion and along 7 precision prior ratios. The solid lines represent the posterior means at a given time, and the shaded bands represent credible regions within two standard deviations of the mean.}
    \label{fig:lambda_x_kx=3}
\end{figure}

\section{\texorpdfstring{Convergence of $\mu_\theta$}{Convergence of mu-theta}}
\label{app:theta}
Here, for both scenarios, we can see a clear stabilised posterior estimate for $\mu_\theta$, along $k_x=2$ and $k_x=3$ orders of motion along each precision prior ratio. Recall that in scenario-different: Lorenz-GM vs. GLV-GP, the Lorenz-GM has only one parameter to estimate, that is, $\rho$ parameter, whereas in scenario-same: GLV-GM vs. GLV-GP, the GLV-GM has 3 parameters, which are the elements above the diagonal elements of the $A$ matrix: $a_{12},a_{13},$ and $a_{23}$. The shaded region around each curve denotes the Bayesian credible interval and corresponds to two standard deviations above and below the posterior mean. 

\subsection{\texorpdfstring{$k_x = 2$}{k x = 2}}
\begin{figure}[H]
    \centering
    \includegraphics[width=0.9\textwidth]{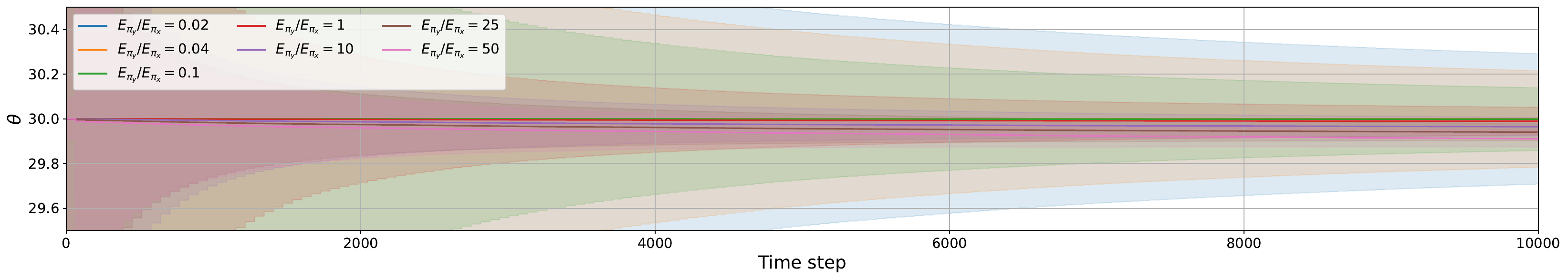}
    \caption{The evolution of the posterior expectation over $\rho$ in scenario-different, with $k_x = 2$ orders of motion across seven precision prior ratios. The solid lines represent the posterior means at a given time, and the shaded bands represent credible regions within two standard deviations of the mean.}
    \label{fig:theta_scenario1_kx2}
\end{figure}

For scenario-different, we can see in Fig.~\ref{fig:theta_scenario1_kx2} that the Lorenz-GM simply does not make much change to the posterior $\mu_{\theta}$ estimates. Interestingly, the posterior uncertainty around these estimates shrink as time passes. 

\begin{figure}[H]
    \centering

    \begin{subfigure}{0.9\textwidth}
        \centering
        \includegraphics[width=\linewidth]{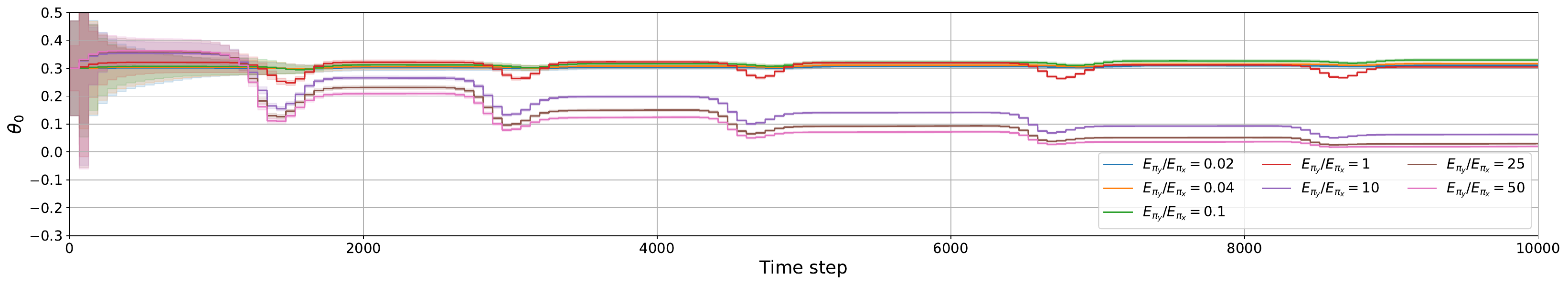}
        \caption{}
        \label{fig:theta_scenario2_kx2_d0}
    \end{subfigure}

    \vspace{0.35cm}

    \begin{subfigure}{0.9\textwidth}
        \centering
        \includegraphics[width=\linewidth]{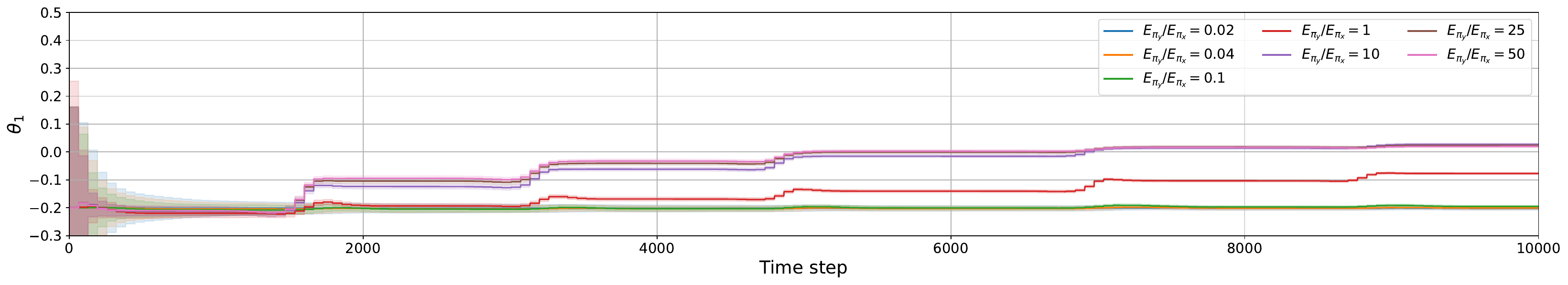}
        \caption{}
        \label{fig:theta_scenario2_kx2_d1}
    \end{subfigure}

    \vspace{0.35cm}

    \begin{subfigure}{0.9\textwidth}
        \centering
        \includegraphics[width=\linewidth]{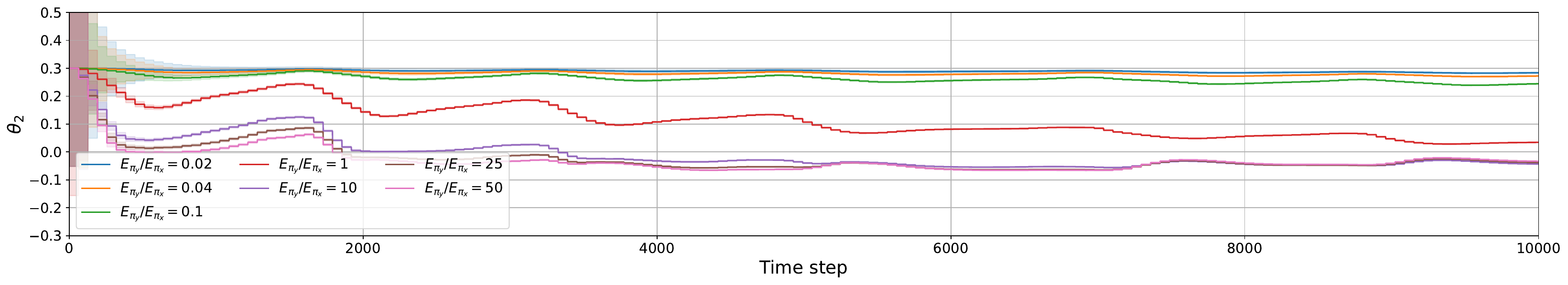}
        \caption{}
        \label{fig:theta_scenario2_kx2_d2}
    \end{subfigure}

    \caption{The evolution of the parameters $a_{12}$, $a_{13}$, and $a_{23}$ of the matrix $A$ in GLV-GM for the scenario-same case with $k_x = 2$ orders of motion, shown in subfigures (a), (b), and (c), respectively, across seven precision prior ratios. The solid lines represent the posterior means at a given time, and the shaded bands represent credible regions within two standard deviations of the mean.}
    \label{fig:theta_scenario2_kx2}
\end{figure}

For scenario-same, we can see the posterior estimates of the 3 parameters in Fig.~\ref{fig:theta_scenario2_kx2_d0}, Fig.~\ref{fig:theta_scenario2_kx2_d1} and Fig.~\ref{fig:theta_scenario2_kx2_d2}. In all cases, we do see a slow stabilisation as time passes, with a quick shrinkage in posterior uncertainty.

\subsection{\texorpdfstring{$k_x = 3$}{k x = 3}}

\begin{figure}[H]
    \centering
    \includegraphics[width=0.9\textwidth]{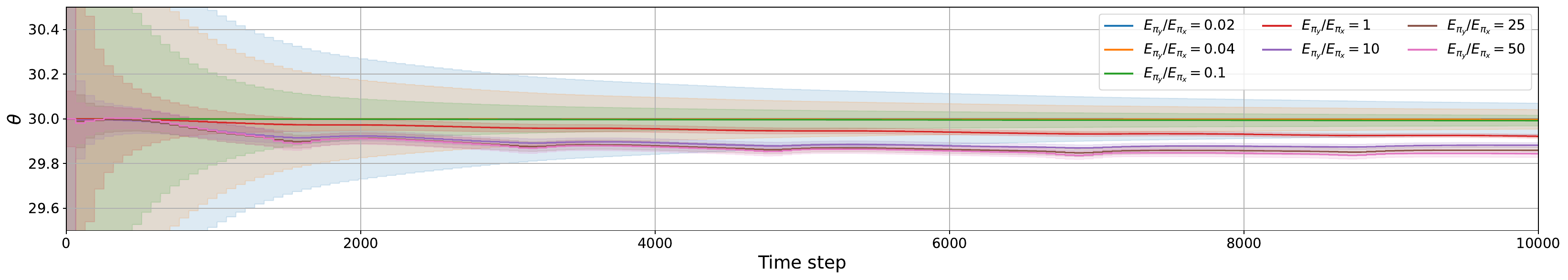}
    \caption{The evolution of the posterior expectation over $\rho$ in scenario-different, with $k_x = 3$ orders of motion across seven precision prior ratios. The solid lines represent the posterior means at a given time, and the shaded bands represent credible regions within two standard deviations of the mean.}
    \label{fig:theta_scenario1_kx3}
\end{figure}

For scenario-different, we can see in Fig.~\ref{fig:theta_scenario1_kx3} that the Lorenz-GM updates the $\mu_{\theta}$ estimates (more than Fig.~\ref{fig:theta_scenario1_kx2}), with a much quicker shrinkage in posterior uncertainty. Most likely, the higher orders of motion, has enabled the GM to be more certain about its parameter estimates.

\begin{figure}[H]
    \centering

    \begin{subfigure}{0.9\textwidth}
        \centering
        \includegraphics[width=\linewidth]{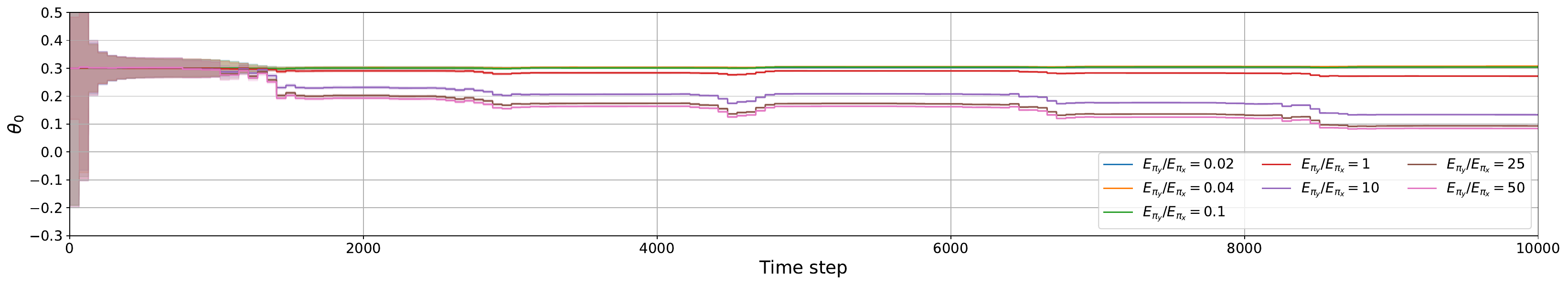}
        \caption{}
        \label{fig:theta_scenario2_kx3_d0}
    \end{subfigure}

    \vspace{0.35cm}

    \begin{subfigure}{0.9\textwidth}
        \centering
        \includegraphics[width=\linewidth]{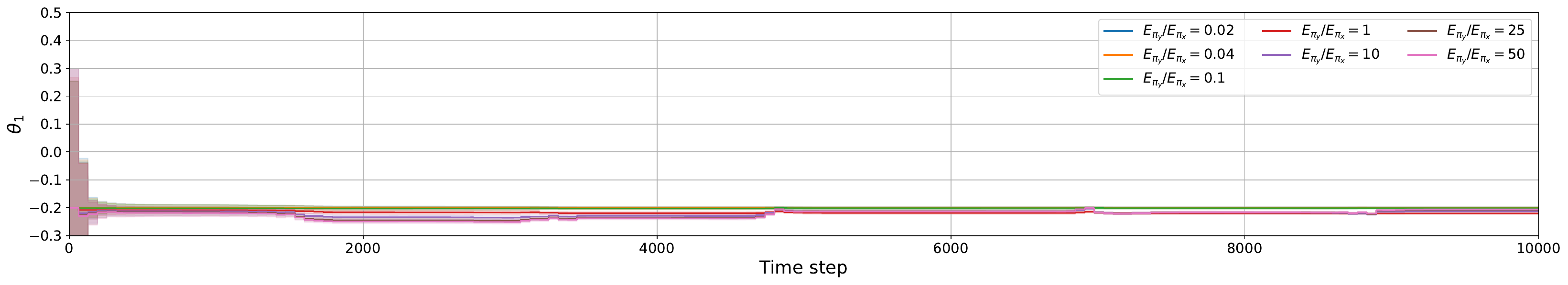}
        \caption{}
        \label{fig:theta_scenario2_kx3_d1}
    \end{subfigure}

    \vspace{0.35cm}

    \begin{subfigure}{0.9\textwidth}
        \centering
        \includegraphics[width=\linewidth]{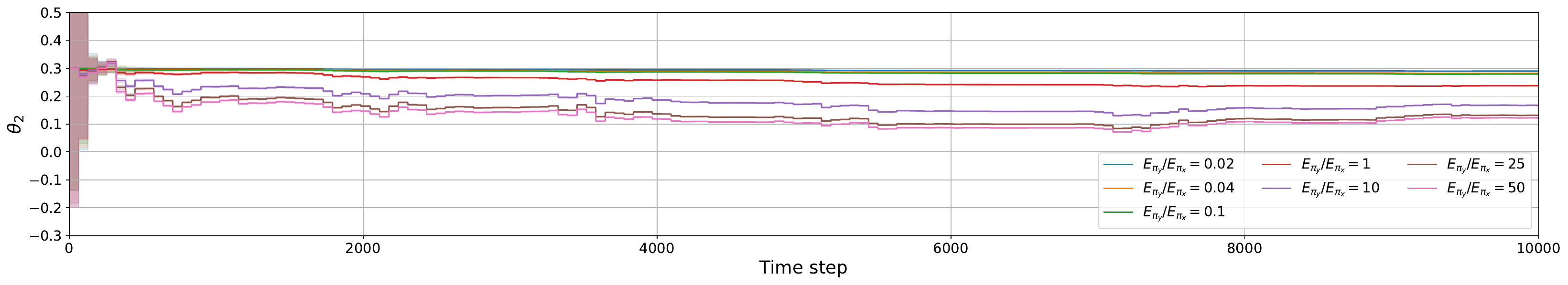}
        \caption{}
        \label{fig:theta_scenario2_kx3_d2}
    \end{subfigure}

    \caption{The evolution of the parameters $a_{12}$, $a_{13}$, and $a_{23}$ of the matrix $A$ in GLV-GM for the scenario-same case with $k_x = 3$ orders of motion, shown in subfigures (a), (b), and (c), respectively, across seven precision prior ratios. The solid lines represent the posterior means at a given time, and the shaded bands represent credible regions within two standard deviations of the mean.}
    \label{fig:theta_scenario2_kx3}
\end{figure}

For scenario-same, we can see the posterior estimates of the 3 parameters in Fig.~\ref{fig:theta_scenario2_kx3_d0}, Fig.~\ref{fig:theta_scenario2_kx3_d1} and Fig.~\ref{fig:theta_scenario2_kx3_d2}. In all cases, we do see a slow stabilisation as time passes, with a quick shrinkage in posterior uncertainty.

Note that the presented models are based on a VFE-based model selection process across various parameter/\allowbreak hyperparameter priors. Conditional on this choice, it would appear that the posterior updates on the hidden states $\bfx_t$ sufficiently "fit the model". As these are \emph{local} variables in the context of the model---directly interacting only with the data $\bfy_t$---they are a lot less constrained. In contrast, the parameter, $\bmtheta$  precision, $\bmlambda$, estimations are more inflexible. The gradient accumulation prevents the updates from over-fitting to a single time point; the changes must explain the behaviour in the data on the \emph{global} scale. Whilst the posteriors of the precisions seem very stable, those of the parameters exhibit some evolution coinciding with the peaks in the data. This suggests that these peaks indeed contain enough information to move the posterior mass towards different values in the parameter space. In terms of the VFE, the numerous local updates on the hidden states reduce the FA more readily then the global updates on the parameters/hyperparameters, and whilst these are assigned well-fitting priors through the model selection, they are not immutable in the presence of sufficient information.

\end{document}